\documentclass{elsarticle}

\usepackage[a4paper,margin=3cm]{geometry}
\usepackage{amssymb}
\usepackage{amsmath}
\usepackage{booktabs}
\usepackage{graphicx}
\usepackage{adjustbox}
\usepackage{float}
\usepackage[hypcap=false]{caption}
\usepackage{subcaption}

\let\PTIncludeGraphics\includegraphics
\renewcommand{\includegraphics}[2][]{%
  \par\centering
  \adjustbox{max width=\linewidth,center}{%
    \PTIncludeGraphics[keepaspectratio,#1]{#2}%
  }%
  \par
}
\usepackage{hyperref}
\usepackage{tabularx}
\usepackage{array}
\usepackage{pdflscape}
\usepackage{url}
\usepackage{enumitem}
\usepackage{hhline}
\usepackage{makecell}
\usepackage{ragged2e}

\makeatletter
\def\ps@pprintTitle{%
 \let\@oddhead\@empty
 \let\@evenhead\@empty
 \def\@oddfoot{}%
 \let\@evenfoot\@oddfoot}
\makeatother

\journal{...}

\begin{document}

\begin{frontmatter}

\title{Dynamic Multi-Pair Trading Strategy in Cryptocurrency Markets with Deep Reinforcement Learning\tnoteref{tn1}}

\tnotetext[tn1]{The replication of the full research is possible due to public GitHub repository which can be accessed at: https://github.com/damianlebiedz/pair-trading-with-rl}

\author[1,fn1]{Damian Lebiedź}

\author[2,fn1]{Robert Ślepaczuk}

\fntext[fn1]{Email addresses: damian.lebiedz@gmail.com (Damian Lebiedź, corresponding author) ORCID: 0009-0008-5762-075X,  rslepaczuk@wne.uw.edu.pl (Robert Ślepaczuk) ORCID: 0000-0001-5227-2014}

\affiliation[1]{organization={Faculty of Economic Sciences, University of Warsaw},
            addressline={ul. Długa 44/50}, 
            city={Warsaw},
            postcode={00-241}, 
            country={Poland}}

\affiliation[2]{organization={Quantitative Finance Research Group, Department of Quantitative Finance and Machine Learning, Faculty of Economic Sciences, University of Warsaw},
            addressline={ul. Długa 44/50}, 
            city={Warsaw},
            postcode={00-241}, 
            country={Poland}}

\begin{abstract}
This study aims to determine whether the application of Deep Reinforcement Learning (DRL) as a specialized execution overlay can enhance pair trading in highly volatile cryptocurrency markets. Although classical implementations of the strategy have proven successful in traditional equities, they frequently exhibit rigidity and suffer from severe divergence risks when applied to high-variance environments. To construct a robust system, we developed a hierarchical "Filter-then-Rank" pair selection methodology and a custom "Fixed Risk, Adaptive Mean" execution model. The system employs a Proximal Policy Optimization (PPO) agent with a Long Short-Term Memory (LSTM) layer to govern execution decisions within strict deterministic risk management boundaries. Evaluated on 1-hour interval data from the Binance USD-M Futures market, the selected RL policy achieved an Out-Of-Sample performance that substantially outperformed the heuristic baseline, although the results varied considerably across five random seeds, with four of the five seeds exceeding the baseline in both CAGR and Sortino Ratio. A stationary circular block bootstrap test of the Agent's risk-adjusted outperformance yields p-values of approximately 0.06, which fall just short of the conventional 5\% significance threshold. Ultimately, this study contributes to the quantitative finance literature by introducing a hybrid architecture that combines statistical arbitrage with DRL execution policies. Furthermore, it applies safe reinforcement learning via deterministic shielding, with results indicating that anchoring a neural policy to statistically robust boundaries mitigates severe divergence risks.
\end{abstract}

\begin{keyword}
Deep Reinforcement Learning \sep Deterministic Shielding \sep Hierarchical Systems \sep Quantitative Finance \sep Statistical Arbitrage
\end{keyword}

\end{frontmatter}

\section{Introduction}
\label{sec:intro}
Statistical arbitrage encompasses a wide range of strategies designed to exploit temporary pricing anomalies and structural inefficiencies between linked assets. Although classical methodologies have been successful in traditional equities, cryptocurrency markets are characterized by high volatility and frequent regime changes \cite{Ardia2019}, which introduce persistent frictions and divergence risks to arbitrage strategies \cite{Makarov2020}. Consequently, traditional pair trading is significantly more difficult to implement in this domain \cite{Fischer2019}. The primary objective of this study is to propose and evaluate a Dynamic Multi-Pair Trading Strategy that utilizes Deep Reinforcement Learning (DRL) as a specialized execution overlay. This research is of significant importance because it addresses the rigidity of traditional implementations of the strategy, proposing a more adaptive, hybrid framework.\\

The main aim of this study is synthesized in the following hypotheses:\\
\textbf{H1}: \emph{A dynamic statistical baseline, driven by a hierarchical pair selection framework and a custom execution concept, outperforms traditional benchmarks in the Out-Of-Sample period.}\\
\textbf{H2}: \emph{Deploying Deep Reinforcement Learning as an execution overlay enhances the profitability and risk-adjusted performance of a heuristic statistical arbitrage baseline.}\\

Moreover, we ask the following research questions:\\
\textbf{RQ1}: \emph{Can a dynamic, multi-pair statistical arbitrage strategy capture a persistent edge within the highly volatile cryptocurrency market?}\\
\textbf{RQ2}: \emph{Does deploying Deep Reinforcement Learning as an execution overlay enhance the profitability of a heuristic baseline?}\\
\textbf{RQ3}: \emph{How do different reinforcement learning configurations, specifically the design of reward functions and observation spaces, or the implementation of a risk management overlay, impact the Out-Of-Sample robustness in high-noise trading environments?}\\

To navigate the complexities of cryptocurrency markets, this study constructs a dynamic statistical baseline utilizing a hierarchical "Filter-then-Rank" pair selection methodology and a custom "Fixed Risk, Adaptive Mean" execution model. This framework trades a "snapshot equilibrium" rather than relying on static \cite{Gatev2006} or continuous rolling \cite{avellaneda2010, Krauss2017} mean-reversion. Furthermore, while contemporary quantitative research increasingly explores conventional end-to-end RL architectures, where agents attempt to simultaneously extract predictive signals and learn execution policies directly from raw market data \cite{guijarro2021}, these unconstrained models frequently suffer from extreme policy instability and Out-Of-Sample overfitting. Therefore, our methodology deliberately decouples these processes. We strictly deploy a Proximal Policy Optimization (PPO) agent with an LSTM architecture exclusively as an execution layer on top of the statistically validated baseline, operating within hard deterministic boundaries. The empirical analysis utilizes 1-hour interval data from the Binance USD-M Futures market, divided into an In-Sample period (2024) and the Out-Of-Sample period (2025).

This research makes several applied and empirical contributions to the quantitative finance literature. First, it provides an empirical application of "safe reinforcement learning via shielding" in high-noise environments by evaluating whether a hybrid architecture can mitigate the brittleness of conventional end-to-end models. Second, it develops a hierarchical pair selection framework that integrates traditional statistical cointegration with strict structural constraints (such as the Hurst exponent and beta coefficients) to systematically isolate high-conviction anomalies. By employing this framework, the study aims to empirically investigate the trade-off between alpha concentration and risk diversification across varying portfolio sizes. Third, it introduces an execution heuristic tailored specifically to absorb the structural drift and regime shifts inherent to highly volatile digital assets, challenging the rigidity of classical continuous frameworks. Finally, the study indicates that anchoring a neural policy to statistically robust boundaries mitigates severe divergence risks, paving the way for safer DRL integration in algorithmic trading. The methodology includes explicit safeguards against look-ahead and survivorship biases.

The structure of this paper is organized as follows. Section 2 provides a comprehensive review of the relevant literature. Section 3 outlines the research methodology, detailing the custom framework, data sources, and the configuration of the DRL algorithm, alongside the evaluation metrics and benchmarks. Section 4 presents the baseline formulation and its phased optimization. Section 5 evaluates the baseline's performance, supported by a sensitivity analysis. Section 6 assesses the strategy using the RL execution overlay, featuring both a sensitivity analysis and a post-hoc ablation study. Ultimately, the final section presents the overarching conclusions of the study, provides answers to the hypotheses and research questions, and outlines directions for future work.\\

\vspace{0.5em}
\section{Literature Review}

Statistical arbitrage has evolved from simple empirical heuristics into complex stochastic and machine learning frameworks. The seminal distance-based approach matched assets by minimum sum of squared historical price deviations during a formation period and traded subsequent divergences, establishing the empirical viability of pair trading in equity markets \citep{Gatev2006}. However, such non-parametric heuristics suffer from severe divergence risks, as they implicitly assume that historical co-movements persist indefinitely while ignoring the underlying forces driving the equilibrium. To formalize this relationship, the literature adopted cointegration as a rigorous econometric alternative, identifying genuine long-term equilibria between non-stationary time series \citep{Engle1987, vidyamurthy2004pairs}.

In the cointegration testing debate, the multivariate Johansen procedure offers theoretical symmetry and higher statistical power, but its eigenvalue decomposition imposes severe computational bottlenecks in high-frequency rolling-window settings; the Engle-Granger two-step heuristic remains highly effective for strictly bivariate spreads and substantially more tractable \citep{johansen1988, Moura2013}. To mitigate the limitations of any single statistical metric, modern frameworks emphasize multi-stage filtering. Augmenting cointegration with explicit structural constraints, in particular the Hurst exponent computed via rescaled-range analysis, helps separate genuine mean-reverting (anti-persistent, $H < 0.5$) spreads from random walks and trending anomalies, while recent contributions further leverage graph clustering to identify topologically interconnected pair universes \citep{Do2010, Hurst1951, Qian2007, ramos2017hurst, korniejczuk2024}.

Beyond pair selection, spread modelling has shifted from static parameters \citep{Gatev2006} toward continu-\
ous-time, time-varying formulations. State-space models with Kalman filtering allow the unobservable spread mean to be dynamically updated as a hidden state, while the Ornstein-Uhlenbeck framework represents the spread as a mean-reverting diffusion process and introduces the dimensionless $s$-score for trade timing based on the spread's half-life \citep{Elliott2005, avellaneda2010}. Despite their theoretical elegance, these continuous models struggle in cryptocurrency markets, which display extreme idiosyncratic volatility, fat-tailed distributions, abrupt regime changes, fragmented liquidity, and substantial execution slippage. Under such conditions, OU mean-reversion speed estimates break down and frictionless assumptions become unrealistic, rendering theoretically optimal continuous hedging practically unviable \citep{Ardia2019, Fischer2019, Krauss2017, Makarov2020}.

To address these non-linearities, Deep Reinforcement Learning (DRL) has emerged as a flexible alternative for algorithmic trading and dynamic portfolio management, offering the capacity to learn execution policies directly from market data without rigid econometric assumptions \citep{jiang2017cryptocurrency, liu2020}. Within statistical arbitrage specifically, DRL agents have been shown to optimize pair-trading execution and dynamically adapt position sizing to changing market variance, outperforming static econometric baselines in cryptocurrency settings \citep{kim2019, Vergara2024, yang2024}. Closest to our setting, \citet{Vergara2024} apply DRL to a statistical arbitrage strategy in the cryptocurrency market; this study retains the statistical pair selection and heuristic baseline, and evaluates the DRL agent as an execution overlay constrained by deterministic risk rules. Architecturally, modern implementations increasingly replace value-based methods with actor-critic algorithms, particularly Proximal Policy Optimization (PPO), whose clipped objective enforces a trust region that prevents destructive policy updates in volatile regimes; LSTM augmentations further address the partial observability of financial time series \citep{schulman2017ppo, engstrom2020, hochreiter1997}.

An equally critical design dimension is the reward function. Standard RL maximizes expected returns without regard for variance, which encourages excessive risk-taking in financial environments. Drawing on Prospect Theory and the formalization of risk-sensitive RL, asymmetric reward shaping that penalizes losses more heavily than it rewards gains internalizes downside risk and stabilizes policy learning \citep{kahneman1979, mihatsch2002}. Nevertheless, unconstrained DRL applied directly to financial series remains fragile: out-of-distribution regimes trigger unstable behaviour, and the literature increasingly warns against backtest overfitting, advocating robust parameter plateaus over isolated mathematical peaks \citep{bailey2014, lopez2018advances, lin2024optimal}.

To mechanically mitigate such instability, the machine learning literature has formalized Safe Reinforcement Learning via Shielding, where a deterministic, non-learned layer enforces hard operational guardrails over a stochastic policy, while strict temporal constraints during training stabilize value estimation \citep{garcia2015comprehensive, alshiekh2018safe, pardo2018time, dulac2019challenges}. Although well documented in robotics, the integration of shielding with classical quantitative risk management to constrain DRL trading agents in high-frequency, friction-heavy markets remains underexplored. Contemporary end-to-end models that attempt to learn feature extraction, cointegration detection, sizing, and execution jointly suffer from high sample complexity and out-of-sample brittleness, particularly when underlying market regimes shift \citep{guijarro2021, zeng2025regimefolio}. Taken together, the current literature suggests that neither classical static frameworks nor unconstrained DRL alone are sufficient to extract robust alpha in cryptocurrency pair trading: the former lack adaptivity, the latter lack safety. This precisely motivates our contribution, namely a decoupled hybrid architecture in which cointegration, Hurst-based structural filters, and dynamically anchored statistical boundaries act as a deterministic shield over a risk-sensitive PPO-LSTM execution policy, explicitly designed to address this gap in the highly volatile cryptocurrency domain.\\

\section{Methodology}

\subsection{Technical Implementation}

Empirical analysis was conducted using a custom end-to-end framework engineered exclusively for this research. To ensure complete methodological transparency, the entire simulation engine was developed from the ground up, leveraging core libraries directly rather than relying on existing backtesting frameworks. This custom architecture allows an unconstrained and precise realization of the proposed execution mechanics. The complete source code is available in the public GitHub repository\footnote{The repository can be accessed at: \url{https://github.com/damianlebiedz/pair-trading-with-rl}}.

The system is built on Python 3.12. This setup allows users to run the framework seamlessly within isolated Docker containers, guaranteeing a consistent runtime environment regardless of the host machine. The framework's architecture is highly modular, orchestrated through dedicated execution scripts. The Reinforcement Learning module is built in strict compliance with the Gymnasium API, ensuring standardized and scalable agent-environment interactions. For the algorithmic backbone, the framework leverages Stable-Baselines3 along with its extension, SB3-Contrib. These libraries provide reliable, highly optimized implementations of state-of-the-art DRL algorithms, such as Recurrent PPO used throughout the experiments. Furthermore, the DRL training process is fully integrated with Weights \& Biases (W\&B).

Readers and researchers interested in replicating the study, or running the framework with their own custom parameters, are referred to the detailed tutorial provided in the repository's README.md file.

Consequently, unless otherwise stated, all figures and tables presented in this study are the author’s own calculations.

\subsection{Data}

This study utilizes historical market data sourced from the official \textit{Binance Data Vision} archive for the USD-M Futures segment. Raw historical dumps were downloaded directly from the exchange's AWS S3 Bucket archives\footnote{Binance Data Vision, \url{https://data.binance.vision/?prefix=data/futures/um/monthly/klines} (accessed: March 16, 2026).}. The analysis presented in this study operates on a 1-hour (1h) timeframe.

The dataset is partitioned into an In-Sample (IS) training period, covering the year 2024, and an Out-Of-Sample (OOS) testing period spanning 2025. To facilitate the dynamic pair selection methodology, the data retrieval process for the initial trading iteration (January 2024) begins on November 1, 2023. Consequently, the contiguous dataset used for this study spans from November 1, 2023, 00:00 UTC, to December 31, 2025, 23:59 UTC. The final data point corresponds to the hourly OHLCV candle opening at 23:00 UTC on the last day of the OOS period, ensuring a complete and uninterrupted observation of the market dynamics throughout the investigated timeframe.\\

\noindent \textbf{Universe Formation and Liquidity Constraints}

The key feature of the data framework is its dynamic, monthly-revised structure. Rather than utilizing a static basket of assets, the study operates on a sequence of rolling windows where a unique universe is constructed for each trading month. This iterative approach ensures that the model works consistently in the most relevant market segments.

For each iteration, the candidate pool is filtered based on the preceding 2-month formation period. During this window, the average daily quoted volume (USDT) is analyzed for all available instruments, allowing the assets to be sorted by liquidity. Crucially, assets that exhibit data gaps during this historical period are strictly excluded \textit{prior} to ranking. This completeness requirement is applied strictly \textit{ex-ante}, utilizing only information available at the time of selection, thereby preventing look-ahead bias. Consequently, for every period, a candidate pool of most liquid assets (by default $N=100$) is advanced to the pair selection pipeline. This liquidity criterion ensures adequate market depth, which is essential for minimizing execution slippage and mitigate microstructural frictions such as bid-ask bounce \cite{Gatev2006, Do2010}.\\

\noindent \textbf{Sample Size and Statistical Power Justification}

While classical pair trading literature frequently evaluates strategies over multi-decade horizons using daily observations (e.g., Gatev et al., 2006), this study deliberately focuses on a high-resolution one-year Out-Of-Sample period. In the context of micro-structural statistical arbitrage, the 1-hour interval yields exactly 8,760 data points per asset in Out-Of-Sample 2025. More importantly, the dynamic selection of 20 pairs evaluated over 12 rolling months translates to 240 pair-month execution episodes within the OOS period alone. This cross-sectional breadth and high data granularity allow a detailed assessment of short-term mean-reverting anomalies. However, since these episodes are not statistically independent, they complement rather than replace validation across multiple market cycles, as discussed among the limitations in Section \ref{sec:conclusion}.

\subsection{Hierarchical Evaluation}
\label{subsec:hierarchical_eval}

The backtesting framework is structured around a strict four-tier hierarchy. This design allows for both granular performance attribution and overall performance simulation.

\begin{itemize}
\item \textbf{Level 1. Single-Pair Monthly Backtest (Micro-Level):} The fundamental building block of the simulation. It represents the isolated execution of the strategy on one specific pair over a discrete one-month trading window. Crucially, at this micro-level, a strict \textit{trade-to-trade compounding} mechanism is applied. The capital base is updated immediately upon the closure of each position, ensuring that the asset quantity for any subsequent trade is determined strictly at its entry and remains frozen throughout the trade's lifespan.
\item \textbf{Level 2. Multi-Pair Monthly Portfolio:} For any given month, the independent results of selected pairs ($n = 20$ by default) are aggregated. This level evaluates the cross-sectional diversification based on a pair selection procedure.
\item \textbf{Level 3. Multi-Period, Multi-Pair Aggregation:} The monthly multi-pair portfolios are sequentially chained across 12 rolling iterations, representing a full year. At this level, a \textit{sequential compounding} mechanism is applied. The cumulative net profit or loss realized at the end of a preceding month directly adjusts the initial capital base allocated for the subsequent month's operations. This accurately reflects real-world portfolio growth and ensures methodological consistency when evaluating the strategy's exponential capital accumulation over time.
\item \textbf{Level 4. Full In-Sample/Out-Of-Sample Analysis (Macro-Level):} The fully merged 12-month multi-pair results for 2024 constitute the complete In-Sample (IS) performance. Correspondingly, the chained 12-month results for 2025 form the Out-Of-Sample (OOS) verification, enabling a comprehensive assessment of the model's capacity to generalize across completely unseen annual market regimes.
\end{itemize}

\begin{figure}[H]
    \caption{Dynamic Rolling Window Hierarchy: Level 1.}
    \label{fig:diagram}
    \centering
    \includegraphics[width=\linewidth]{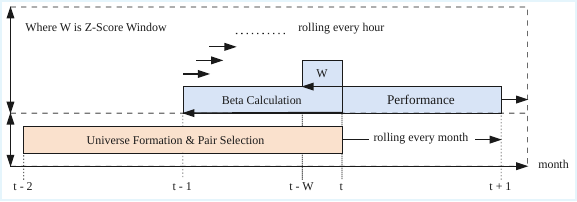}
    \justifying
    \noindent \justifying \noindent \scriptsize Note: 
    The diagram illustrates the strictly chronological data flow inside Level 1, designed to eliminate look-ahead bias. The 2-month Formation Window ($t-2, t-1$) is completely isolated from the 1-month Trading Window ($t$).
\end{figure}

\subsection{Market Frictions and Leverage Mechanics}
\label{subsec:market_fr_and_lev}

\noindent \textbf{Transaction Fees}

In realistic quantitative backtesting, explicitly accounting for market frictions is essential to evaluate the true viability of a strategy. By default, this study applies a 0.05\% transaction fee per execution, aligning with the standard USDT taker fee rate for regular users on the Binance for USD-M Futures\footnote{Binance Futures Fee Structure, \url{https://www.binance.com/en/fee/futureFee} (accessed: March 16, 2026).}. However, to test the strategy's sensitivity to unmodeled frictions, the assumptions verification phase includes a 0.10\% friction rate. This elevated rate serves as a composite proxy, encapsulating both the baseline 0.05\% commission and an additional 0.05\% penalty designed to simulate execution slippage in highly volatile cryptocurrency regimes. This is a simplified proxy rather than a realistic execution model, as funding payments, spreads, market impact, partial fills and latency are not simulated.\\

\noindent \textbf{Leverage Mechanics and Liquidation Threshold}

To facilitate a fair comparison of risk-adjusted returns, the strategy's capital exposure is scaled to align its structural risk profile with the higher-volatility benchmarks \cite{asness2012}.

As established in Subsection \ref{subsec:hierarchical_eval}, the strategy operates on a strict \textit{trade-to-trade compounding} basis. To preserve methodological consistency, the leveraged variant natively adopts exactly the same mechanism.

Formally, for a given leverage factor $L$, the strategy allocates a specific base margin upon generating an entry signal. This margin is multiplied by $L$ to establish the position's \textit{Notional Value}. Crucially, both the step-wise profit and loss (PnL) and the transaction fees are explicitly derived from this amplified nominal exposure. Throughout the duration of the trade, the absolute quantity of engaged leveraged assets remains fixed. This ensures that the leveraged strategy is governed by the identical underlying logic as the unleveraged baseline, with performance deviations driven solely by the scaled exposure and proportionally magnified market frictions.

Furthermore, simulating a highly leveraged margin account requires the incorporation of a strict liquidation engine to maintain mathematical validity and market realism. In this framework, a "Net PnL Liquidation Threshold" is enforced. A position is force-closed (margin call) at the exact moment the unrealized net PnL (inclusive of expected exit transaction fees) reaches $100\%$ of the initially allocated margin. Additionally, if the pair-specific equity is fully exhausted, the strategy is declared bankrupt, permanently halting trading operations for that particular asset pair. This accurately captures the absolute "Risk of Ruin" inherent in leveraged cryptocurrency markets \cite{thorp2011}. While real-world exchanges implement a Maintenance Margin requirement that triggers liquidation slightly before the full exhaustion of the margin balance, this study adopts a 100\% threshold as a simplification to ensure mathematical tractability and focus on the fundamental risk of ruin.

\subsection{Pair Selection}

To construct a robust bucket of assets, we implement a two-stage "Filter-then-Rank" (Composite Scoring with Hard Penalty) methodology, relying on a multi-stage selection instead of a single statistical metric \cite{Do2010, Krauss2017}. Our architecture intentionally separates the evaluation of a pair's statistical quality from its structural usefulness, ensuring that only pairs exhibiting both strong equilibrium and active mean-reverting properties are selected.

The pair selection process is executed iteratively for each trading month. For a given period (e.g., January 2024), the algorithm utilizes the preceding 2-month formation window (November–December 2023) and performs selection on the universe of $N=100$ most liquid assets based on their average daily volume during these two months. Subsequently, the \textit{Pair Selection} pipeline evaluates all potential combinations within this specific candidate pool.

Before any statistical evaluation, all raw asset close prices within the 2-month formation window are transformed into natural logarithms. This transformation will ensure that the modeled spreads represent symmetric percentage differentials.\\

\noindent \textbf{Stage 1: Statistical Scoring (Quality Assessment)}

In the first stage, we evaluate the fundamental relationship between two assets to determine their long-term equilibrium and short-term linear dependence.
For every potential pair combination formed from the $N=100$ candidate pool, we compute a \textit{Raw Score} defined as a uniformly weighted sum of its normalized cointegration strength and its coefficient of determination:
\begin{equation}
    \text{Raw Score} = 0.5 \cdot (1 - p_{EG}) + 0.5 \cdot R^2
\end{equation}
where $p_{EG}$ is the p-value derived from the Augmented Engle-Granger two-step method \cite{Engle1987}, and $R^2$ is the squared Pearson correlation coefficient. While the cointegration ensures that the spread does not wander indefinitely over the long term, the $R^2$ metric complements this by capturing the short-term variance explained by the linear fit. This composite metric allows us to rank pairs from the most to the least statistically robust.

The equal weighting (50/50) of cointegration strength and the coefficient of determination ($R^2$) serves as a balanced heuristic designed to identify pairs that exhibit both a long-term equilibrium and short-term signal clarity. Crucially, rather than applying an arbitrary absolute acceptance threshold (e.g., Raw Score $> 0.70$), the methodology enforces a strict cardinality constraint by dynamically selecting only the top $n$ pairs (with a default of $n=20$) out of 4,950 potential combinations. This mechanism strictly isolates the extreme right tail of the distribution (e.g., the top 0.4\% for the default parameter). Because the Composite Score is a convex combination of two metrics strictly bounded within the [0, 1] interval, the ranking favors combinations with strong composite scores. A pair exhibiting a strong correlation but weak cointegration (or vice versa) may be outranked by combinations excelling in both. The top $n$ ranking may therefore reduce the sensitivity to the exact 50/50 weighting; however, the scoring weights and the Hurst threshold were not systematically sensitivity-tested in this study.\\

\noindent \textbf{Stage 2: Structural Filtering (Suitability Validation)}

While a high \textit{Raw Score} indicates a strong historical relationship, it does not guarantee suitability for a mean-reversion strategy. A highly correlated pair might be in a diverging trend, or its relationship might be inverted. To address this, we apply strict structural filters using the Hurst exponent ($H$) \cite{Hurst1951} and the hedge ratio ($\beta$).

Crucially, both $H$ and $\beta$ are estimated over the exact same 2-month formation window used for Stage 1. The hedge ratio $\beta$ is derived with Ordinary Least Squares (OLS) regression ($p_{A,t} = \alpha + \beta p_{B,t} + \epsilon_t$). Rather than blending these metrics into the continuous Composite Score - which risks allowing an exceptionally high correlation to mask disqualifying structural flaws, we treat them as hard constraints:
\begin{itemize}
    \item \textbf{Mean-Reversion Constraint ($H < 0.5$):} The Hurst exponent quantifies the long-term memory of a time series. A value of $H = 0.5$ signifies a pure random walk. A value of $H > 0.5$ indicates persistent (trending) behavior. Therefore, only spreads exhibiting anti-persistence ($H < 0.5$) possess the mean-reverting properties required to validate the core premise of statistical arbitrage \cite{Qian2007, ramos2017hurst}.
    \item \textbf{Hedge Feasibility Constraint ($\beta > 0$):} A non-positive beta indicates an inverse or non-existent linear relationship, making standard long/short hedge ratio construction inappropriate.
\end{itemize}

\vspace{0.5em}
\noindent \textbf{Final Composite Scoring Model}

\begin{equation}
\text{Final Score} = 
\begin{cases} 
0.5 \cdot (1 - p_{EG}) + 0.5 \cdot R^2 & \text{if } H < 0.5 \text{ and } \beta > 0 \\ 
0.0 & \text{if } H >= 0.5 \text{ or } \beta \le 0 
\end{cases}
\end{equation}

\vspace{0.5em}
Following this evaluation, the pairs are ranked in descending order, based on their Final Score, and the top n (where $n=20$ by default in this analysis), pairs are selected for the subsequent trading month. For this final basket, the total capital allocated for the given trading month is distributed on a strict equal-weight basis, assigning exactly 1/n of the available funds to each pair.\\

\noindent \textbf{Methodological Considerations in Cointegration Testing}

While the Johansen cointegration test \cite{johansen1988} is frequently highlighted in econometric literature for its theoretical superiority, specifically its symmetry (independence from the choice of the dependent variable) and higher statistical power in identifying cointegrating vectors, the Engle-Granger (EG) two-step approach \cite{Engle1987} was deliberately selected for this framework due to specific architectural and computational requirements.

First, the framework operates on a dynamic rolling-window basis evaluating $N=100$ assets, resulting in $N(N-1)/2 = 4950$ unique pair combinations tested in a single trading month. Implementing the Johansen test, which requires complex eigenvalue decomposition for each combination iteratively across the dataset, introduces severe computational bottlenecks ($O(N^2)$ complexity per rolling step). The EG test provides a computationally tractable heuristic that aligns with the requirements of high-frequency backtesting engines.

Second, the primary limitation of the EG test, its sensitivity to variable ordering which can lead to asymmetric cointegration estimations (e.g., asset A cointegrates with asset B, but not vice versa, potentially distorting the hedge ratio), is explicitly acknowledged and mitigated within the broader pair selection pipeline. The framework does not rely solely on the EG test. As detailed in Stage 2, any structural flaws or spurious asymmetries passed by the EG test are heavily filtered out by strict deterministic constraints, namely the Hurst exponent requirement ($H < 0.5$) and the Hedge Feasibility constraint ($\beta > 0$).

Ultimately, the EG test serves strictly as an initial, fast mathematical sieve in Stage 1. By coupling it with robust secondary filters and focusing strictly on bivariate relationships (where Johansen's multivariate advantages are less critical), the framework successfully isolates high-quality mean-reverting pairs while maintaining computational efficiency.\\

\noindent \textbf{Survivorship Bias and Look-ahead Mitigation}

The selection pipeline is designed to be strictly immune to survivorship bias. If an asset was available and met the liquidity criteria during the 2-month formation window, it remains a valid candidate for the subsequent trading month, even if it was delisted during that active testing period. The trading engine recognizes this discontinuity and simulates an immediate emergency liquidation of the open position at the last available quoted price, fully realizing and booking any incurred losses.

\subsection{Baseline Strategy Framework}

The framework employs a strict \textit{Signal on Close, Execute on Open} architecture to eliminate look-ahead bias. To mirror real-world execution, raw hourly candles (indexed by open time) are re-indexed to their exact \textit{close time}. For example, a candle covering price action from \texttt{00:00:00} to \texttt{00:59:59} is timestamped at \texttt{01:00:00}. Consequently, for a trading month starting at \texttt{00:00:00}, the first actionable data point is the fully formed candle at \texttt{01:00:00}.

At any discrete step $t$, statistical indicators are calculated exclusively using the finalized close price from $t-1$. If a signal is generated, the simulated order is executed at the \textit{open price} of the candle beginning at $t$. This structural separation between the signal-triggering price (historical close) and the fill price (subsequent open) ensures that the model acts only on information available at that exact moment. 

Crucially, because the \textit{Pair Selection} pipeline is performed monthly, a strict \textit{terminal liquidation} policy is enforced at the end of each testing window. At the final timestamp of the period (e.g., 2024-02-01 00:00:00), any open trades are forcefully closed at the last available close price. This mandatory liquidation ensures that PnL is fully realized within the correct period and that the subsequent month begins with a neutral cash position, ready for the newly reconstituted set of pairs.

\subsubsection{The Fixed Risk, Adaptive Mean Model}

The core execution logic of the strategy deviates from the classical, strictly stationary pair trading frameworks (e.g., Gatev et al., 2006) to better accommodate the persistent micro-structural drift characteristic of cryptocurrency markets. Rather than executing a continuous rolling mean-reversion, the proposed architecture trades a \textit{snapshot equilibrium}. We conceptualize this approach as the \textit{Fixed Risk, Adaptive Mean} model. 

The system operates in a discrete action space, where the portfolio exposure is either long leg ($1.0$), short leg ($-1.0$), or neutral/out-of-market ($0.0$) \cite{yang2024}. When a signal is generated, the strategy commits 100\% of the available capital allocated to that specific pair.

To ensure scale-invariance, all price data are transformed into natural logarithms. Let $p_{A,t}$ and $p_{B,t}$ denote the log-prices at the current time step $t$, and let $t_0$ represent the timestamp of position entry. The decision-making proxy, the Conditional Z-Score ($Z_t$), is derived based on the operational state of the portfolio:

\begin{equation}
Z_t =
\begin{cases}
\frac{(p_{A,t} - \beta_t p_{B,t}) - \mu_t}{\sigma_t} & \text{if } \text{Position}_{t-1} = 0 \text{ (Market State)} \\
\frac{(p_{A,t} - \beta_{t_0} p_{B,t}) - \mu_t}{\sigma_{t_0}} & \text{if } \text{Position}_{t-1} \neq 0 \text{ (Position State)}
\end{cases}
\end{equation}

\vspace{0.5em}
A critical feature of this equation is the decoupling of the estimation horizons. While the hedge ratio ($\beta$) is estimated over a long-term, one-month rolling window to capture the stable, wider relationship between the pair, the Z-Score components ($\mu$ and $\sigma$) are evaluated over a much shorter, one-week rolling window (\textit{Z-Score Window = 168} hours by default in this analysis). This dual-window approach allows the strategy to anchor its physical exposure to a robust, long-term ratio, while remaining agile enough to exploit short-term, local micro-structural price anomalies.

In the \textit{Market State}, the strategy uses rolling beta ($\beta_t$) and standard deviation ($\sigma_t$) to evaluate new opportunities against current market conditions. Once a position is initiated, the strategy transitions to the \textit{Position State}, where the hedge ratio ($\beta_{t_0}$) and volatility anchor ($\sigma_{t_0}$) are frozen. However, the mean ($\mu_t$) remains adaptive (live) in both states to account for structural price drift.

This behavior is governed by three foundational pillars:\\

\noindent \textbf{State-Dependent Hedge Ratio ($\beta_{t_0}$)}

In the classical pair trading literature, strategies often rely on a strict 1:1 dollar-neutral capital allocation ($\beta = 1.0$, $w_A = 0.5, w_B = 0.5$) \cite{Gatev2006, vidyamurthy2004pairs}. However, this equal-weighting approach carries a significant structural drawback. If one asset exhibits significantly higher idiosyncratic volatility than its paired counterpart, a rigid 1:1 capital split leaves the portfolio heavily exposed to unhedged directional market risk.

To address this limitation, the proposed strategy utilizes a dynamic hedging mechanism operating across two distinct states:

\begin{itemize}
\item \textbf{Market State (Out-of-Position):} While searching for entry opportunities, $\beta_t$ is continuously re-calculated at each discrete time step $t$ using an OLS regression over a rolling one-month lookback window. This window spans from the first close price of the preceding month to the current close price at $t$. The system is strictly immune to look-ahead bias due to the \textit{Signal on Close, Execute on Open} architecture, ensuring that signals derived from finalized data are executed only at the subsequent market \texttt{open}.
During this state, the spread is evaluated as:
\begin{equation}
\text{Spread}_t = p_{A,t} - \beta_{t} p_{B,t}
\end{equation}
\item \textbf{Position State (In-Position):} Upon generating an entry signal at the close of a candle $t_0$, the optimal hedge ratio $\beta_{t_0}$ (derived from the one-month window ending at $t_0$) is immediately locked for the entire duration of the trade. During this state, the spread for all $t > t_0$ is evaluated as:
\begin{equation}
\text{Spread}_t = p_{A,t} - \beta_{t_0} p_{B,t}
\end{equation}
This ensures that the strategy tracks the performance of the exact physical inventory held, maintaining the mathematical integrity of the trade against subsequent shifts in the rolling market correlation. Capital weights are established strictly at $t_0$ using $w_A = 1 / (1 + \beta_{t_0})$ and $w_B = \beta_{t_0} / (1 + \beta_{t_0})$.
\end{itemize}

Immediately upon position liquidation, the system reverts to the Market State, resuming the calculation of the spread and Z-Score using the rolling market beta ($\beta_t$) to identify the next mean-reversion opportunity.

This structural freezing is a practical necessity. Continuously rebalancing to match a rolling beta hedge ratio during an active trade would generate excessive portfolio turnover, entirely eroding the statistical edge through cumulative transaction commissions \cite{avellaneda2010}.\\

\noindent \textbf{State-Dependent Standard Deviation ($\sigma_{t_0}$, \textit{Volatility Anchor})}

During an active position, the standard deviation of the spread - acting as the denominator in the Z-Score equation - is strictly frozen at its entry value ($\sigma_{t_0}$). This architectural decision prevents the phenomenon of "denominator error" or false Mean-Reversion Signals. In the highly volatile markets, a sudden exogenous volatility shock will cause the rolling standard deviation to spike. If the denominator is left unfrozen, it mathematically dilutes the Z-Score, forcing it artificially toward zero. Freezing $\sigma_{t_0}$ at entry solidifies the initial unit of risk/reward, defending the mathematical integrity of the exit logic against purely noisy volatility expansions.

\begin{itemize}
\item \textbf{Market State (Out-of-Position):} 
The strategy utilizes a rolling standard deviation ($\sigma_t$), calculated over the specified Z-Score Window (168 hours by default).
\item \textbf{Position State (In-Position):} 
Upon position initiation, the standard deviation is strictly locked at its entry value ($\sigma_{t_0}$). The Z-score is then calculated using this constant value.
\end{itemize}

\noindent \textbf{Live Mean ($\mu_t$)}

Conversely, the rolling mean of the spread ($\mu_t$) is continuously updated based on the current Z-Score Window:
\begin{equation}
    \mu_t = \frac{1}{W} \sum_{k=0}^{W-1} \text{Spread}_{t-k}
\end{equation}
If the mean is frozen at entry and the market undergoes a permanent regime shift, the strategy risks holding a stranded asset indefinitely, waiting for a historical equilibrium that no longer exists. A live mean allows the strategy to slowly adapt to the new equilibrium.

\subsubsection{Stop Loss Architecture}

To protect the portfolio from structural breakdowns and prolonged regime shifts, the framework implements a robust, three-tier risk management architecture. The system relies on predefined mathematical boundaries rather than arbitrary percentage drops, ensuring that risk scales proportionally to the statistical properties of the spread.\\

\noindent \textbf{Deterministic Stop Loss Threshold}

The primary Stop Loss ($\text{SL}$) parameter is defined mathematically as a static multiplier of the predefined Entry Threshold ($\text{Entry}_{thr}$), rather than a multiplier of the actual execution Z-Score ($Z_{t_0}$):
\begin{equation}
    \text{SL}_{thr} = \text{Entry}_{thr} \times \text{SL}
\end{equation}

This deterministic approach is crucial when operating on discrete, hourly intervals. In highly volatile markets, an hourly candle might close with an extreme Z-Score spike (e.g., 5.50). If the strategy calculated the SL dynamically based on the execution Z-Score, the risk boundary would widen disproportionately (to 11.0), exposing the portfolio to unpredictable drawdowns. By fixing the SL Threshold to the Entry parameter, the maximum allowed deviation remains constant. A late entry means the position has less room to maneuver before hitting the SL, penalizing the imperfect execution latency without compromising the global risk limits.\\

\noindent \textbf{SL Lock (\textit{Regime Filter})}

To prevent the strategy from repeatedly taking losses, the model employs an \textit{SL Lock} mechanism. If a position is liquidated via Stop Loss, the specific pair is immediately blacklisted from opening new trades. This lock is lifted only when the spread definitively proves it has returned to an equilibrium state - when the Z-Score crosses the Exit Threshold boundary (e.g., returning to the mean, if Exit Threshold = 0.0). This regime filter effectively forces the strategy to stay out of trending, non-stationary market phases.\\

\noindent \textbf{Time Decay Stop Loss} 

To prevent capital from being tied up in stagnant, non-reverting spreads, the strategy introduces a temporal feature known as \textit{Time Decay SL}. By default, this mechanism enforces a strict time limit on active trades based on the Z-Score Window.

If a trade remains open for an extended period, the original SL Threshold begins to linearly contract towards the Exit Threshold. In this study, the decay initiates once the trade duration reaches 50\% of the Z-Score Window and steadily tightens at each calculation step. By the time the trade duration equals the full Z-Score Window length, the SL Threshold converges completely with the Exit Threshold, forcing a mandatory liquidation. This temporal tightening mechanism penalizes prolonged market exposure.

\subsubsection{Signal Generation and Execution Mechanics}

The chronological flow of a single algorithmic step is defined as follows:

\begin{enumerate}
    \item \textbf{Calculation ($t$):} At the close of the hourly candle at time $t$, all rolling statistics and the conditional Z-Score ($Z_t$) are computed using definitive historical data.
    
    \item \textbf{Decision ($t$):} Initially, the algorithm verifies the SL Lock status. If the pair is currently locked due to a recent stop-loss liquidation, all entry signals are suppressed until $Z_t$ crosses the $\text{Exit}_{thr}$ boundary, proving a return to equilibrium. If the pair is tradable, the system evaluates $Z_t$ and its predecessor ($Z_{t-1}$) against the thresholds ($\text{Entry}_{thr}$, $\text{Exit}_{thr}$) and the defined risk boundary ($\text{SL}_{thr}$):
    
    \begin{itemize}
        \item \textbf{Long Entry:} Triggered if the Z-Score crosses below the negative entry threshold from above ($Z_{t-1} > -\text{Entry}_{thr} \ge Z_t$). The signal is only valid if the divergence has not breached the Stop Loss boundary ($Z_t > -{SL}_{thr}$).
        \item \textbf{Short Entry:} Triggered if the Z-Score crosses above the positive entry threshold from below ($Z_{t-1} < \text{Entry}_{thr} \le Z_t$). Similarly, the signal is valid only if $Z_t < {SL}_{thr}$.
        \item \textbf{Standard Exit (Take Profit):} If a position is active, a liquidation signal is triggered when the Z-Score reverts to the equilibrium mean, defined by the Exit Threshold (crossing $0.0$ by default). For a Long position, this occurs when $Z_t \ge -\text{Exit}_{thr}$; for a Short position, when $Z_t \le \text{Exit}_{thr}$.
        \item \textbf{Stop Loss Exit (Risk Liquidation):} A mandatory liquidation is triggered if the divergence expands beyond the predefined risk boundary. For a Long position, this occurs when $Z_t \le - \text{SL}_{thr}$; for a Short position, when $Z_t \ge \text{SL}_{thr}$.
    \end{itemize}
    
    \item \textbf{Execution ($t+1$):} If an entry or exit signal is generated at the close of $t$, the simulated order is routed and executed at the open price of the subsequent candle. This structural separation ensures the model acts only on fully finalized information, reflecting the practical constraints of computational processing.
\end{enumerate}

\subsection{Reinforcement Learning Implementation}

The integration of Reinforcement Learning (RL) in this study departs from conventional end-to-end trading architectures, where agents are expected to derive both signals and execution logic from raw market data \cite{jiang2017cryptocurrency}. Instead, we implement the RL agent as a \textit{Dynamic Execution Overlay}.

\subsubsection{Algorithm}

To handle temporal dependencies, we utilize Proximal Policy Optimization (PPO) \cite{schulman2017ppo} with a Long Short-Term Memory (LSTM) architecture \cite{hochreiter1997}. The recurrent layer allows the agent to maintain an internal "memory" of recent market dynamics, which is crucial for identifying shifting regimes.

The key hyperparameters of the PPO algorithm and the recurrent network architecture are summarized in Table \ref{tab:rl_hyperparameters}.

\begin{table}[H]
    \centering
    \footnotesize
    \renewcommand{\arraystretch}{1.2}
    \caption{PPO and LSTM Network Hyperparameters.}
    \label{tab:rl_hyperparameters}
    \vspace{12pt}
    \begin{tabularx}{\linewidth}{l >{\raggedright\arraybackslash}X >{\centering\arraybackslash}X}
    \toprule
        \textbf{Category} & \textbf{Parameter} & \textbf{Value} \\
    \midrule
        \textbf{Optimization} & Learning Rate & 0.0003 \\
         & Batch Size & 256 \\
         & Optimization Epochs & 10 \\
         & Clipping Range ($\epsilon$) & 0.2 \\[4pt]
         
        \textbf{RL Dynamics} & Gamma ($\gamma$) & 0.999 \\
         & Entropy Coefficient ($c_2$) & 0.01 \\
         & Number of Steps ($n_{steps}$) & 256 \\[4pt]
         
        \textbf{Architecture} & LSTM Hidden Size & 128 \\
         & Number of LSTM Layers & 1 \\
         & Shared LSTM & False \\
         & Enable Critic LSTM & True \\[4pt]
         
        \textbf{Training} & Passes per Pair & 20 \\
         & Seed & 42 \\
    \bottomrule
    \end{tabularx}
    
    \vspace{12pt}
    
    \justifying \noindent \scriptsize Note: The choice of a high discount factor ($\gamma = 0.999$) reflects the objective of prioritizing long-term portfolio stability over immediate localized gains. Furthermore, the inclusion of an \textit{LSTM-based Critic} is a deliberate architectural decision; it enables the value network to more accurately estimate the expected reward by considering the historical context of the spread's trajectory. Finally, the \textit{Passes per Pair} parameter dictates the number of training iterations executed over a single, one-month episode provided in the training dataset.
\end{table}

By setting this \textit{Passes per Pair} multiplier to 20, the agent iterates through the entire historical training set multiple times, effectively defining the total number of training epochs required for policy stabilization. This approach ensures that the model is sufficiently exposed to the underlying market dynamics to achieve convergence. Crucially, this repetition does not introduce data leakage despite the use of a recurrent architecture (LSTM). Since the implementation resets the LSTM hidden states at the conclusion of each 168-step episode, the agent is prevented from memorizing global chronological sequences and is instead forced to learn generalizable features within the local context of each pair.

Regarding the discount factor, $\gamma = 0.999$ was selected to align with the strictly finite horizon of the environment. In a 168-hour episode, this value ensures that rewards at the end of the window still retain approximately 84.6\% of their original weight ($\approx 0.999^{168}$). Such a high discount factor is essential for pair trading strategies to prevent myopic (short-sighted) behavior, as mean reversion often requires holding positions for a significant portion of the available time. While values very close to 1.0 can cause instability in infinite-horizon problems, the fixed duration of these episodes naturally bounds the optimization process, ensuring robust learning without the need for lower $\gamma$ values.

\subsubsection{Observation Space}
\label{subsubsec:obs_space}

To address the "curse of dimensionality" and identify the most impactful features for policy convergence, we designed a modular observation space:

\begin{itemize}
    \item \textbf{Autonomous:} Focuses on fundamental trade mechanics. The state vector $\mathbf{s}_t \in \mathbb{R}^3$ includes the current Z-Score, the agent's active position status, and the normalized time spent in the position ($t_{pos} / W$, where $W$ is \textit{Z-Score Window}).
    \item \textbf{Standard:} Adds the baseline's heuristic guidance. The state vector $\mathbf{s}_t \in \mathbb{R}^4$ extends the Autonomous set by including the discrete signal ($1, 0, -1$) generated by the baseline statistical model.
    \item \textbf{Full:} The state vector $\mathbf{s}_t \in \mathbb{R}^5$ expands the Standard set by incorporating the Hurst exponent ($H_t$) \cite{Hurst1951} computed over the identical lookback window as $\beta_t$.
\end{itemize}

To ensure stable policy convergence and mitigate the risk of vanishing or exploding gradients during training, the heterogeneous features of the selected observation space undergo dynamic normalization. Utilizing a running mean and variance algorithm \cite{engstrom2020}, the state space is continuously standardized to a zero mean and unit variance, with extreme market outliers systematically clipped. This critical preprocessing step guarantees that the neural network receives symmetrically scaled inputs, effectively resolving the inherent scale disparities between different features before the state representation is fed into the agent's policy network.

\vspace{0.5em}
\subsubsection{Reward Function}
\label{subsubsec:rew_func}

The reward function is the core component that aligns the agent's policy with risk-adjusted profitability. We implement three distinct schemes, all incorporating an asymmetric loss aversion coefficient $\lambda$ inspired by \textit{Prospect Theory} \cite{kahneman1979} and formalized technically within the Risk-Sensitive Reinforcement Learning framework \cite{mihatsch2002}. This ensures that negative returns are penalized more severely than positive gains are rewarded. In all cases, a portfolio bankruptcy event results in a terminal penalty of -1.0.\\

\noindent \textbf{1. Step-based Asymmetric Reward (\textit{StepPnL})}

This approach provides a dense reward signal. Its formulation evaluates the net return at each discrete time step $t$. This forces the agent to internalize the continuous cost of capital, market fluctuations, and execution fees ($f_t$):
\begin{equation}
\label{eq:rew}
    r_t = \frac{\Delta PnL_t - f_t}{\text{Equity}_t}, \quad 
    R_t = \begin{cases} r_t & \text{if } r_t \geq 0 \\ \lambda \cdot r_t & \text{if } r_t < 0 \end{cases}
\end{equation}
To evaluate the impact of downside risk penalization, two configurations for the multiplier were tested in this study: $\lambda = 1.0$ (representing a symmetric baseline) and $\lambda = 1.2$ (introducing an asymmetric penalty for negative returns).

To ensure numerical stability during the gradient update process, the reward function incorporates a terminal state logic. If $Equity_t \leq 0$, the environment triggers a bankruptcy event, and the agent receives a fixed penalty of -1.0, bypassing the division in Eq. \ref{eq:rew}. Furthermore, since the In-Sample training is conducted strictly on an unleveraged basis (1x leverage), the equity fluctuations remain within a stable range, preventing extreme reward scaling issues that could otherwise arise in highly leveraged scenarios.

Importantly, this approach ensures that leverage remains transparent to the agent's policy. Since the observation space excludes capital-related variables and focuses solely on statistical indicators and position status, the transition to a 10x leverage during testing does not change the agent's inputs. Even if higher leverage leads to a faster 'risk of ruin', the agent’s logic remains unaffected because it is incapable of perceiving the margin level, and once liquidated, it simply ceases to make further actions while outside the market.\\

\noindent \textbf{2. Trade-based Realized Reward (\textit{TradePnL})}

To mitigate the noise inherent in financial time series, this scheme provides a sparse reward signal only when a trade is finalized. This allows the agent to ignore temporary adverse excursions:
\begin{equation}
    r_{trade} = \frac{PnL_{trade}}{\text{Equity}_t}, \quad 
    R_t = \begin{cases} 
    r_{trade} & \text{if } \textit{trade\_ended} \text{ and } r_{trade} \geq 0 \\ 
    \lambda \cdot r_{trade} & \text{if } \textit{trade\_ended} \text{ and } r_{trade} < 0 \\ 
    0 & \text{otherwise} 
    \end{cases}
\end{equation}
Consistent with the \textit{StepPnL} approach, the identical multiplier configurations ($\lambda \in \{1.0, 1.2\}$) were applied to penalize finalized losing trades.\\

\noindent \textbf{3. Hybrid Signal-Action Reward (\textit{HybridAction})}

This formulation combines realized \textit{TradePnL} ($r_{trade}$) with an immediate reinforcement signal ($\Phi_t$) triggered during position entry \cite{yang2024}. Let $S_t \in \{-1, 0, 1\}$ denote the baseline mean-reverting signal and $A_t \in \{-1, 0, 1\}$ denote the agent's chosen action. The dynamic component $\Phi_t$ is defined relative to the total round-trip fee ($2f_t$) and a tuning multiplier $m$:
\begin{equation}
    \Phi_t = \begin{cases} 
    +2f_t \cdot m & \text{if } A_t = S_t \text{ and } S_t \neq 0 \quad \text{(Action Bonus)} \\ 
    -2f_t \cdot m & \text{if } A_t = 0 \text{ and } S_t \neq 0 \quad \text{(Omission Penalty)} \\ 
    -4f_t \cdot m & \text{if } A_t = -S_t \text{ and } S_t \neq 0 \quad \text{(Contrarian Penalty)} \\
    0 & \text{otherwise}
    \end{cases}
\end{equation}
In this study, we set the multiplier to $m = 0.2$, effectively making the signal-following bonus equal to 40\% of the transaction cost. The final reward for this hybrid scheme is the sum of the sparse trade reward and the dense action signal: 
\begin{equation}
    R_t = R_t^{trade} + \Phi_t
\end{equation}

It is crucial to explicitly state that the HybridActionReward scheme is designed as a deliberate test of the "guided exploration" paradigm, rather than a scenario of fully autonomous enhancement. The inclusion of the Omission Penalty (where the agent is penalized for remaining neutral when the baseline signal triggers, i.e., $A_t=0$ and $S_t \neq 0$) inherently hard-codes the baseline heuristic into the reward landscape. The architectural objective here is not to create an independent decision-maker, but to empirically investigate whether explicitly forcing the agent to mimic the validated statistical signals accelerates policy convergence, or conversely, if such circular logic constrains the neural network's capacity to optimize micro-execution timing. Acknowledging this internal constraint makes the subsequent Out-Of-Sample ablation study between fully autonomous rewards (e.g., StepPnL) and explicitly guided rewards interpretable. HybridActionReward is therefore only an evaluated experimental variant (Agents 13--18, see Table \ref{tab:agent} in Section \ref{sec:rl_eval}); the selected Agent 2 uses the StepPnLReward.

\subsubsection{Training Diagnostic Metrics}

To ensure transparency, all training sessions were monitored using the Weights \& Biases (W\&B) platform. Consequently, all the training diagnostic figures presented in this study represent the author's own elaboration of data captured during the agents' training and logged via W\&B.

To monitor the effectiveness of the training phase, four primary diagnostic metrics are utilized. These indicators provide insight into the optimization process of the Proximal Policy Optimization (PPO) algorithm \cite{schulman2017ppo}:\\

\noindent \textbf{1. Mean Episode Reward ($R_{mean}$)} \\

The episode reward represents the total sum of step-based rewards ($r_t$) accumulated by the agent over a discrete one-month trading window, which constitutes a single episode in this environment. The total reward for an episode of length $T$ is defined as:
\begin{equation}
    R_{episode} = \sum_{t=1}^{T} r_t
\end{equation}
To continuously assess training progress, the learning framework tracks the Mean Episode Reward ($R_{mean}$), calculated internally as a rolling average over up to 100 most recent episodes. An upward trajectory indicates that agents are successfully learning to better maximize the reward function over time.\\

\noindent \textbf{2. Value Loss ($\mathcal{L}^{VF}$)} \\

Value loss measures the mean squared error (MSE) between the state-value function estimated by the Critic network ($V_{\theta}(s_t)$) and the actual future discounted rewards observed during training ($V_t^{targ}$). It is defined as:
\begin{equation}
    \mathcal{L}^{VF} = \mathbb{E}_t \left[ (V_{\theta}(s_t) - V_t^{targ})^2 \right]
\end{equation}
A decreasing and stabilizing value loss indicates that the Critic is becoming increasingly accurate at estimating the expected cumulative reward for a given market state.\\

\noindent \textbf{3. Explained Variance ($EV$)} \\

The explained variance quantifies the proportion of the variance in the future discounted rewards that is successfully captured by the Critic's value function. It is a critical indicator of the signal-to-noise ratio within the agent's internal representation:
\begin{equation}
    EV = 1 - \frac{\text{Var}(V_t^{targ} - V_{\theta}(s_t))}{\text{Var}(V_t^{targ})}
\end{equation}
The dynamic of $EV$ depends on the characteristics of the reward function. A consistent upward trend indicates that the Critic is increasingly able to distinguish structural patterns from stochastic noise.\\

\noindent \textbf{4. Policy Entropy ($H$)} \\

Entropy measures the degree of randomness in the agent's action distribution ($\pi$). It serves as a primary mechanism to balance the fundamental exploration-exploitation trade-off \cite{sutton2018} during training, ensuring that the agent does not prematurely converge to a suboptimal policy ($A$ denotes the discrete action space):
\begin{equation}
    H(\pi(\cdot|s_t)) = -\sum_{a \in \mathcal{A}} \pi(a|s_t) \log \pi(a|s_t)
\end{equation}
The stabilization of entropy at a non-zero plateau ensures that the policy remains adaptive and continues to explore the state space, preventing the model from prematurely collapsing into a rigid and potentially \textit{overfitted} state.

\subsubsection{Agent Execution and Risk Management Overlay}
\label{subsubsec:agent_exec_risk_overlay}

Transitioning from the baseline heuristic model, the framework introduces a Deep Reinforcement Learning (DRL) agent to govern trade execution. Rather than relying on static $Z_t$ entry thresholds, the agent continuously evaluates the market state and autonomously outputs discrete actions (Long, Short, or Neutral/Out-of-Market).

However, DRL policies applied to financial time series are notoriously susceptible to out-of-distribution (OOD) market regimes. In unseen environments, unconstrained agents often exhibit unpredictable behaviors during structural breakdowns \cite{garcia2015comprehensive}. To ensure capital preservation and mitigate this inherent unpredictability, we implement a heuristic \textit{Risk Management Overlay} during the deployment phase. 

This architecture acts as a deterministic, non-learned shielding layer that enforces strict execution guardrails over the RL agent's stochastic actions  \cite{alshiekh2018safe}. While the agent maintains full autonomy to enter positions and can choose to close them early, the overlay enforces a hard boundary based on the validated baseline parameters: a static Take-Profit (Exit Threshold = 0.0) and a rigorous Stop Loss ($\text{SL}$). If the agent fails to liquidate a position before these boundaries are breached, the overlay overrides the neural network and forces a mandatory exit. In the evaluated configuration, the overlay also applies the SL Lock and the progressive Time Decay SL. The time-based close at the end of the Z-Score Window and the margin-exhaustion liquidation (Subsection \ref{subsec:market_fr_and_lev}) are separate execution constraints that remain active even when the overlay is disabled.

This dual-layer approach, combining a dynamic RL execution policy with a static risk overlay, serves to truncate unbounded losses during permanent regime changes. By isolating the RL agent's decision-making from the heuristic risk overlay, the system assigns each layer the task it is best suited for: Reinforcement Learning optimizes execution timing, while deterministic rules bound the downside that a learned policy cannot reliably control under out-of-distribution conditions. The ablation in Table \ref{tab:rl_oos_mechanism} supports this division of roles, as disabling the overlay increases the maximum drawdown from 35.27\% to 65.09\%. Unlike formally verified shields \cite{alshiekh2018safe}, these limits are heuristic rather than provably safe.

\subsubsection{Agent Training Environment and Autonomy}
\label{subsubsec:agent_training_autonomy}

During the In-Sample training phase, the heuristic Risk Management Overlay (detailed in Section \ref{subsubsec:agent_exec_risk_overlay}) is explicitly disabled. The Reinforcement Learning agent operates in a fully autonomous mode, granting the neural policy unrestricted control over the actions.

The training process is executed directly on the historical data structured by the baseline strategy's In-Sample (2024) backtest. By default, the environment simulates 12 monthly iterations across a portfolio of 20 dynamically selected pairs, yielding a total of 240 distinct, single-pair monthly episodes. The DRL algorithm iteratively traverses these isolated backtests, learning execution policies based on the exact market regimes.

This unconstrained environment is computationally essential for effective exploration \cite{dulac2019challenges}. Imposing deterministic boundaries during training would prematurely truncate episodes, artificially censoring the reward signal and biasing the value function estimation \cite{pardo2018time}. By forcing the agent to independently execute exit actions, the network learns to natively associate delayed liquidations with decaying portfolio value. The agent is trained at 1x leverage to prevent 'Policy Collapse', a state where extreme bankruptcy penalties during exploration could lead to an overly passive policy. Since the observation space contains no capital-, margin- or leverage-related variables (see Subsection \ref{subsubsec:rew_func}), the learned mapping from market states to actions is identical at 1x and 10x leverage; leverage scales the financial consequences of the actions, not the decisions themselves. Likewise, the overlay is applied only at deployment by design: as a non-learned shield, it overrides the policy only when the validated baseline boundaries are breached and does not alter the policy's inputs. In the Out-Of-Sample phase, the Agent and the Baseline operate under the same 10x leverage, fee assumptions and exit boundaries (Exit Threshold, Stop Loss, SL Lock and Time Decay SL), so the performance difference between them stems from the execution decisions of the learned policy.

However, to prevent infinite episode horizons caused by permanently diverging, non-stationary pairs, the training environment enforces a single structural constraint: Time Decay Stop (\textit{Temporal Truncation}). If a position reaches this maximum duration limit (Z-Score Window = 168 hours by default), the environment forces a mandatory liquidation at the current market price.

Crucially, while the other heuristic exit rules are disabled, the underlying mathematical architecture of the environment remains identical to the baseline model. The \textit{Fixed Risk, Adaptive Mean} framework is strictly enforced throughout the training phase to ensure consistency across environments.

\subsection{Performance Metrics}
\label{subsec:performance_metrics}

To comprehensively evaluate the performance, we employ a standard suite of quantitative metrics. Let $V_t$ denote a portfolio equity value at time $t$, and $R_t$ denote the periodic return.\\

\noindent \textbf{Compound Annual Growth Rate (CAGR)}

\begin{equation}
    \text{CAGR} = \left( \frac{V_T}{V_0} \right)^{\frac{1}{Y}} - 1
\end{equation}
where $V_T$ is the final portfolio value, $V_0$ is the initial allocated capital, and $Y$ represents the total duration of the testing period expressed in years. Because multi-period aggregation applies capital compounding, the CAGR directly reflects the true exponential growth rate of the simulated portfolio.\\

\noindent \textbf{Annualized Volatility ($\sigma_{ann}$)}

\begin{equation}
    \sigma_{ann} = \sqrt{AF \cdot \frac{1}{N-1} \sum_{t=1}^{N} (R_t - \bar{R})^2}
\end{equation}
where $N$ is the number of observations, $R_t$ is the return at time $t$, $\bar{R}$ is the mean return, and $AF$ is the annualization factor (e.g., 8760 for hourly data).\\

\noindent \textbf{Maximum Drawdown (MDD)} \cite{magdon2004}

\begin{equation}
    \text{MDD} = \max_{t \in [0, T]} \left( \frac{P_t - V_t}{P_t} \right)
\end{equation}
where $P_t = \max_{\tau \in [0, t]} V_\tau$ is the historical peak (high-water mark) of the portfolio value up to time $t$. This metric assesses the largest peak-to-trough drop, capturing extreme tail risks.\\

\noindent \textbf{Sharpe Ratio (Ann.)} \cite{sharpe1966}

\begin{equation}
    \text{Sharpe} = \frac{\text{CAGR} - R_f}{\sigma_{ann}}
\end{equation}
where $R_f$ is the risk-free rate.\\

\noindent \textbf{Sortino Ratio (Ann.)} \cite{sortino1994}

\begin{equation}
    \text{Sortino} = \frac{\text{CAGR} - R_f}{\sigma_{d}}
\end{equation}
where $\sigma_d$ is the annualized downside deviation, defined as:
\begin{equation}
    \sigma_d = \sqrt{AF \cdot \frac{1}{N} \sum_{t=1}^{N} \min(0, R_t - R_{target})^2}
\end{equation}
with the minimum acceptable return ($R_{target}$) set to zero.\\

\noindent \textbf{Calmar Ratio} 

The Calmar ratio, frequently employed in the evaluation of systematic trading strategies (e.g., \citet{korniejczuk2024}), is defined as:

\begin{equation}
    \text{Calmar} = \frac{\text{CAGR}}{\text{MDD}}
\end{equation}

\vspace{0.5cm}
\noindent \textbf{Operational Metrics}

In addition to return and risk metrics, we report several operational statistics that provide deeper insight into the strategy's structural efficiency and the agent's behavior. The percentage-based trade metrics are calculated from realized trade PnL relative to the allocated entry capital, incorporating the leverage applied during execution:
\begin{itemize}
    \item \textbf{Win Count and Lose Count:} The absolute number of profitable and unprofitable round-trip transactions, respectively, executed across the portfolio, net of the transaction fees recorded at position closing.
    \item \textbf{Average Win Return and Average Loss Return:} The mean percentage return generated exclusively by the winning trades, and the mean percentage loss incurred by the losing trades.
    \item \textbf{Average Trade Return:} The overall mean percentage return calculated across all executed trades, reflecting the per-trade expected value of the strategy, including the leverage applied during execution.
    \item \textbf{Average Trade Duration:} The mean duration, measured in hourly steps, that a position is held open before liquidation.
\end{itemize}

\vspace{0.5cm}
\noindent \textbf{Metric Evaluation Paradigm}

As the empirical evaluation of the strategy deliberately employs a leverage factor (detailed in Subsection \ref{subsec:market_fr_and_lev}), it is important to emphasize that all reported statistics, both portfolio-level risk/return metrics (e.g., CAGR, Annual Volatility, Max Drawdown) and trade-level operational metrics, are calculated strictly on a post-leverage basis. This ensures that the evaluated performance reflects the true operational reality of the margin account.

\subsection{Benchmark Selection}
\label{subsec:benchmark_selection}

To evaluate the performance of the baseline strategy, its cumulative returns were compared against two distinct benchmarks:

\begin{itemize}
    \item \textbf{Buy-and-Hold Bitcoin (B\&H BTC):} Represents a passive long-only exposure to the BTCUSDT trading pair. To ensure a fair comparison, it accounts for a 0.05\% transaction commission incurred during both the initial entry and the final liquidation.
    \item \textbf{Equal-Weight Portfolio (B\&H EWP):} Represents a passive allocation across the monthly universe ($N=100$ assets). At the start of each month, capital is rebalanced to an equal 1\% weight per asset. The portfolio is held passively until the next month, allowing weights to drift. This benchmark incorporates:
    \begin{itemize}
        \item \textit{Monthly Turnover Costs:} A 0.05\% fee is applied to all buy/sell orders during the universe update and rebalancing.
        \item \textit{Delisting Handling:} Assets with missing data are treated as forced liquidations; the recovered capital (net of fees) is held as cash until the end of the month.
    \end{itemize}
    
    It is important to note that the B\&H EWP benchmark operates on a strict \textit{month-to-month compounding} basis. The aggregate portfolio equity at the end of a given month serves as the initial capital base for the subsequent month's equal-weight reallocation. This ensures a mathematically fair and methodologically consistent comparison with the strategy's sequentially compounded equity curve.

\end{itemize}

\noindent \textbf{Leverage and Risk Alignment}

In this study, the performance of the strategies is evaluated using a leverage factor (see Subsection \ref{subsec:market_fr_and_lev}), while the benchmarks are evaluated unleveraged (1x). This discrepancy is intentional, grounded in the fundamental risk profiles of the respective approaches. Statistical arbitrage strategies historically exhibit significantly lower baseline volatility and shallower drawdowns compared to passive directional exposure \cite{Gatev2006, Krauss2017}. To facilitate a fair comparison of returns, the strategy's capital exposure is scaled to align its structural risk profile with the native risk metrics of the benchmarks \cite{asness2012}. However, this scaling is not a formal volatility- or drawdown-matching procedure; the 1x BTC and EWP benchmarks therefore serve as descriptive market context, while the primary evaluation is the comparison between Agent 2 and the baseline, both at 10x leverage under identical fee assumptions.

\vspace{0.5cm}
\noindent \textbf{Mathematical Formulation of Benchmarks}

Let $P_{i,t}$ denote the price of asset $i$ at time $t$, and $R_{i,t} = (P_{i,t} / P_{i,t-1}) - 1$ denote its periodic return. The cumulative return of the benchmarks is formalized as follows:

\begin{enumerate}
    \item \textbf{B\&H BTC ($V_{BTC,T}$):} With initial value $V_0$ and commission $c = 0.0005$:
    \begin{equation}
        V_{BTC,T} = V_0 (1 - c) \left[ \prod_{t=1}^{T} (1 + R_{BTC,t}) \right] (1 - c)
    \end{equation}
    
    \item \textbf{B\&H EWP ($V_{EWP,m}$):} For month $m$ and universe $\mathcal{U}_m$, the return (net of aggregate rebalancing costs $c_m$) is:
    \begin{equation}
        R_{EWP,m} = \left[ \frac{1}{N} \sum_{i \in \mathcal{U}_m} \left( \prod_{t \in m} (1 + R_{i,t}) \right) \right] - c_m
    \end{equation}
    where for any asset $i$ delisted at time $\tau \in m$, $R_{i,t} = 0$ for all $t > \tau$, simulating a transition to cash. The total portfolio value $V_{EWP,M}$ after $M$ months is obtained through the sequential compounding of these net monthly returns:
    \begin{equation}
        V_{EWP,M} = V_0 \prod_{m=1}^{M} (1 + R_{EWP,m})
    \end{equation}
\end{enumerate}

\section{Baseline Strategy Formulation}
\label{sec:bas_strategy_form}

The primary objective of the baseline strategy is to establish a robust training dataset for the Reinforcement Learning (RL) agent. Extensive full-space grid search optimization in algorithmic trading frequently leads to In-Sample (IS) overfitting and severe Out-Of-Sample (OOS) degradation \cite{bailey2014}. To mitigate this, we implemented a phased, heuristic parameter selection methodology.

By optimizing the strategy in sequential stages, we prioritize the discovery of "parameter plateaus" over isolated performance spikes, which drastically improves OOS generalization \cite{lin2024optimal}. This methodology allows us to first prove the efficacy of the core mean-reversion engine before applying secondary risk and allocation filters, creating a stable environment with distinct trading and stop-loss boundaries for the subsequent Deep Reinforcement Learning phase \cite{kim2019, zeng2025regimefolio}.

\subsection{Baseline Backtesting Methodology}

To rigorously prevent look-ahead bias, baseline parameter optimization is conducted strictly within the one-year In-Sample (IS) period (2024). The IS evaluation process for any given parameter set follows a structured protocol:
\begin{itemize}
    \item \textbf{Monthly Granularity:} The one-year backtest is decomposed into 12 discrete monthly iterations. In each iteration, the strategy trades pairs, which are dynamically re-selected at the start of each month based on the selection procedure described in Section 3.5.
    \item \textbf{Metric Distribution:} By analyzing the distribution (median, mean, and interquartile range) of the Sortino Ratio (Ann.) across these 12 monthly data points, we can identify parameter regions that demonstrate consistent profitability across the 12 months of the In-Sample year, rather than relying on a single aggregate return.
\end{itemize}

To ensure a robust and unbiased hyperparameter optimization process, the performance distributions in Figures \ref{fig:entry_coarse}-\ref{fig:sl_zoom_20} are constructed using separately evaluated monthly Sortino Ratios (without leverage, with default fee rate = 0.05\%). At this stage of evaluation, capital compounding is deliberately excluded. This methodological safeguard ensures that performance metrics are not disproportionately skewed by late-stage months, where the accumulated capital is higher. By maintaining constant capital boundaries within each isolated month, the optimization process correctly isolates the raw predictive edge of the parameters.\\

\vspace{0.5em}
\noindent \textbf{Static Architecture and Sequential Optimization}

Before initiating the optimization sequence, several structural parameters were strictly locked. These static parameters act as the invariant core of the strategy.
\begin{itemize}
    \item \textbf{Pairs = 20:} Portfolio size; capped to maintain highly cointegrated allocations. Varying this size during optimization risks cherry-picking historically "lucky" pair combinations rather than evaluating the system's underlying logic.
    \item \textbf{Z-Score Window = 168:} Fixed at one week to capture short-term micro-structural anomalies. While this specific horizon might appear arbitrary, it is grounded in two justifications:
    \begin{itemize}
        \item First, regarding market dynamics, a 168-hour window (roughly 1/4 of a trading month) ensures rolling metrics remain highly responsive to short-term reversions. This allows the spread to diverge and revert multiple times per episode, aligning with empirical evidence that shorter windows best adapt to sudden regime shifts in highly volatile cryptocurrency markets \cite{Krauss2017, Fischer2019}.
        \item Second, the Z-Score Window and Entry Threshold are inherently interdependent; expanding the window smooths the standard deviation, directly altering threshold breach frequencies. Optimizing both simultaneously risks severe data-mining bias and curve-fitting. In sequential optimization, it is methodologically sounder to deterministically lock the physical time horizon first, and subsequently optimize the statistical boundary (Entry Threshold) to fit that constraint.
    \end{itemize}
    \item \textbf{Exit Threshold = 0.0}: Fixed deterministically without optimization because it serves as the fundamental theoretical anchor of the strategy. Statistical arbitrage relies strictly on mean-reversion, meaning that the predictive edge is exhausted exactly when the spread reverts to its historical equilibrium. Locking the exit exactly at zero is not an optimization oversight, but a strict boundary condition that defines the strategy's mathematical identity, eliminating the need for empirical grid search.
\end{itemize}

Crucially, to ensure these static choices are not the result of arbitrary curve-fitting, their robustness is explicitly verified in the Sensitivity Analysis (Table \ref{tab:oos_sensitivity}). Having established these rigid constraints, we isolate the dynamic execution parameters: the Entry Threshold and the Stop Loss multiplier.

To optimize these variables, we performed a sequential optimization procedure instead of a full-space global grid search. In a multi-dimensional global search, it is notoriously difficult to isolate the marginal contribution of any individual parameter. Consequently, such exhaustive searches frequently gravitate toward mathematically optimal but structurally fragile combinations that overfit to a historical market noise \cite{lopez2018advances}. Therefore, we applied a phased, hierarchical approach, methodologically aligned with modern multi-stage quantitative systems \cite{zeng2025regimefolio}. 

By structurally freezing secondary parameters while evaluating the primary ones, we isolate the fundamental statistical validity of the core mean-reverting edge. Specifically, the Entry Threshold is first optimized independently, without the interference of a Stop Loss, to identify a natural, unconstrained market regime plateau. The Stop Loss multiplier is subsequently fitted only after this pure entry edge is secured. This sequential search comprehensively maps the performance landscape, prioritizing the discovery of broad, resilient "parameter plateaus" over isolated, over-optimized peaks.

\subsubsection{Stage 1: Entry Threshold Optimization}

The Stop Loss (SL) multiplier has been turned off to evaluate the quality of the pure entry signal. A coarse grid search was executed over a broad parameter space:
$$ \text{Entry Threshold} \in \{2.0, 2.25, 2.5, 2.75, 3.0, 3.25, 3.5, 3.75, 4.0\} $$

\begin{figure}[H]
    \caption{Coarse Grid Search for the Entry Threshold.}
    \label{fig:entry_coarse}
    \centering
    \includegraphics[width=\linewidth]{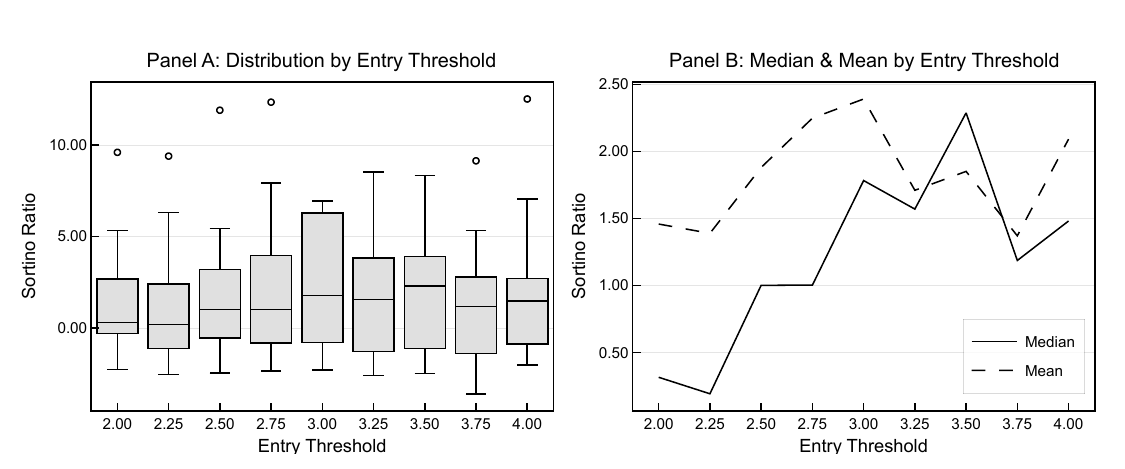}
    \justifying
    \noindent \justifying \noindent \scriptsize Note: The Stop Loss (SL) multiplier is disabled to evaluate the pure entry signal. Panel A displays the distribution of monthly Sortino Ratios in the one-year In-Sample backtest (2024), while Panel B tracks the mean and median of this distribution. Constant parameters: Exit Threshold = 0.0, Z-Score Window = 168, Pairs = 20.
\end{figure}

As illustrated in Figure \ref{fig:entry_coarse}, conducting the initial search without a structural Stop Loss reveals the raw predictive power of the mean-reversion signal. Two distinct regions of interest emerge, exhibiting localized peaks in the median Sortino ratio: the 3.0 threshold and the 3.5 threshold.

However, relying solely on isolated performance spikes without assessing the surrounding parameter landscape frequently leads to In-Sample (IS) overfitting. Statistically, requiring a higher Z-Score divergence significantly reduces the absolute number of executed trades. This diminished sample size naturally increases the probability of the algorithm curve-fitting to rare, idiosyncratic market anomalies, highly profitable outliers that artificially inflate IS metrics but exhibit poor Out-Of-Sample (OOS) generalization. Consequently, any apparent optimum at these elevated levels must be approached with a strong prior of skepticism.

To rigorously differentiate between structurally sound "parameter plateaus" and fragile "parameter islands", high-resolution analysis with a fine-grained step size of 0.05 was conducted around both candidates.

\begin{figure}[H]
    \caption{High-Resolution Local Sensitivity Analysis: 3.5 Entry Threshold.}
    \label{fig:entry_zoom_3_5}
    \centering
    \includegraphics[width=\linewidth]{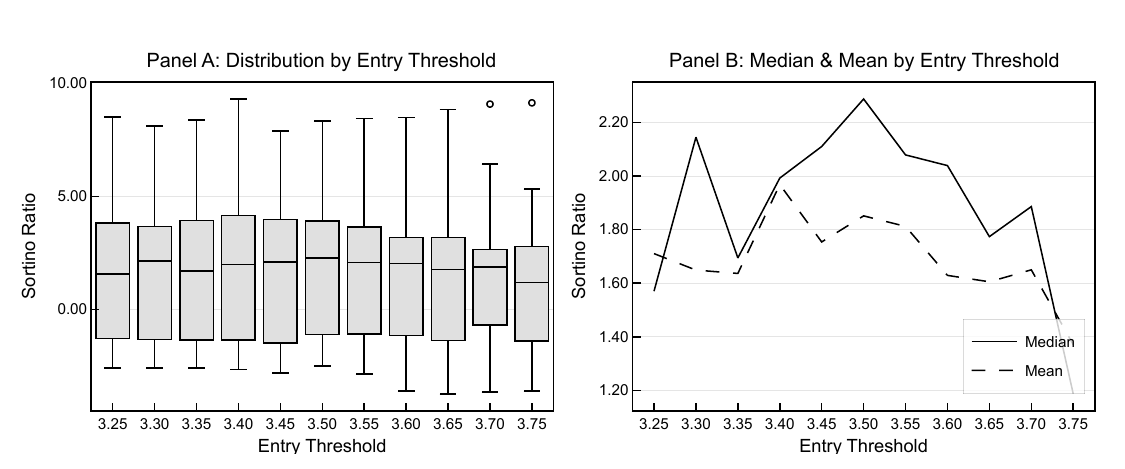}
    \justifying
    \noindent \justifying \noindent \scriptsize Note: 
    The search is conducted around the 3.5 region with a 0.05 step size (SL disabled). Panel A displays the distribution of monthly Sortino Ratios in the one-year In-Sample backtest (2024), while Panel B tracks the mean and median of this distribution. Constant parameters: Exit Threshold = 0.0, Z-Score Window = 168, Pairs = 20.
\end{figure}

High-resolution analysis around the 3.5 threshold (Figure \ref{fig:entry_zoom_3_5}) reveals a highly alarming statistical profile. Although the median Sortino ratio remains high around 3.5, the mean expected value is significantly lower. This severe divergence between the median and the mean indicates a highly left-skewed (negatively skewed) return distribution. In practice, this signifies that the strategy's typical monthly performance is artificially inflated by rare outliers, while the mathematical expectation is dragged down by severe drawdown events. This confirms that the 3.50 threshold is, with high certainty, a fragile, overfitted artifact.

\begin{figure}[H]
    \caption{High-Resolution Local Sensitivity Analysis: 3.0 Entry Threshold.}
    \label{fig:entry_zoom_3_0}
    \centering
    \includegraphics[width=\linewidth]{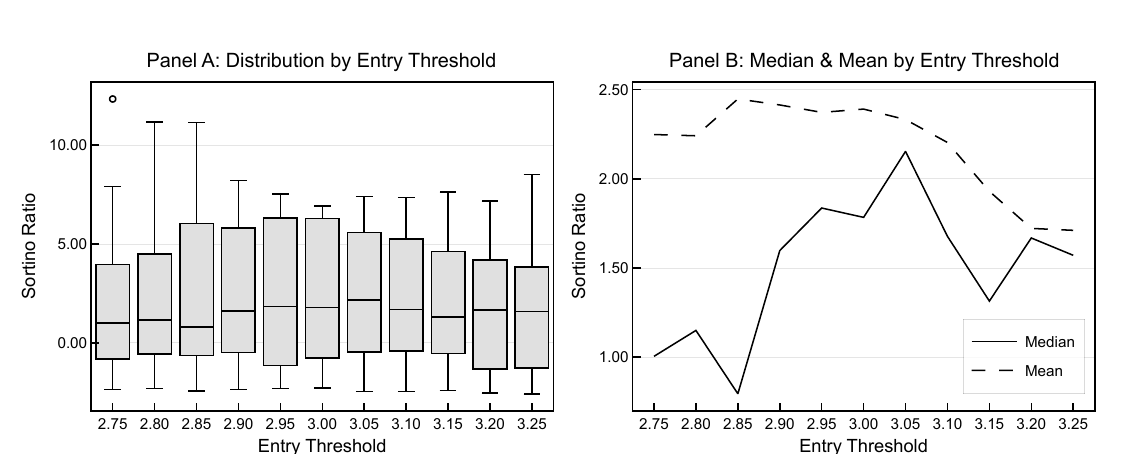}
\justifying
    \noindent \justifying \noindent \scriptsize Note: 
    The search is conducted around the 3.0 region with a 0.05 step size (SL disabled). Panel A displays the distribution of monthly Sortino Ratios in the one-year In-Sample backtest (2024), while Panel B tracks the mean and median of this distribution. Constant parameters: Exit Threshold = 0.0, Z-Score Window = 168, Pairs = 20.
\end{figure}

In contrast, the local analysis surrounding the 3.0 region (Figure \ref{fig:entry_zoom_3_0}) highlights a better view. Even though the median looks unstable, what is more important is the mean's behavior: unlike in the previous landscape, it is naturally above the median and significantly more stable. The graceful degradation of performance across adjacent parameters ensures that the model's edge is derived from a structural mean-reverting property rather than a coincidental data fit. The mean curve shows the steadily degrading plateau. Consequently, the value of an Entry Threshold = 3.0 was selected for the next stage.

\subsubsection{Stage 2: Stop Loss Optimization}

In this stage, the Entry Threshold was locked at the newly validated 3.0 to explore the broader Stop Loss parameter space:

$$ \text{Stop Loss} \in \{1.25, 1.5, 1.75, 2.0, 2.25, 2.5, 2.75, 3.0, 3.25\} $$

\begin{figure}[H]
    \caption{Coarse Grid Search for the Stop Loss.}
    \label{fig:sl_coarse}
    \centering
    \includegraphics[width=\linewidth]{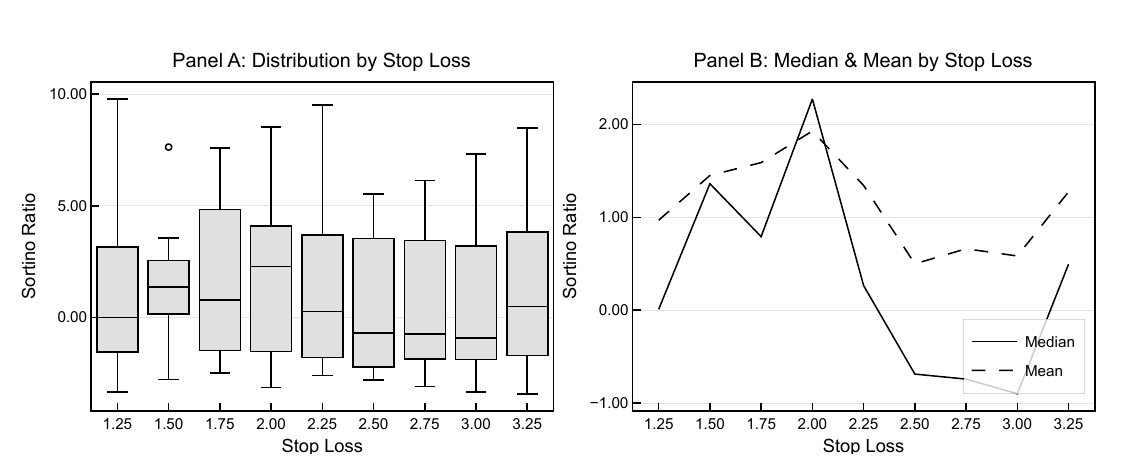}
    \justifying
    \noindent \justifying \noindent \scriptsize Note: Entry Threshold locked at the 3.0 optimum from Stage 1. Panel A displays the distribution of monthly Sortino Ratios in the one-year In-Sample backtest (2024), while Panel B tracks the mean and median of this distribution. Constant parameters: Entry Threshold = 3.0, Exit Threshold = 0.0, Z-Score Window = 168, Pairs = 20.
\end{figure}

The coarse search highlighted an optimum at SL = 2.0 (Figure \ref{fig:sl_coarse}). However, another region of interest emerged at the SL multiplier of 1.5. Visually, this parameter appeared highly attractive, exhibiting a tight Interquartile Range with the entire boxplot body positioned strictly above zero.

To conclusively differentiate between an overfitted artifact and a structurally sound parameter, similarly to the previous analysis of Entry Threshold, high-resolution Coarse-to-Fine analysis was conducted around both local optima.

\begin{figure}[H]
    \caption{High-Resolution Local Sensitivity Analysis: 1.5 Stop Loss.}
    \label{fig:sl_zoom_15}
    \centering
    \includegraphics[width=\linewidth]{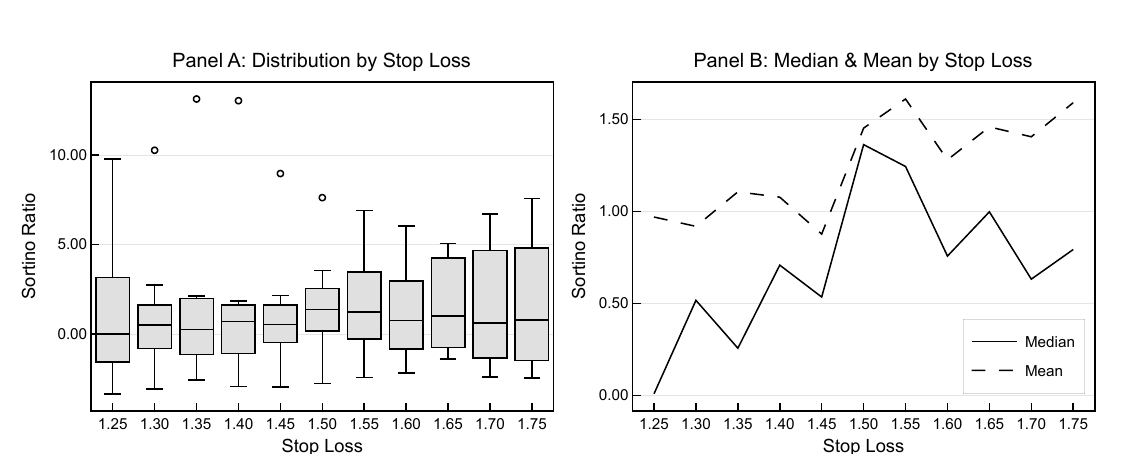}
    \justifying
    \noindent \justifying \noindent \scriptsize Note: 
    The search is conducted around the 1.5 region with a 0.05 step size. Panel A displays the distribution of monthly Sortino Ratios in the one-year In-Sample backtest (2024), while Panel B tracks the mean and median of this distribution. Constant parameters: Entry Threshold = 3.0, Exit Threshold = 0.0, Z-Score Window = 168, Pairs = 20.
\end{figure}

High-resolution analysis around the 1.50 value of the Stop Loss multiplier (Figure \ref{fig:sl_zoom_15}) revealed visible fragility. Performance exhibited severe discontinuity, where a minimal parameter shift of $\pm 0.05$ precipitated an explosion in negative variance. This indicates that the 1.50 threshold is likely an overfitted In-Sample artifact.

\begin{figure}[H]
    \caption{High-Resolution Local Sensitivity Analysis: 2.0 Stop Loss.}
    \label{fig:sl_zoom_20}
    \centering
    \includegraphics[width=\linewidth]{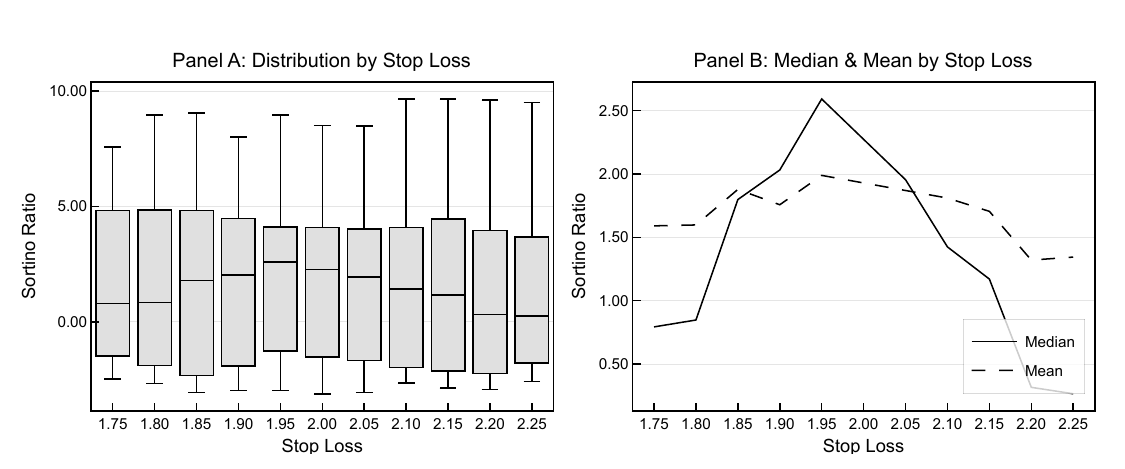}
    \justifying
    \noindent \justifying \noindent \scriptsize Note: 
    The search is conducted around the 2.0 region with a 0.05 step size. Panel A displays the distribution of monthly Sortino Ratios in the one-year In-Sample backtest (2024), while Panel B tracks the mean and median of this distribution. Constant parameters: Entry Threshold = 3.0, Exit Threshold = 0.0, Z-Score Window = 168, Pairs = 20.
\end{figure}

In contrast, the region surrounding 2.0 (Figure \ref{fig:sl_zoom_20}) exhibited a smooth convex optimization landscape. Although the median at Stop Loss = 2.0 exceeds the mean, the critical difference lies in the behavior of the mean itself. Instead of collapsing, the mean forms a highly stable and resilient plateau across the range from 1.85 to 2.10. 

This relationship ($Median > Mean$) is structurally expected when optimizing a Stop Loss parameter. A rigid stop-loss mechanism forcefully truncates adverse excursions, realizing predefined losses to prevent catastrophic tail risks. Consequently, the mathematical mean is naturally pulled down by these executed stop-outs, while the "typical" performance (the median) remains elevated because the threshold is wide enough not to be triggered by standard market noise. The key indicator of robustness here is that the mean maintains a stable plateau and the variance remains tightly controlled, efficiently balancing the trade-off between premature liquidations and tail-risk exposure.

Furthermore, while the absolute mathematical peak of the median occurred marginally lower at 1.95, we deliberately selected 2.0 as the final parameter. Selecting the absolute In-Sample peak introduces an unnecessary risk of micro-overfitting. Choosing 2.0 prioritizes structural simplicity, aligns symmetrically with the 3.0 entry threshold (equating to a strict Stop Loss Threshold = 6.0).\\

\section{Baseline Strategy Evaluation}
\label{sec:baseline_eval}

Having established a structurally robust parameter configuration (Entry Threshold = 3.0, Exit Threshold = 0.0, Stop Loss = 2.0, Z-Score Window = 168, Pairs = 20) through hierarchical optimization, the finalized baseline strategy was evaluated over the full In-Sample (IS) period and subsequently validated on the unseen Out-Of-Sample (OOS) dataset. The primary goal of this evaluation is not to present a flawless trading system, but to verify that the baseline generates a stable, positive expectancy in unseen market conditions, thereby providing a legitimate environment for the subsequent Reinforcement Learning training phase.

Prior to presenting the empirical results, it is imperative to reiterate that the baseline strategy is evaluated under a strict 10x leverage regime. As formalized in Subsection \ref{subsec:market_fr_and_lev}, this native, \textit{trade-to-trade} leverage mechanism is applied to systematically scale the strategy's capital exposure, thereby aligning its structural risk profile with the inherently high-variance benchmarks.

\subsection{In-Sample \& Out-Of-Sample Baseline Performance Analysis}

\begin{figure}[H]
    \caption{Cumulative Performance of the Baseline Strategy}
    \label{fig:baseline-vs-all}
    \centering
    \includegraphics[width=\linewidth]{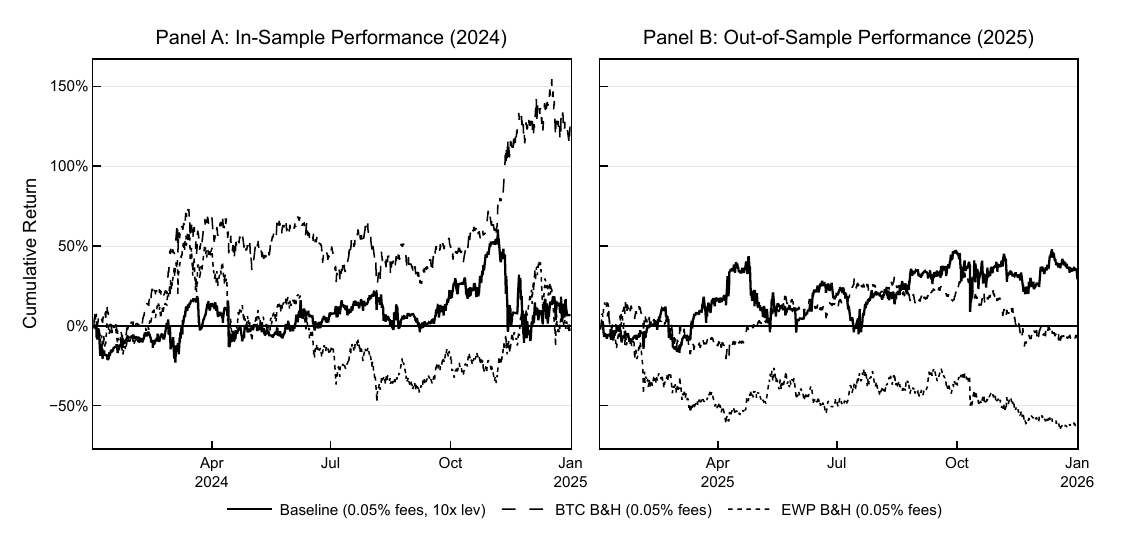}
    \justifying
    \noindent \justifying \noindent \scriptsize Note: 
    The figure compares the optimized baseline strategy against Buy-and-Hold Bitcoin (BTC) and Equal-Weight Portfolio (EWP) benchmarks. Panel A presents the In-Sample period (2024), and Panel B covers the Out-Of-Sample period (2025). Baseline parameters: Entry Threshold = 3.0, Stop Loss = 2.0, Exit Threshold = 0.0, Z-Score Window = 168, Pairs = 20. All curves account for 0.05\% transaction fees, with the baseline utilizing 10x leverage (see Subsection \ref{subsec:market_fr_and_lev}).
\end{figure}

\begin{table}[H]
    \centering
    \footnotesize
    \renewcommand{\arraystretch}{1.2}
    \caption{Out-Of-Sample Performance of the Baseline Strategy Against Benchmarks (2025).}
    \label{tab:oos-baseline}
    \vspace{12pt}
    \begin{tabularx}{\linewidth}{l*{3}{>{\centering\arraybackslash}X}}
    \toprule
        Metric & Baseline (10x) & BTC B\&H & EWP B\&H \\ 
    \midrule
        CAGR & 30.40\% & -6.43\% & -62.73\% \\
        Annual Volatility & 61.78\% & 44.36\% & 61.45\% \\
        Max Drawdown & 34.13\% & 34.75\% & 69.36\% \\
        \addlinespace[4pt]
        Win Count & 371 & - & - \\
        Loss Count & 245 & - & - \\
        Win Rate & 60.23\% & - & - \\
        \addlinespace[4pt]
        Avg Win Return & 22.89\% & - & - \\
        Avg Loss Return & -31.43\% & - & - \\
        Avg Trade Return & 1.28\% & - & - \\
        Avg Trade Duration & 55.04 & - & - \\
        \addlinespace[4pt]
        Sharpe Ratio (Ann.) & 0.4921 & -0.1451 & -1.0208 \\
        Sortino Ratio (Ann.) & 0.5360 & -0.1848 & -1.2795 \\
        Calmar Ratio & 0.8908 & -0.1852 & -0.9044 \\
    \bottomrule
    \end{tabularx}
    \justifying \noindent \scriptsize Note: Baseline: 0.05\% fees, leverage 10x; BTC/EWP B\&H benchmarks: 0.05\% fees. All reported metrics, both portfolio-level (CAGR, Annual Volatility, Max Drawdown, and Risk-Adjusted Ratios) and percentage-based trade-return metrics, are calculated on a post-leverage basis (see Subsection \ref{subsec:performance_metrics}).
\end{table}

The results demonstrate that the leveraged baseline strategy successfully outperformed both the B\&H BTC and B\&H EWP benchmarks. Specifically, the baseline achieved a CAGR of 30.40\%, along with an Annual Volatility of 61.78\% and a Maximum Drawdown of 34.13\%. This asymmetric risk-return profile is further reflected in a positive Sortino Ratio of 0.5360. However, it is crucial to reiterate that the primary objective of this baseline was to establish a stable foundation. By verifying its positive expectancy and structural robustness under rigorous margin conditions, the baseline provides a legitimate environment for the subsequent Deep Reinforcement Learning training phase.

\vspace{0.5em}
\subsection{Sensitivity Analysis}

To verify the structural stability of the baseline strategy, a comprehensive two-tier sensitivity analysis was performed. The first tier evaluates broad parameter variations, specifically testing wide grid increments (e.g., 0.5 steps for execution thresholds and 48-hour shifts for the Z-Score Window). This macro-level evaluation tests the strategy's resilience against major parameter miscalibration. The second tier functions as a structural ablation study, designed to test the foundational assumptions of the trading engine. Specifically, we assess the performance impact of the following modifications:
\begin{itemize}
    \item \textbf{Varying Fee Structures:} Evaluating the strategy's sensitivity to transaction costs.
    \item \textbf{Beta Hedge Deactivation:} Reverting from the dynamic empirical hedge to a strictly dollar-neutral allocation (fixed $\beta = 1$).
    \item \textbf{Risk Overlay Ablation:} Disabling specific Stop Loss (SL) mechanisms (\textit{SL Lock},
    \textit{Time Decay SL}) to isolate and quantify their individual contributions to tail-risk reduction.
\end{itemize}

The corresponding equity curves for each parameter perturbation (Entry Threshold, Exit Threshold, Stop Loss, Pairs and Z-Score Window) are reported in \ref{app:sensitivity}, Figures \ref{fig:sens_entry}--\ref{fig:sens_z_score_win}.

\begin{landscape}
\vspace*{\fill}
\renewcommand{\arraystretch}{1.2}
\begin{center}
\footnotesize
\captionof{table}{Sensitivity Analysis of Out-Of-Sample Baseline Strategy Performance (2025).}
\vspace{12pt}
\label{tab:oos_sensitivity}
\begin{tabularx}{\linewidth}{l*{11}{>{\centering\arraybackslash}X}}
\toprule
 & Baseline & \multicolumn{2}{c}{Entry Threshold} & \multicolumn{2}{c}{Exit Threshold} & \multicolumn{2}{c}{Stop Loss} & \multicolumn{2}{c}{Pairs} & \multicolumn{2}{c}{Z-Score Window} \\
 & - & 2.5 & 3.5 & -0.5 & 0.5 & 1.5 & 2.5 & 10 & 30 & 120 & 216 \\
\midrule
CAGR & 30.40\% & -13.97\% & -25.18\% & -18.09\% & 29.89\% & -31.57\% & 22.29\% & 15.99\% & -18.94\% & -43.50\% & -15.20\% \\
Annual Volatility & 61.78\% & 72.61\% & 51.51\% & 65.74\% & 58.73\% & 46.27\% & 70.42\% & 80.15\% & 51.89\% & 60.47\% & 62.26\% \\
Max Drawdown & 34.13\% & 46.66\% & 55.70\% & 48.06\% & 38.09\% & 40.68\% & 49.93\% & 45.16\% & 39.93\% & 52.42\% & 60.50\% \\[4pt]
Win Count & 371 & 564 & 213 & 324 & 413 & 277 & 395 & 188 & 526 & 466 & 313 \\
Loss Count & 245 & 381 & 152 & 277 & 233 & 325 & 216 & 127 & 382 & 310 & 215 \\
Win Rate & 60.23\% & 59.68\% & 58.36\% & 53.91\% & 63.93\% & 46.01\% & 64.65\% & 59.68\% & 57.93\% & 60.05\% & 59.28\% \\[4pt]
Avg Win Return & 22.89\% & 20.37\% & 26.14\% & 25.77\% & 20.23\% & 24.25\% & 22.68\% & 22.19\% & 23.79\% & 19.00\% & 25.89\% \\
Avg Loss Return & -31.43\% & -27.00\% & -38.17\% & -30.71\% & -32.78\% & -20.40\% & -37.75\% & -30.72\% & -31.68\% & -29.06\% & -35.57\% \\
Avg Trade Return & 1.28\% & 1.27\% & -0.64\% & -0.26\% & 1.11\% & 0.14\% & 1.32\% & 0.86\% & 0.45\% & -0.20\% & 0.87\% \\
Avg Trade Duration & 55.04 & 47.89 & 56.59 & 63.00 & 45.06 & 39.19 & 63.13 & 52.50 & 54.03 & 41.30 & 65.90 \\[4pt]
Sharpe Ratio (Ann.) & 0.4921 & -0.1924 & -0.4888 & -0.2752 & 0.5090 & -0.6823 & 0.3165 & 0.1996 & -0.3651 & -0.7193 & -0.2441 \\
Sortino Ratio (Ann.) & 0.5360 & -0.2171 & -0.4943 & -0.2975 & 0.5393 & -0.7204 & 0.3517 & 0.1919 & -0.4141 & -0.7351 & -0.2710 \\
Calmar Ratio & 0.8908 & -0.2994 & -0.4521 & -0.3765 & 0.7847 & -0.7760 & 0.4463 & 0.3542 & -0.4744 & -0.8298 & -0.2512 \\
\bottomrule
\end{tabularx}
\justifying \noindent \scriptsize Note: Baseline (0.05\% fees, leverage 10x): Entry Threshold = 3.0, Exit Threshold = 0.0, Stop Loss = 2.0, Pairs = 20, Z-Score Window = 168. The 10x leverage is applied to the baseline to scale
its inherently lower structural volatility and align its risk profile with the unleveraged benchmarks (see Subsection 3.9). Consequently, all
calculated performance metrics represent the post-leverage performance of the strategy (see Subsection 3.8).
\end{center}
\vspace*{\fill}
\end{landscape}

\noindent \textbf{Assumptions Verification}

The equity curves for each assumption ablation (Fee Rate, Beta Hedge, SL Lock and Time Decay SL) are provided in \ref{app:sensitivity}, Figures \ref{fig:mechanism_fee}--\ref{fig:mechanism_time_decay_sl}.

\begin{table}[H]
    \centering
    \footnotesize
    \renewcommand{\arraystretch}{1.2}
    \caption{Assumptions Verification: Out-Of-Sample Baseline Strategy Performance (2025).}
    \label{tab:oos-fee_hedge_sensitivity}
    \vspace{12pt}
    \begin{tabularx}{\linewidth}{l*{6}{>{\centering\arraybackslash}X}}
    \toprule
        Metric & Baseline & \makecell{Fee Rate \\ 0.00\%} & \makecell{Fee Rate \\ 0.10\%} & \makecell{Unhedged \\ } & \makecell{SL Lock \\ Disabled} & \makecell{Time Decay \\ SL \\ Disabled} \\
    \midrule
        CAGR & 30.40\% & 80.52\% & -6.02\% & -19.39\% & 1.98\% & -1.13\% \\
        Annual Volatility & 61.78\% & 61.35\% & 62.24\% & 58.46\% & 67.33\% & 64.32\% \\
        Max Drawdown & 34.13\% & 31.64\% & 39.26\% & 59.57\% & 46.13\% & 39.79\% \\[4pt]
        Win Count & 371 & 373 & 370 & 361 & 420 & 371 \\
        Loss Count & 245 & 243 & 246 & 240 & 287 & 243 \\
        Win Rate & 60.23\% & 60.55\% & 60.06\% & 60.07\% & 59.41\% & 60.42\% \\[4pt]
        Avg Win Return & 22.89\% & 23.26\% & 22.45\% & 22.78\% & 23.93\% & 22.77\% \\
        Avg Loss Return & -31.43\% & -31.20\% & -31.79\% & -33.04\% & -34.58\% & -33.64\% \\
        Avg Trade Return & 1.28\% & 1.77\% & 0.79\% & 0.49\% & 0.18\% & 0.44\% \\
        Avg Trade Duration & 55.04 & 55.04 & 55.04 & 53.29 & 57.97 & 60.31 \\[4pt]
        Sharpe Ratio (Ann.) & 0.4921 & 1.3125 & -0.0968 & -0.3316 & 0.0294 & -0.0175 \\
        Sortino Ratio (Ann.) & 0.5360 & 1.4329 & -0.1050 & -0.3591 & 0.0327 & -0.0194 \\
        Calmar Ratio & 0.8908 & 2.5450 & -0.1534 & -0.3255 & 0.0429 & -0.0284 \\
    \bottomrule
    \end{tabularx}\\
    \justifying \noindent \scriptsize Note: Baseline (0.05\% fees, leverage 10x): Beta Hedge = True, SL Lock = True, Time Decay SL = True. The 10x leverage is applied to the baseline to scale its inherently lower structural volatility and align its risk profile with the unleveraged benchmarks (see Subsection 3.9). Consequently, all
    calculated performance metrics represent the post-leverage performance of the strategy (see Subsection 3.8).
\end{table}

The comprehensive sensitivity analysis reveals that while the baseline strategy captures a genuine mean-reverting edge, its Out-Of-Sample success is highly conditional on strict structural constraints. The ablation study (Table \ref{tab:oos-fee_hedge_sensitivity}) demonstrates significant degradation when the foundational assumptions are removed. This suggests that dynamic hedging and strict regime filters are important components for limiting tail-risk events. Regarding parameter perturbations (Table \ref{tab:oos_sensitivity}), the model shows severe vulnerability to overly restrictive settings and hyper-reactive lookback periods, highlighting the inherent limitations and fragility of relying on static heuristics in highly noisy cryptocurrency markets. 

Ultimately, these findings perfectly motivate the subsequent integration of Deep Reinforcement Learning. The baseline successfully validates the core mathematical architecture (the \textit{Fixed Risk, Adaptive Mean} framework) and justifies the necessity of maintaining strict risk overlays. However, its significant sensitivity to rigid execution parameters underscores the need for a more adaptive decision engine. Consequently, the RL agent introduced in the next section is deployed precisely to replace these rigid thresholds with a dynamic, non-linear execution policy, operating safely within the boundaries of the verified risk management framework.

\section{RL Strategy Evaluation}
\label{sec:rl_eval}

\begin{table}[H]
    \centering
    \footnotesize
    \renewcommand{\arraystretch}{1.2}
    \caption{Agent Configuration Overview.}
    \label{tab:agent}
    \vspace{12pt}
    \setlength{\tabcolsep}{30pt}
    \begin{tabularx}{\linewidth}{c 
    >{\raggedright\arraybackslash}X 
    >{\raggedright\arraybackslash}X 
    c}
    \toprule
    Agent & Reward & Space & $\lambda$ \\
    \midrule
    1  & StepPnLReward  & Autonomous & 1.0 \\
    2  & StepPnLReward  & Autonomous & 1.2 \\
    3  & StepPnLReward  & Standard   & 1.0 \\
    4  & StepPnLReward  & Standard   & 1.2 \\
    5  & StepPnLReward  & Full       & 1.0 \\
    6  & StepPnLReward  & Full       & 1.2 \\
    [4pt]
    7  & TradePnLReward & Autonomous & 1.0 \\
    8  & TradePnLReward & Autonomous & 1.2 \\
    9  & TradePnLReward & Standard   & 1.0 \\
    10 & TradePnLReward & Standard   & 1.2 \\
    11 & TradePnLReward & Full       & 1.0 \\
    12 & TradePnLReward & Full       & 1.2 \\
    [4pt]
    13 & HybridActionReward & Autonomous & 1.0 \\
    14 & HybridActionReward & Autonomous & 1.2 \\
    15 & HybridActionReward & Standard   & 1.0 \\
    16 & HybridActionReward & Standard   & 1.2 \\
    17 & HybridActionReward & Full       & 1.0 \\
    18 & HybridActionReward & Full       & 1.2 \\
    \bottomrule
    \end{tabularx}\\
    \justifying \noindent \scriptsize 
    Note: Agent indexing corresponds to experimental configuration grid combining reward function, agent space, and $\lambda$ reward parameter (see Subsections \ref{subsubsec:obs_space} and \ref{subsubsec:rew_func}).
\end{table}
\vspace{1em}

\subsection{Training Analysis}

To simplify the identification of long-term trends, an Exponential Moving Average (EMA) was applied for the related training metrics. The two Actor-side diagnostics (Mean Episode Reward and Entropy Loss) are presented below in Figures \ref{fig:wandb_reward} and \ref{fig:wandb_entr_loss}, while the auxiliary Critic-side diagnostics (Value Loss and Explained Variance) are reported in \ref{app:sensitivity}, Figures \ref{fig:wandb_value_loss} and \ref{fig:wandb_expl_var}. In these visualizations, raw current values are displayed in the background, while the foreground illustrates the EMA-smoothed trajectories.

\begin{figure}[H]
    \caption{Training Diagnostic: Mean Episode Reward.}
    \label{fig:wandb_reward}
    \centering
    \includegraphics[width=\linewidth]{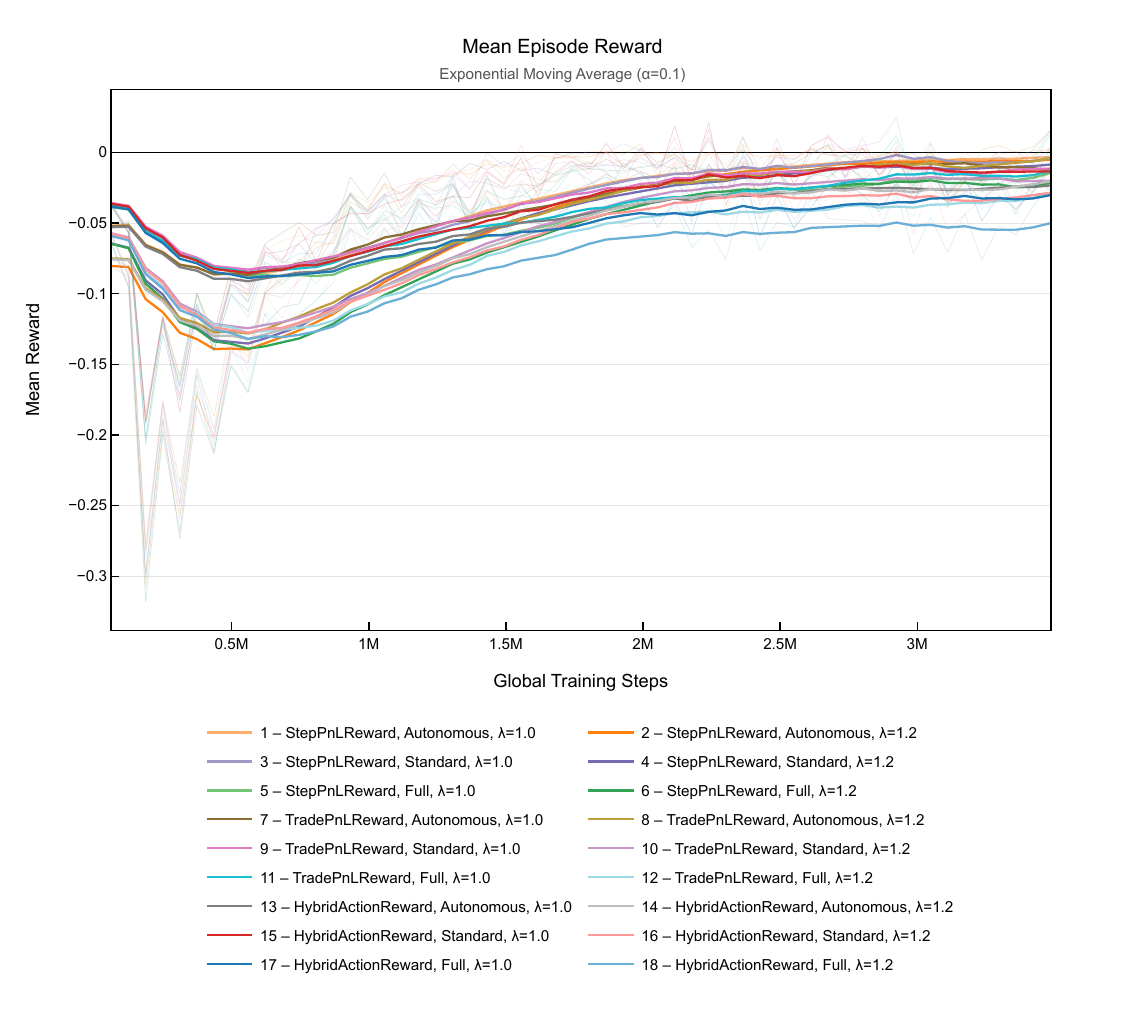}
    \justifying
    \noindent \justifying \noindent \scriptsize Note: 
    The figure illustrates mean episode reward obtained by various RL agent configurations during single 1-month training episodes. Raw current values are displayed in the background, with an Exponential Moving Average ($\alpha=0.1$) overlay highlighting long-term trends. An upward trend
    indicates that the agents learns to better maximize the reward function over time.
\end{figure}

\begin{figure}[H]
    \caption{Training Diagnostic: Entropy Loss.}
    \label{fig:wandb_entr_loss}
    \centering
    \includegraphics[width=\linewidth]{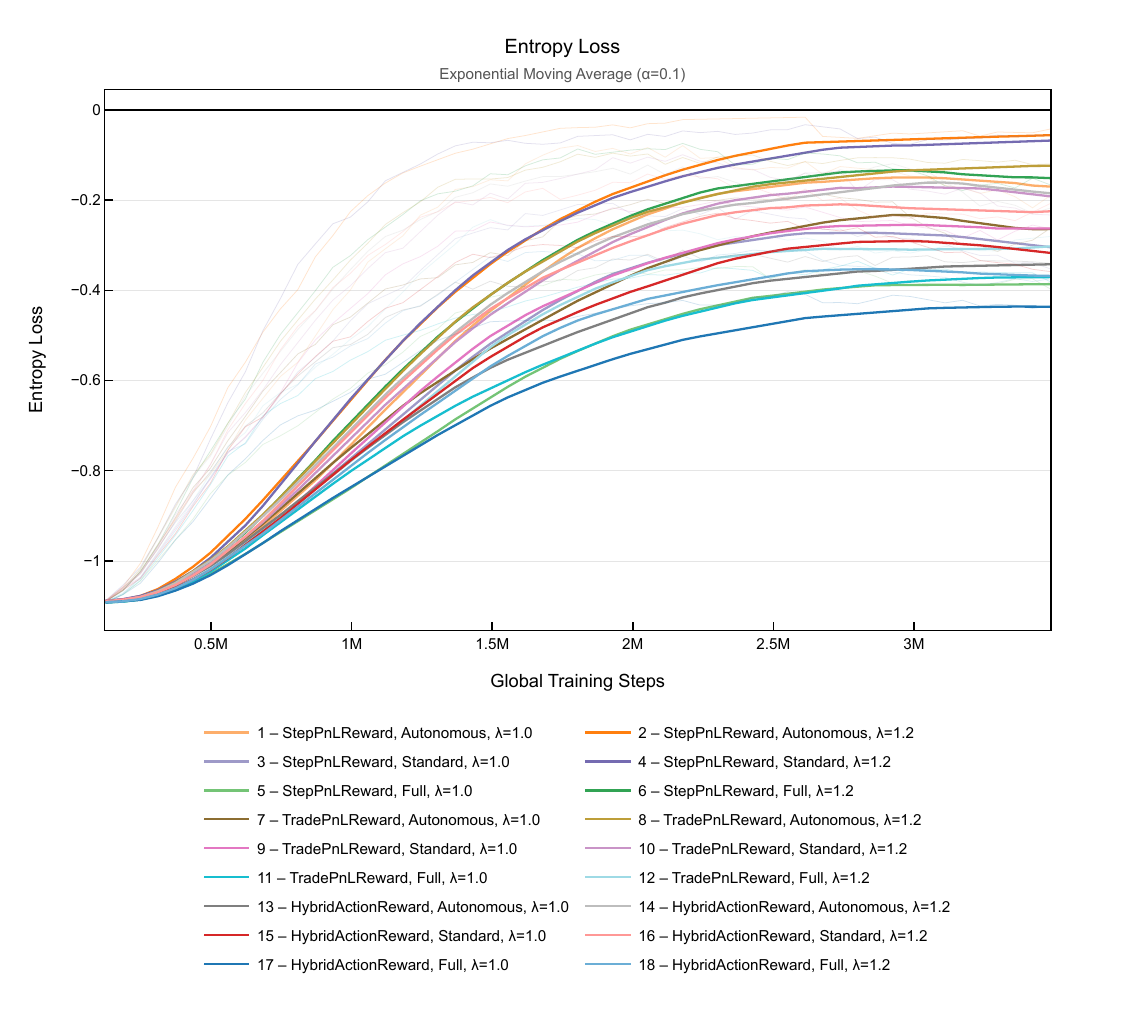}
    \justifying
    \noindent \justifying \noindent \scriptsize Note: 
    The figure illustrates measure of policy randomness obtained by various RL agent configurations during single 1-month training episodes. Raw current values are displayed in the background, with an Exponential Moving Average ($\alpha=0.1$) overlay highlighting long-term trends. The stabilization at a non-zero plateau demonstrates a healthy balance between exploration of learned patterns and continued exploration of the state space.
    \end{figure}

Training diagnostic metrics confirm robust and stable learning dynamics across all configurations. The Mean Episode Reward followed a consistent upward trajectory, recovering from initial losses caused by transaction costs to converge on a stable, near-zero or positive plateau. This progress was supported by the Critic network, where Value Loss narrowed to a consistent band (0.10–0.20) for all evaluated agents (see \ref{app:sensitivity}, Figure \ref{fig:wandb_value_loss}). 

While the Explained Variance (\ref{app:sensitivity}, Figure \ref{fig:wandb_expl_var}) peaked at a high level above 0.70, we observed a notably slower growth rate for agents utilizing the StepPnLReward. This is inherently tied to the dense nature of the step-based function, which forces the Critic to account for hourly price fluctuations and execution fees at every discrete interval. Such high-frequency noise reduces the signal-to-noise ratio, making the state-value function significantly more difficult to approximate than in sparser, trade-based reward schemes.

Furthermore, while the Explained Variance converging above 0.70 might intuitively raise concerns about the Critic network overfitting to training trajectories, evaluating it on a holdout validation set is methodologically flawed within the PPO framework. The Critic is strictly an auxiliary network used exclusively during In-Sample training for advantage estimation. During Out-Of-Sample deployment, the Critic is entirely discarded, and execution is governed solely by the Actor network. Consequently, the Critic's ability to generalize is structurally irrelevant. The true measure of the system's resistance to overfitting lies exclusively in the Actor's Out-Of-Sample performance, evaluated in subsequent sections.

Finally, the stabilization of Entropy Loss at a non-zero plateau (-0.20 to -0.40) ensured a healthy balance between exploitation and continued exploration, preventing the policy from premature convergence or overfitting.

\subsection{Model Selection}

Having confirmed healthy learning dynamics, the definitive model selection was conducted based on the In-Sample (2024) backtest performance. Evaluating Reinforcement Learning agents solely on standard quantitative metrics (such as the Sortino Ratio) frequently leads to "metric hacking" \cite{lopez2018advances}.

Crucially, both the baseline formulation and the RL model selection must be conducted on the identical In-Sample dataset. This is not a methodological oversight, but a strict architectural necessity of the hybrid framework. The RL agent is not an independent end-to-end trading system; it acts strictly as an execution overlay designed to optimize the exact mean-reverting regimes isolated by the baseline. Furthermore, during Out-Of-Sample deployment, the agent's actions are constrained by the deterministic Risk Management Overlay, which relies entirely on heuristics (Entry Threshold, Stop Loss) derived from the baseline's In-Sample behavior. Evaluating the agent on a different, unseen validation set before applying the risk overlay would sever the structural link between the learned policy and the baseline environment it was designed to navigate.

To ensure that the selected RL agent genuinely acts as an intelligent optimization layer over the established statistical foundation, the model selection followed a two-stage hierarchical framework focused on structural consistency:

\begin{itemize}
    \item \textbf{Regime-Fidelity Filter (Trade Duration Alignment):} As a primary filter, we required candidate agents to closely match the heuristic Baseline's Average Trade Duration ($\sim$55 hours). In financial Reinforcement Learning, unconstrained agents frequently suffer from "reward hacking," learning to exploit specific historical data sequences that fail to generalize Out-Of-Sample. To mitigate this, the duration alignment acts as a strict structural regularizer. It forces the agent to operate within the time regime and half-life explicitly confirmed during the pair selection phase, preventing the neural network from drifting into noisy, high-frequency scalping or long-term trend-following.
    \item \textbf{Win Rate Maximization:} Among the duration-aligned candidates, the final model was selected by maximizing the Win Rate. In the context of thin-margin statistical arbitrage, the win rate serves as a highly reliable proxy for signal precision and execution timing accuracy.
\end{itemize}

This dual-filter approach ensures that the RL agent enhances execution logic without increasing tail-risk exposure. Crucially, the evaluation of In-Sample performance for model selection was conducted with the heuristic Risk Management Overlay explicitly disabled. Since the parameters of this overlay were optimized over this exact same historical period (as detailed in Section \ref{sec:bas_strategy_form}), applying them during the agent evaluation phase would introduce a form of look-ahead bias and artificially mask the agent's true learned policy. By evaluating the agents in an unconstrained environment, we isolate and assess the pure predictive capability of the neural network. This ensures that the selected model genuinely internalized the market dynamics and learned optimal execution timing autonomously, rather than relying on the heuristically fitted guardrails.

The Risk Management Overlay is subsequently activated only for the Out-Of-Sample deployment to serve as a deterministic safety shield against unseen regimes. Consequently, this architectural asymmetry precludes the direct comparison and correlation of aggregate performance metrics between the IS and OOS periods. The unconstrained IS phase naturally generates highly variable equity curves that do not reflect the behavior of the final shielded system. Thus, attempting to correlate IS and OOS aggregate performance would be methodologically flawed. To ensure robust OOS generalization, the model selection process evaluates the intrinsic quality of the agent's decisions instead, relying on Average Trade Duration and Win Rate to isolate its true predictive edge.

\begin{landscape}
\vspace*{\fill}
\renewcommand{\arraystretch}{1.2}
\begin{center}
\footnotesize
\captionof{table}{Comparison of In-Sample RL Models Performance (2024).}
\label{tab:is_models}
\resizebox{\linewidth}{!}{
    \begin{tabular}{l*{18}{c}}
    \hline
 & 1 & 2 & 3 & 4 & 5 & 6 & 7 & 8 & 9 & 10 & 11 & 12 & 13 & 14 & 15 & 16 & 17 & 18 \\
    \hline
    CAGR & -14.86\% & 72.98\% & 294.66\% & -1.07\% & 11.32\% & 266.42\% & -31.29\% & 26.14\% & -13.82\% & 92.03\% & 10.70\% & 256.15\% & -49.02\% & 368.11\% & 15.05\% & 26.62\% & -55.91\% & 322.34\% \\
    Annual Volatility & 173.04\% & 94.22\% & 196.04\% & 149.25\% & 172.63\% & 80.91\% & 146.66\% & 172.07\% & 161.42\% & 206.75\% & 153.73\% & 144.60\% & 203.62\% & 172.81\% & 141.02\% & 164.00\% & 174.32\% & 179.33\% \\
    Max Drawdown & 81.87\% & 57.11\% & 76.89\% & 71.47\% & 82.19\% & 31.81\% & 79.92\% & 70.52\% & 83.32\% & 84.39\% & 83.52\% & 70.35\% & 86.28\% & 79.29\% & 75.90\% & 84.64\% & 87.96\% & 75.65\% \\[4pt]

    Win Count & 578 & 759 & 926 & 714 & 833 & 1009 & 618 & 677 & 812 & 771 & 955 & 636 & 873 & 771 & 691 & 620 & 1359 & 770 \\
    Loss Count & 415 & 276 & 482 & 429 & 476 & 455 & 403 & 436 & 448 & 465 & 484 & 350 & 493 & 439 & 377 & 430 & 695 & 437 \\
    Win Rate & 58.21\% & 73.33\% & 65.77\% & 62.47\% & 63.64\% & 68.92\% & 60.53\% & 60.83\% & 64.44\% & 62.38\% & 66.37\% & 64.50\% & 63.91\% & 63.72\% & 64.70\% & 59.05\% & 66.16\% & 63.79\% \\[4pt]

    Avg Win Return & 30.16\% & 15.49\% & 23.76\% & 20.85\% & 24.23\% & 11.67\% & 28.32\% & 28.53\% & 23.35\% & 25.53\% & 20.23\% & 26.83\% & 23.83\% & 26.82\% & 25.54\% & 29.55\% & 14.90\% & 24.26\% \\
    Avg Lose Return & -40.37\% & -36.34\% & -39.47\% & -30.50\% & -39.51\% & -19.63\% & -40.56\% & -39.62\% & -40.49\% & -38.26\% & -35.89\% & -39.38\% & -39.47\% & -39.68\% & -40.89\% & -38.41\% & -26.10\% & -36.50\% \\
    Avg Trade Return & 0.69\% & 1.67\% & 2.11\% & 1.58\% & 1.05\% & 1.94\% & 1.13\% & 1.84\% & 0.65\% & 1.53\% & 1.35\% & 3.33\% & 0.98\% & 2.70\% & 2.09\% & 1.72\% & 1.02\% & 2.26\% \\
    Avg Trade Duration & 107.85 & 45.61 & 87.61 & 63.32 & 92.35 & 22.04 & 98.64 & 102.05 & 89.41 & 92.91 & 76.99 & 92.73 & 93.61 & 92.63 & 92.40 & 101.00 & 51.21 & 86.80 \\[4pt]

    Sharpe Ratio (Ann.) & -0.0859 & 0.7746 & 1.5031 & -0.0072 & 0.0656 & 3.2926 & -0.2134 & 0.1519 & -0.0856 & 0.4451 & 0.0696 & 1.7714 & -0.2408 & 2.1301 & 0.1068 & 0.1623 & -0.3207 & 1.7975 \\
    Sortino Ratio (Ann.) & -0.1155 & 0.9764 & 2.0082 & -0.0085 & 0.0860 & 3.7930 & -0.2751 & 0.2031 & -0.1167 & 0.6027 & 0.0891 & 2.2566 & -0.3105 & 2.9274 & 0.1377 & 0.2161 & -0.4038 & 2.3540 \\
    Calmar Ratio & -0.1815 & 1.2779 & 3.8321 & -0.0149 & 0.1377 & 8.3758 & -0.3916 & 0.3707 & -0.1659 & 1.0905 & 0.1282 & 3.6411 & -0.5681 & 4.6426 & 0.1983 & 0.3145 & -0.6356 & 4.2610 \\
    \hline
    \end{tabular}
}
\scriptsize
\justifying \noindent \scriptsize 
    Note: The 10x leverage is applied to the agents to scale its inherently lower structural volatility and align its risk profile with the unleveraged benchmarks (see Subsection 3.9). Consequently, all
    calculated performance metrics represent the post-leverage performance of the strategy (see Subsection 3.8). Crucially, to assess the pure learned policy and avoid look-ahead bias from heuristically optimized boundaries, Risk Management Overlay = False during this In-Sample evaluation. Agents: 1 – StepPnLReward, Autonomous, $\lambda=1.0$, 2 – StepPnLReward, Autonomous, $\lambda=1.2$, 3 – StepPnLReward, Standard, $\lambda=1.0$, 4 – StepPnLReward, Standard, $\lambda=1.2$, 5 – StepPnLReward, Full, $\lambda=1.0$, 6 – StepPnLReward, Full, $\lambda=1.2$, 7 – TradePnLReward, Autonomous, $\lambda=1.0$, 8 – TradePnLReward, Autonomous, $\lambda=1.2$, 9 – TradePnLReward, Standard, $\lambda=1.0$, 10 – TradePnLReward, Standard, $\lambda=1.2$, 11 – TradePnLReward, Full, $\lambda=1.0$, 12 – TradePnLReward, Full, $\lambda=1.2$, 13 – HybridActionReward, Autonomous, $\lambda=1.0$, 14 – HybridActionReward, Autonomous, $\lambda=1.2$, 15 – HybridActionReward, Standard, $\lambda=1.0$, 16 – HybridActionReward, Standard, $\lambda=1.2$, 17 – HybridActionReward, Full, $\lambda=1.0$, 18 – HybridActionReward, Full, $\lambda=1.2$.
\end{center}
\vspace*{\fill}
\end{landscape}

Table \ref{tab:is_models} presents the In-Sample performance metrics across all evaluated RL configurations. According to that, three agents (2, 4, and 17) demonstrated Avg Trade Duration similar to the Baseline (respectively 45.61, 63.32, and 51.21 vs 55.04 of Baseline). The Agent 2, trained with the \textit{StepPnLReward} ($\lambda=1.2$) and operating within the \textit{Autonomous} Space demonstrated the highest In-Sample Win Rate among all candidates (73.33\%).

It is crucial to note the discrepancy: the majority of other agents trade, on average, twice as long as the Baseline, while Agent 6 trades for less than half as long (22.04 hours). Although Agent 6 achieved the highest In-Sample Sortino Ratio (3.7930), we interpreted its drastically shortened holding period as a sign that it may have decoupled from the statistical anomaly. By effectively reducing the expected mean-reverting half-life by more than 50\%, Agent 6 may have transitioned from trading the cointegrated equilibrium to exploiting transient microstructural noise. Such high-frequency behavior rarely generalizes to Out-Of-Sample cryptocurrency markets due to persistent frictions like transaction costs and slippage. Therefore, Agent 2 was preferred over Agent 6. This interpretation is consistent with the post-hoc Out-Of-Sample results (Table \ref{tab:oos_models}), in which the Sortino Ratio of Agent 6 fell from 3.7930 to 0.9673.

This alignment suggests that the RL Agent 2 internalized the natural half-life of the mean-reverting spread, optimizing entry and exit precision without drifting into directional risk or high-frequency overfitting. Consequently, this structurally sound policy was selected as the definitive RL Strategy for Out-Of-Sample verification.

\subsection{Out-Of-Sample Performance Analysis}

\begin{figure}[H]
    \caption{Cumulative Out-Of-Sample Performance of the Agent 2 Strategy.}
    \label{fig:rl_baseline_all}
    \centering
    \includegraphics[width=\linewidth]{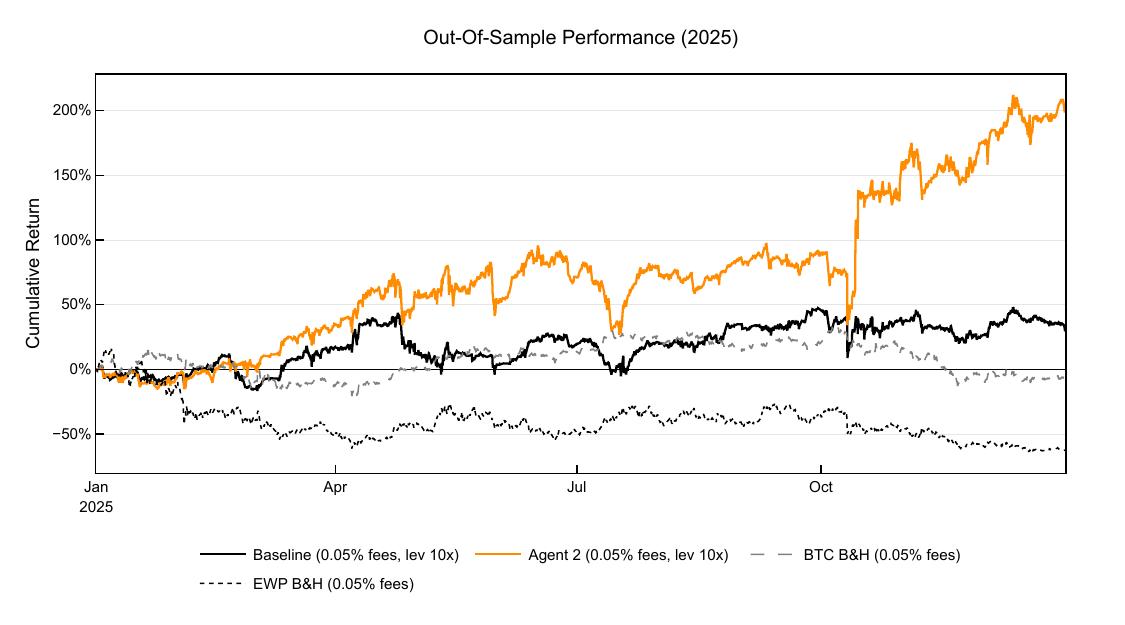}
    \justifying
    \noindent \justifying \noindent \scriptsize Note: The figure compares the Agent 2 performance against Baseline, Buy-and-Hold Bitcoin (BTC) and Equal-Weight Portfolio (EWP) benchmarks in the Out-Of-Sample period (2025). Agent 2 parameters: StepPnLReward, Autonomous, $\lambda=1.2$. Baseline parameters: Entry Threshold = 3.0, Stop Loss = 2.0, Exit Threshold = 0.0, Z-Score Window = 168, Pairs = 20. All curves account for 0.05\% transaction fees, with the Agent 2 and Baseline utilizing 10x leverage.
\end{figure}

\begin{table}[H]
    \centering
    \footnotesize
    \renewcommand{\arraystretch}{1.2}
    \caption{Out-Of-Sample Performance of Agent 2 Against Baseline and Benchmarks (2025).}
    \label{tab:oos-baseline-agent2}
    \vspace{12pt}
    \begin{tabularx}{\linewidth}{l*{4}{>{\centering\arraybackslash}X}}
    \toprule
        Metric & Baseline (10x) & Agent 2 (10x) & BTC B\&H & EWP B\&H \\ 
    \midrule
        CAGR & 30.40\% & 199.45\% & -6.43\% & -62.73\% \\
        Annual Volatility & 61.78\% & 70.68\% & 44.36\% & 61.45\% \\
        Max Drawdown & 34.13\% & 35.27\% & 34.75\% & 69.36\% \\
        \addlinespace[4pt]
        Win Count & 371 & 816 & - & - \\
        Loss Count & 245 & 299 & - & - \\
        Win Rate & 60.23\% & 73.18\% & - & - \\
        \addlinespace[4pt]
        Avg Win Return & 22.89\% & 13.62\% & - & - \\
        Avg Loss Return & -31.43\% & -30.62\% & - & - \\
        Avg Trade Return & 1.28\% & 1.76\% & - & - \\
        Avg Trade Duration & 55.0406 & 33.0726 & - & - \\
        \addlinespace[4pt]
        Sharpe Ratio (Ann.) & 0.4921 & 2.8220 & -0.1451 & -1.0208 \\
        Sortino Ratio (Ann.) & 0.5360 & 3.2494 & -0.1848 & -1.2795 \\
        Calmar Ratio & 0.8908 & 5.6557 & -0.1852 & -0.9044 \\
    \bottomrule
    \end{tabularx}
    \justifying \noindent \scriptsize Note: Baseline: 0.05\% fees, 10x leverage; Agent 2 (StepPnLReward, Autonomous, $\lambda=1.2$): 0.05\% fees, 10x leverage; Benchmarks: 0.05\% fees. The 10x leverage is applied to the baseline and agent strategies to scale their inherently lower structural volatility and align their risk profile with the unleveraged benchmarks (see Subsection \ref{subsec:benchmark_selection}). Consequently, all calculated performance metrics represent the post-leverage performance of the strategies (see Subsection \ref{subsec:performance_metrics}).
\end{table}

As illustrated in Figure \ref{fig:rl_baseline_all}, Agent 2 outperforms both the baseline strategy and the benchmarks in the Out-Of-Sample period. The RL execution overlay achieved a Sortino Ratio (Ann.) of 3.2494, improving upon the baseline's 0.5360 (while both include 0.05\% fees and leverage of 10x). Furthermore, the agent effectively enhanced execution timing, raising the win rate to 73.18\% from 60.23\% without increasing the maximum drawdown, which remained remarkably stable at 35.27\% compared to the baseline's 34.13\%. However, this result corresponds to a single training run (seed 42) and should not be regarded as representative; the dispersion across random seeds is analyzed in Subsection \ref{subsubsec:seed_variance}.

\subsection{Sensitivity Analysis}

The corresponding equity curves for each parameter perturbation of Agent 2 (Entry Threshold, Exit Threshold, Stop Loss, Z-Score Window and Pairs) are reported in \ref{app:sensitivity}, Figures \ref{fig:rl_sens_entry}--\ref{fig:rl_sens_top_n}, while the equity curves for each assumption ablation of Agent 2 (Fee Rate, Beta Hedge, SL Lock, Time Decay SL and Risk Management Overlay) are provided in \ref{app:sensitivity}, Figures \ref{fig:rl_mech_fee}--\ref{fig:rl_mech_auto}.

\begin{landscape}
\vspace*{\fill}
\renewcommand{\arraystretch}{1.2}
\begin{center}
\footnotesize
\captionof{table}{Sensitivity Analysis of Out-Of-Sample Agent Performance (2025).}
\vspace{12pt}
\label{tab:rl_oos_sensitivity}
\begin{tabularx}{\linewidth}{l*{11}{>{\centering\arraybackslash}X}}
\toprule
 & Agent 2 & \multicolumn{2}{c}{Entry Threshold} & \multicolumn{2}{c}{Exit Threshold} & \multicolumn{2}{c}{Stop Loss} & \multicolumn{2}{c}{Pairs} & \multicolumn{2}{c}{Z-Score Window} \\
 & - & 2.5 & 3.5 & -0.5 & 0.5 & 1.5 & 2.5 & 10 & 30 & 120 & 216 \\
\midrule
CAGR & 199.45\% & 126.90\% & 170.41\% & 154.91\% & 175.12\% & 53.70\% & 201.41\% & 345.79\% & 78.54\% & -18.46\% & 136.82\% \\
Annual Volatility & 70.68\% & 58.75\% & 74.88\% & 71.35\% & 70.98\% & 54.30\% & 78.09\% & 92.89\% & 61.30\% & 70.64\% & 69.27\% \\
Max Drawdown & 35.27\% & 28.41\% & 37.23\% & 39.14\% & 34.89\% & 31.30\% & 38.72\% & 38.75\% & 27.56\% & 57.23\% & 47.33\% \\[4pt]
Win Count & 816 & 769 & 828 & 819 & 842 & 715 & 835 & 425 & 1161 & 965 & 678 \\
Loss Count & 299 & 341 & 278 & 307 & 308 & 388 & 267 & 146 & 465 & 402 & 243 \\
Win Rate & 73.18\% & 69.28\% & 74.86\% & 72.74\% & 73.22\% & 64.82\% & 75.77\% & 74.43\% & 71.40\% & 70.59\% & 73.62\% \\[4pt]
Avg Win Return & 13.62\% & 13.72\% & 13.45\% & 13.62\% & 13.36\% & 14.03\% & 13.58\% & 13.68\% & 13.77\% & 11.93\% & 14.81\% \\
Avg Loss Return & -30.62\% & -25.58\% & -33.65\% & -31.06\% & -30.46\% & -22.31\% & -34.33\% & -31.41\% & -31.18\% & -26.92\% & -33.75\% \\
Avg Trade Return & 1.76\% & 1.65\% & 1.61\% & 1.44\% & 1.62\% & 1.25\% & 1.98\% & 2.15\% & 0.92\% & 0.51\% & 1.99\% \\
Avg Trade Duration & 33.07 & 28.65 & 35.56 & 33.25 & 31.79 & 26.02 & 36.65 & 31.02 & 32.53 & 26.39 & 40.42 \\[4pt]
Sharpe Ratio (Ann.) & 2.8220 & 2.1598 & 2.2758 & 2.1713 & 2.4671 & 0.9889 & 2.5793 & 3.7225 & 1.2811 & -0.2613 & 1.9753 \\
Sortino Ratio (Ann.) & 3.2494 & 2.4857 & 2.6384 & 2.4933 & 2.8351 & 1.0993 & 3.0275 & 3.8762 & 1.5107 & -0.2738 & 2.2620 \\
Calmar Ratio & 5.6557 & 4.4670 & 4.5771 & 3.9577 & 5.0195 & 1.7154 & 5.2010 & 8.9238 & 2.8491 & -0.3225 & 2.8905 \\
\bottomrule
\end{tabularx}
\justifying \noindent \scriptsize Note: Agent 2 (StepPnLReward, Autonomous, $\lambda=1.2$; 0.05\% fees, lev 10x): Exit Threshold = 0.0, Stop Loss = 2.0, Pairs = 20, Z-Score Window = 168, Risk Management Overlay = True. The 10x leverage is applied to the agent to scale
its inherently lower structural volatility and align its risk profile with the unleveraged benchmarks (see Subsection 3.9). Consequently, all
calculated performance metrics represent the post-leverage performance of the strategy (see Subsection 3.8).
\end{center}
\vspace*{\fill}
\end{landscape}

\begin{landscape}
\vspace*{\fill}
\renewcommand{\arraystretch}{1.2}
\begin{center}
\footnotesize
\captionof{table}{Assumptions Verification: Out-Of-Sample Agent Performance (2025).}
\vspace{12pt}
\label{tab:rl_oos_mechanism}
\begin{tabularx}{\linewidth}{l*{7}{>{\centering\arraybackslash}X}}
\toprule
 & Agent 2 & \multicolumn{2}{c}{Fee Rate} & \multicolumn{1}{c}{Unhedged} & \multicolumn{1}{c}{SL Lock} & \multicolumn{1}{c}{Time Decay SL} & \multicolumn{1}{c}{Risk Management Overlay} \\
 & - & 0.00\% & 0.10\% & (No Beta Hedge) & Disabled & Disabled & Disabled \\
\midrule
CAGR & 199.45\% & 446.83\% & 63.29\% & 300.72\% & 115.22\% & 189.08\% & 78.11\% \\
Annual Volatility & 70.68\% & 70.41\% & 70.99\% & 66.50\% & 79.63\% & 71.93\% & 91.55\% \\
Max Drawdown & 35.27\% & 31.47\% & 43.56\% & 46.95\% & 52.70\% & 37.69\% & 65.09\% \\[4pt]
Win Count & 816 & 820 & 802 & 801 & 964 & 815 & 817 \\
Loss Count & 299 & 295 & 313 & 292 & 439 & 300 & 252 \\
Win Rate & 73.18\% & 73.54\% & 71.93\% & 73.28\% & 68.71\% & 73.09\% & 76.43\% \\[4pt]
Avg Win Return & 13.62\% & 14.06\% & 13.36\% & 14.01\% & 15.60\% & 13.65\% & 13.10\% \\
Avg Loss Return & -30.62\% & -30.54\% & -29.72\% & -30.68\% & -31.74\% & -30.93\% & -37.17\% \\
Avg Trade Return & 1.76\% & 2.26\% & 1.26\% & 2.07\% & 0.79\% & 1.65\% & 1.25\% \\
Avg Trade Duration & 33.07 & 33.07 & 33.07 & 32.40 & 34.70 & 34.86 & 43.27 \\[4pt]
Sharpe Ratio (Ann.) & 2.8220 & 6.3459 & 0.8915 & 4.5221 & 1.4469 & 2.6285 & 0.8532 \\
Sortino Ratio (Ann.) & 3.2494 & 7.3387 & 1.0204 & 5.6702 & 1.6845 & 3.0431 & 0.9874 \\
Calmar Ratio & 5.6557 & 14.1993 & 1.4531 & 6.4052 & 2.1862 & 5.0170 & 1.2000 \\
\bottomrule
\end{tabularx}
\justifying \noindent \scriptsize Note: Agent 2 (StepPnLReward, Autonomous, $\lambda=1.2$; 0.05\% fees, leverage 10x): Exit Threshold = 0.0, Stop Loss = 2.0, Pairs = 20, Z-Score Window = 168, Risk Management Overlay = True. The 10x leverage is applied to the agent to scale its inherently lower structural volatility and align its risk profile with the unleveraged benchmarks (see Subsection 3.9). Consequently, all calculated performance metrics represent the post-leverage performance of the strategy (see Subsection 3.8). Risk Management Overlay Disabled removes the heuristic Take-Profit and Stop Loss enforcement and the associated SL Lock, but retains the time-based close at the end of the Z-Score Window and the margin-exhaustion liquidation.
\end{center}
\vspace*{\fill}
\end{landscape}

The sensitivity analysis confirms that the RL execution overlay largely mirrors the structural robustness of the baseline. While disabling the Beta Hedge surprisingly yielded higher CAGR and Sortino Ratio (Ann.), this outperformance initially suggested a potential increase in unhedged directional market exposure. Theoretically, removing the hedging component shifts the strategy's risk profile towards a more concentrated exposure to individual asset volatility. However, to verify whether this result stems from a higher risk premium or a genuine statistical edge, a formal econometric validation is presented in Subsection \ref{subsubsec:beta_hedge_ablation}.

Most importantly, the ablation study highlights the importance of the deterministic shielding layer. When the Risk Management Overlay is disabled, the portfolio suffers a catastrophic maximum drawdown of 65.09\% and annual volatility of 91.55\%. This starkly illustrates that while DRL excels at optimizing localized execution timing, it cannot inherently safeguard against out-of-distribution shocks without hard-coded risk guardrails, supporting the proposed hybrid architecture in the evaluated setting.

Moreover, the sensitivity analysis regarding portfolio cardinality (10 and 30 pairs vs default 20) confirms the structural robustness of the system and the efficacy of the pair selection mechanism. Reducing the portfolio to the top 10 pairs yields a significantly higher CAGR of 345.79\% and an improved Sortino Ratio (Ann.) of 3.8762, demonstrating that the "Filter-then-Rank" methodology successfully concentrates capital on pairs with the highest predictive expectancy. Conversely, expanding the portfolio to 30 pairs leads to performance degradation (CAGR: 78.54\%, Sortino Ratio: 1.5107) as lower-ranked pairs with weaker cointegration enter the basket. However, this expansion facilitates greater diversification, resulting in lower annual volatility (61.30\% vs. 70.68\%) and a reduced Maximum Drawdown (27.56\% vs. 35.27\%) compared to the 20-pair baseline. This predictable behavior across varying portfolio sizes confirms that the strategy avoids over-parameterization and follows a clear alpha-concentration versus risk-diversification trade-off.

Furthermore, the evaluation of transaction costs highlights the structural resilience of the strategy against severe market frictions. The isolated predictive engine demonstrates a massive edge in a zero-fee simulation, achieving a Sortino Ratio (Ann.) of 7.3387 and a CAGR of 446.83\%. Although the model comfortably absorbs the standard 0.05\% baseline commission, its true durability emerges under stress-test conditions. Applying a harsh 0.10\% friction rate, a proxy encapsulating both commission and execution slippage (see Subsection \ref{subsec:market_fr_and_lev}), still yields a CAGR of 63.29\% and a Sortino Ratio of 1.0204.

\subsection{Post-hoc Ablation Study}

Having evaluated the strategy on unseen data without look-ahead bias, this section conducts a retrospective, post-hoc analysis of the entire Out-Of-Sample period. By evaluating all configurations after the fact, we aim to deconstruct how different architectures and stochastic initializations ultimately impacted the final policy generalization.

\subsubsection{Evaluation of the Beta Hedge Constraint}
\label{subsubsec:beta_hedge_ablation}

The anomaly observed during the assumptions verification phase, where the Unhedged (BetaHedge = False) variant of Agent 2 exhibited a higher CAGR and Sortino Ratio compared to the Hedged model, motivated a post-hoc investigation. Initially, this outperformance was suspected to be an artifact of unhedged directional exposure, in which the strategy simply captures the underlying market trend rather than exploiting a mean-reverting edge. To empirically validate this hypothesis and isolate the true source of the generated returns, an Ordinary Least Squares (OLS) regression was conducted. The daily returns of both the Hedged and Unhedged variants of Agent 2 were regressed against the daily returns of the BTC benchmark over the entire Out-Of-Sample period (2025).

\begin{table}[H]
\centering
\footnotesize
\renewcommand{\arraystretch}{1.2}
\caption{Post-Hoc Ablation Regression Analysis: Agent 2 Hedged vs Unhedged Against BTC Benchmark.}
\label{tab:ablation-regression}
\begin{tabularx}{\linewidth}{l*{5}{>{\centering\arraybackslash}X}}
\toprule
Strategy & $\alpha$ & $p$-value ($\alpha$) & $\beta$ & $p$-value ($\beta$) & $R^2$ \\
\midrule
Agent 2 (Hedged) & 0.0038 & 0.0594 & 0.1789 & 0.0497 & 0.0106 \\
Agent 2 (Unhedged) & 0.0044 & 0.0201 & 0.0461 & 0.5929 & 0.0008 \\
\bottomrule
\end{tabularx}
\justifying\noindent\scriptsize Note: Ordinary Least Squares (OLS) regression of daily strategy returns against the BTC benchmark returns in the Out-Of-Sample period. $\alpha$ represents the daily abnormal return intercept, while $\beta$ represents the market exposure coefficient. All curves (Agents and BTC Benchmark) account for 0.05\% transaction fees,
with the Agents utilizing 10x leverage.
\end{table}

The empirical results presented in Table \ref{tab:ablation-regression} show no statistically significant directional market exposure for the unhedged variant. For the Unhedged strategy, the estimated market exposure coefficient ($\beta$) is remarkably low at 0.0461 and statistically insignificant (p-value = 0.5929). Conversely, it exhibits a positive and statistically significant daily abnormal return intercept ($\alpha$) of 0.0044 (p-value = 0.0201). Paradoxically, enforcing the dynamic hedging module (Beta Hedge = True) introduces a minor but statistically significant market correlation ($\beta$ = 0.1789, p-value = 0.0497) while simultaneously diminishing the significance of the $\alpha$ component (p-value = 0.0594). These findings suggest a low BTC exposure of the unhedged policy in the evaluated period, although a one-factor regression does not establish complete market neutrality. A plausible explanation for the performance difference is that the mechanical beta hedge forces the portfolio to maintain costly short positions in a highly noisy environment, injecting microstructural friction rather than isolating risk.

\subsubsection{Evaluation of Agent Configurations}

\begin{figure}[H]
    \caption{Out-Of-Sample Equity Curves: StepPnLReward.}    
    \label{fig:oos_all_rl_step}
    \centering
    \includegraphics[width=\linewidth]{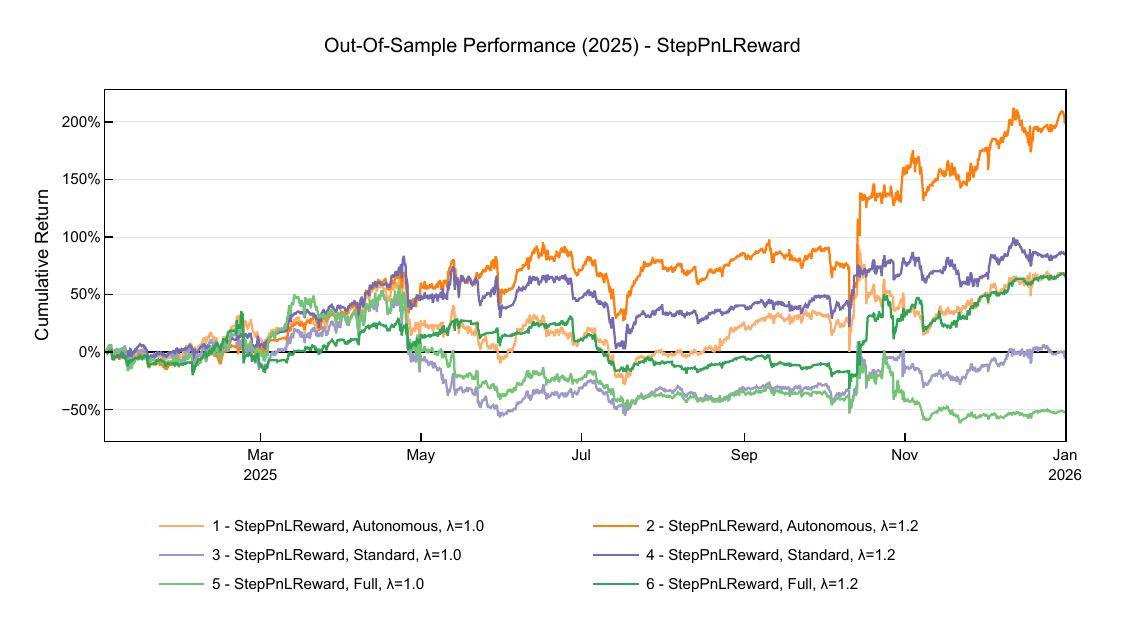}
    \justifying
    \noindent \justifying \noindent \scriptsize Note: The figure displays the Out-Of-Sample (2025) performance of RL agents trained using the dense StepPnLReward function. The plotted curves represent combinations of different observation spaces (Autonomous, Standard, Full) and loss aversion multipliers ($\lambda=1.0$ and $\lambda=1.2$). All simulations incorporate 0.05\% fees and 10x leverage.
\end{figure}

\begin{figure}[H]
    \caption{Out-Of-Sample Equity Curves: TradePnLReward.}    
    \label{fig:oos_all_rl_trade}
    \centering
    \includegraphics[width=\linewidth]{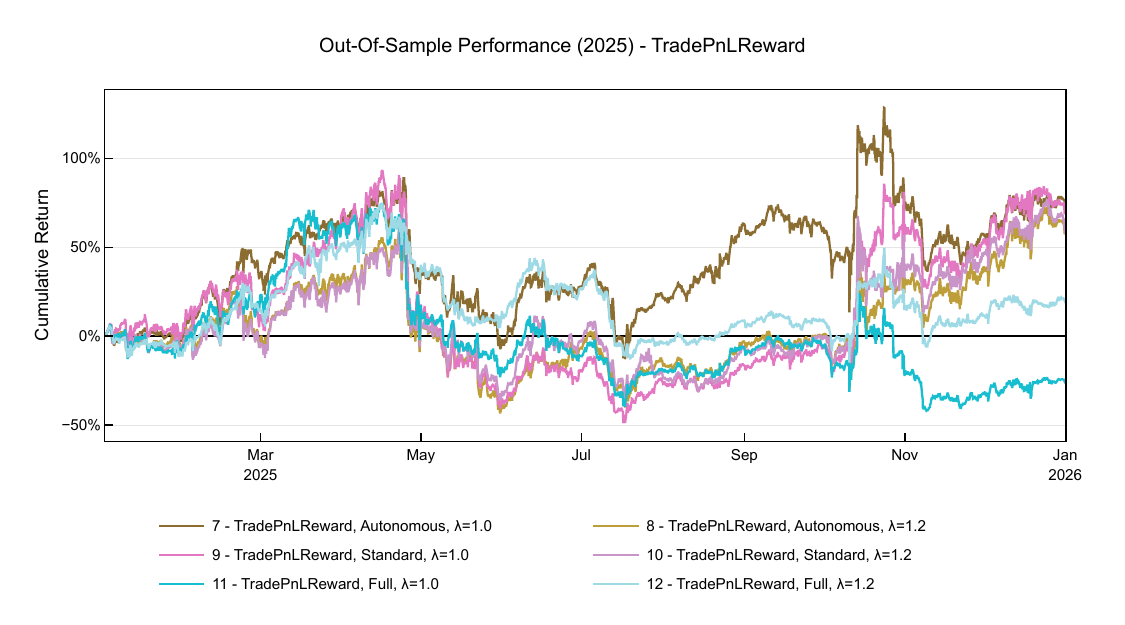}
    \justifying
    \noindent \justifying \noindent \scriptsize Note: The figure displays the Out-Of-Sample (2025) performance of RL agents trained using the sparse TradePnLReward function. The plotted curves represent combinations of different observation spaces (Autonomous, Standard, Full) and loss aversion multipliers ($\lambda=1.0$ and $\lambda=1.2$). All simulations incorporate 0.05\% fees and 10x leverage.
\end{figure}

\begin{figure}[H]
    \caption{Out-Of-Sample Equity Curves: HybridActionReward.}
    \label{fig:oos_all_rl_hybrid}
    \centering
    \includegraphics[width=\linewidth]{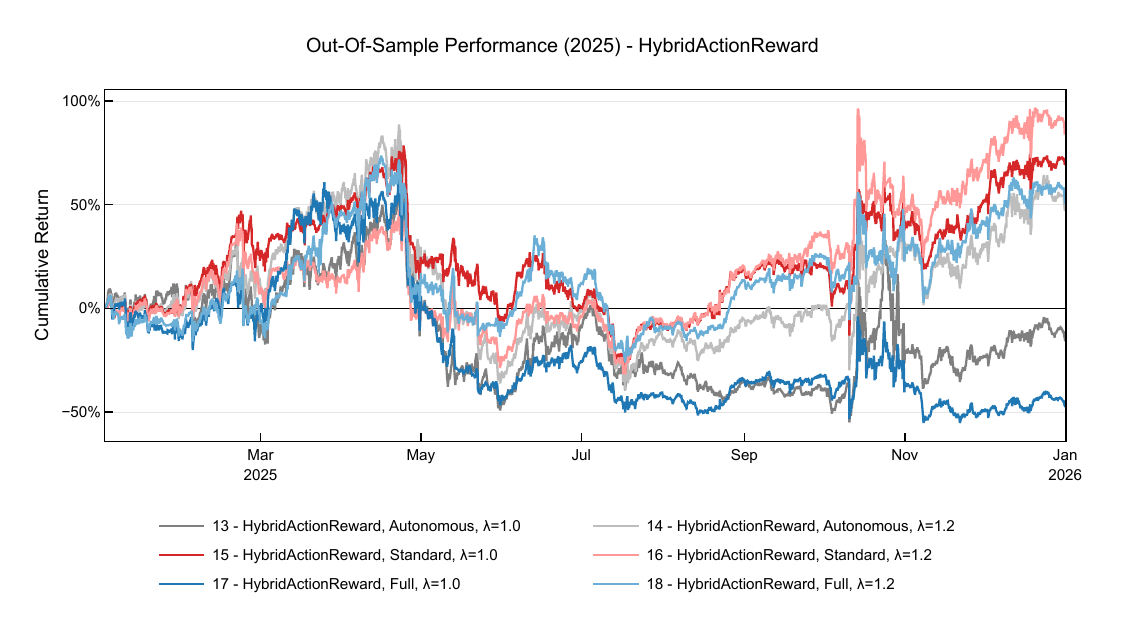}
    \justifying
    \noindent \justifying \noindent \scriptsize Note: The figure displays the Out-Of-Sample (2025) performance of RL agents trained using the HybridActionReward function. The plotted curves represent combinations of different observation spaces (Autonomous, Standard, Full) and loss aversion multipliers ($\lambda=1.0$ and $\lambda=1.2$). All simulations incorporate 0.05\% fees and 10x leverage.
\end{figure}

Having established the robustness of Agent 2 through Out-Of-Sample (OOS) testing and sensitivity analysis, we now conduct a post-hoc evaluation of all RL configurations. Figures \ref{fig:oos_all_rl_step}, \ref{fig:oos_all_rl_trade}, and \ref{fig:oos_all_rl_hybrid} illustrate the isolated OOS equity curves of all trained agents by their reward functions.

\begin{landscape}
\vspace*{\fill}
\renewcommand{\arraystretch}{1.2}
\begin{center}
\footnotesize
\captionof{table}{Comparison of Out-Of-Sample RL Models Performance (2025).}
\label{tab:oos_models}
\resizebox{\linewidth}{!}{
    \begin{tabular}{l*{18}{c}}
    \hline
 & 1 & 2 & 3 & 4 & 5 & 6 & 7 & 8 & 9 & 10 & 11 & 12 & 13 & 14 & 15 & 16 & 17 & 18 \\
    \hline
    CAGR & 63.24\% & 199.45\% & -5.19\% & 84.16\% & -52.60\% & 65.98\% & 74.22\% & 56.97\% & 67.55\% & 56.36\% & -27.28\% & 17.60\% & -16.62\% & 45.64\% & 68.81\% & 83.53\% & -48.01\% & 49.50\% \\
    Annual Volatility & 84.55\% & 70.68\% & 113.17\% & 64.72\% & 107.69\% & 70.19\% & 83.58\% & 105.09\% & 100.97\% & 100.33\% & 90.51\% & 65.64\% & 134.32\% & 107.82\% & 77.98\% & 86.25\% & 110.14\% & 90.24\% \\
    Max Drawdown & 57.36\% & 35.27\% & 71.73\% & 44.28\% & 75.75\% & 49.77\% & 54.05\% & 64.87\% & 73.58\% & 60.13\% & 66.96\% & 51.33\% & 72.28\% & 67.81\% & 61.17\% & 53.87\% & 73.07\% & 57.49\% \\[4pt]

    Win Count & 807 & 816 & 1656 & 622 & 1996 & 1136 & 885 & 1212 & 1449 & 1010 & 1522 & 668 & 1876 & 1315 & 710 & 801 & 2039 & 1053 \\
    Loss Count & 349 & 299 & 542 & 411 & 721 & 605 & 351 & 474 & 479 & 423 & 509 & 262 & 595 & 491 & 319 & 359 & 975 & 428 \\
    Win Rate & 69.81\% & 73.18\% & 75.34\% & 60.21\% & 73.46\% & 65.25\% & 71.60\% & 71.89\% & 75.16\% & 70.48\% & 74.94\% & 71.83\% & 75.92\% & 72.81\% & 69.00\% & 69.05\% & 67.65\% & 71.10\% \\[4pt]

    Avg Win Return & 15.65\% & 13.62\% & 9.97\% & 16.15\% & 7.83\% & 9.26\% & 13.22\% & 13.06\% & 10.96\% & 15.11\% & 8.78\% & 13.30\% & 10.15\% & 12.38\% & 16.33\% & 16.37\% & 8.37\% & 12.86\% \\
    Avg Lose Return & -32.95\% & -30.62\% & -28.52\% & -21.15\% & -20.54\% & -15.10\% & -30.99\% & -30.13\% & -29.40\% & -32.01\% & -24.01\% & -31.43\% & -29.54\% & -29.99\% & -32.72\% & -32.60\% & -16.34\% & -27.59\% \\
    Avg Trade Return & 0.98\% & 1.76\% & 0.48\% & 1.31\% & 0.30\% & 0.79\% & 0.66\% & 0.92\% & 0.94\% & 1.20\% & 0.56\% & 0.70\% & 0.59\% & 0.86\% & 1.13\% & 1.21\% & 0.38\% & 1.17\% \\
    Avg Trade Duration & 48.90 & 33.07 & 37.66 & 35.43 & 28.35 & 15.29 & 43.93 & 44.87 & 39.62 & 49.96 & 31.53 & 43.73 & 38.34 & 43.27 & 49.59 & 50.88 & 26.11 & 40.54 \\[4pt]

    Sharpe Ratio (Ann.) & 0.7480 & 2.8220 & -0.0459 & 1.3004 & -0.4884 & 0.9401 & 0.8879 & 0.5421 & 0.6690 & 0.5617 & -0.3014 & 0.2681 & -0.1237 & 0.4234 & 0.8823 & 0.9685 & -0.4359 & 0.5485 \\
    Sortino Ratio (Ann.) & 0.8800 & 3.2494 & -0.0530 & 1.4993 & -0.5777 & 0.9673 & 1.0406 & 0.6234 & 0.8104 & 0.6766 & -0.3587 & 0.3272 & -0.1424 & 0.4870 & 1.0473 & 1.1606 & -0.5076 & 0.6322 \\
    Calmar Ratio & 1.1026 & 5.6557 & -0.0723 & 1.9008 & -0.6944 & 1.3258 & 1.3732 & 0.8781 & 0.9181 & 0.9374 & -0.4074 & 0.3428 & -0.2299 & 0.6731 & 1.1248 & 1.5505 & -0.6570 & 0.8610 \\
    \hline
    \end{tabular}
}
\scriptsize
\justifying \noindent \scriptsize 
    Note: The 10x leverage is applied to the agents to scale its inherently lower structural volatility and align its risk profile with the unleveraged benchmarks (see Subsection 3.9). Consequently, all
    calculated performance metrics represent the post-leverage performance of the strategy (see Subsection 3.8). Risk Management Overlay = True (see Subsection \ref{subsubsec:agent_exec_risk_overlay}). Agents: 1 – StepPnLReward, Autonomous, $\lambda=1.0$, 2 – StepPnLReward, Autonomous, $\lambda=1.2$, 3 – StepPnLReward, Standard, $\lambda=1.0$, 4 – StepPnLReward, Standard, $\lambda=1.2$, 5 – StepPnLReward, Full, $\lambda=1.0$, 6 – StepPnLReward, Full, $\lambda=1.2$, 7 – TradePnLReward, Autonomous, $\lambda=1.0$, 8 – TradePnLReward, Autonomous, $\lambda=1.2$, 9 – TradePnLReward, Standard, $\lambda=1.0$, 10 – TradePnLReward, Standard, $\lambda=1.2$, 11 – TradePnLReward, Full, $\lambda=1.0$, 12 – TradePnLReward, Full, $\lambda=1.2$, 13 – HybridActionReward, Autonomous, $\lambda=1.0$, 14 – HybridActionReward, Autonomous, $\lambda=1.2$, 15 – HybridActionReward, Standard, $\lambda=1.0$, 16 – HybridActionReward, Standard, $\lambda=1.2$, 17 – HybridActionReward, Full, $\lambda=1.0$, 18 – HybridActionReward, Full, $\lambda=1.2$.
\end{center}
\vspace*{\fill}
\end{landscape}

Two critical observations emerge from this retrospective analysis. First, overly complex or sparse reward schemes (such as \textit{TradePnL} and \textit{HybridAction}) suffered severe OOS degradation. Second, in terms of observation spaces, an unexpected divergence was observed: agents operating without the explicit baseline signal embedded in their state representation frequently demonstrated superior adaptability.

However, this representational autonomy during the learning phase does not invalidate the necessity of the statistical baseline. As shown by the prior ablation study, the deterministic Risk Management Overlay, derived directly from the baseline's Entry Threshold and Stop Loss multiplier, remains an important component of Out-Of-Sample performance. This highlights a critical architectural distinction: while Deep RL demands an unconstrained observation space to effectively optimize execution timing, its final policy must remain strictly governed by external heuristic guardrails to prevent catastrophic tail-risk events.

These findings indicate a potential structural deficiency within the Full observation space. However, it remains to be verified whether this performance decline is due to abstract informational friction or measurable statistical redundancies. To resolve this, Subsection \ref{subsubsec:stat_audit} provides a Statistical Audit of the state representation, specifically examining the interaction between the Z-Score and the Hurst exponent through correlation and Variance Inflation Factor (VIF) analysis.

\subsubsection{Diagnosing State-Space Dimensionality Overhead}
\label{subsubsec:stat_audit}

To diagnose the root cause of performance degradation in the Full space, a statistical audit of the state vector components was conducted based on 240 training episodes (corresponding to the 12-month backtests spanning In-Sample 2024 for the top 20 pairs, the same as used for agent training). Five episodes (2.1\% of the sample) were excluded from the calculations due to zero trade activity, which would preclude reliable estimation of variance and correlation.

\begin{table}[H]
\centering
\footnotesize
\renewcommand{\arraystretch}{1.2}
\caption{Global Interaction Analysis: Pooled Correlation Matrix of State Variables.}
\label{tab:training-correlation-matrix}
\begin{tabularx}{\linewidth}{l*{5}{>{\centering\arraybackslash}X}}
\toprule
Metric & Z-Score & Hurst & Position & Signal & Norm. Time in Position \\
\midrule
Z-Score & 1.0000 & -0.0397 & -0.6378 & -0.2110 & -0.0366 \\
Hurst & -0.0397 & 1.0000 & 0.0109 & 0.0097 & 0.0273 \\
Position & -0.6378 & 0.0109 & 1.0000 & 0.2182 & 0.0724 \\
Signal & -0.2110 & 0.0097 & 0.2182 & 1.0000 & 0.0119 \\
Norm. Time in Position & -0.0366 & 0.0273 & 0.0724 & 0.0119 & 1.0000 \\
\bottomrule
\end{tabularx}
\justifying\noindent\scriptsize Note: The correlation matrix is computed on the pooled dataset aggregated across $N=235$ valid In-Sample 2024 training episodes to capture global feature interactions. Normalized Time in Position = Time in Position / Z-Score Window (where Z-Score Window = 168), see Subsection \ref{subsubsec:obs_space}.
\end{table}

\begin{table}[H]
\centering
\footnotesize
\renewcommand{\arraystretch}{1.2}
\caption{Multicollinearity Diagnostics: Pooled Variance Inflation Factor (VIF) Assessment.}
\label{tab:training-vif}
\begin{tabularx}{\linewidth}{l*{5}{>{\centering\arraybackslash}X}}
\toprule
Metric & Z-Score & Hurst & Position & Signal & Norm. Time in Position \\
\midrule
VIF & 1.7040 & 1.1432 & 1.7149 & 1.0597 & 1.1482 \\
\bottomrule
\end{tabularx}
\justifying\noindent\scriptsize Note: Variance Inflation Factor (VIF) values quantify the impact of multicollinearity on the variance of estimated coefficients. The VIF values are computed on the pooled dataset aggregated across $N=235$ valid In-Sample 2024 training episodes. A VIF value below 5 indicates the absence of severe multicollinearity. Normalized Time in Position = Time in Position / Z-Score Window (where Z-Score Window = 168), see Subsection \ref{subsubsec:obs_space}.
\end{table}

As illustrated in Tables \ref{tab:training-correlation-matrix} and \ref{tab:training-vif}, the aggregated data analysis directly addresses the hypothesis of feature multicollinearity. The global Variance Inflation Factor (VIF) for the Hurst exponent ($H_t$) is notably low (1.14), well below the critical threshold of 5. Furthermore, $H_t$ exhibits a near-zero global correlation with the primary Z-Score signal. This empirically rejects the premise that the performance degradation is driven by gradient instability caused by highly collinear inputs.Instead, the degradation may be rooted in noise injection and state-space dimensionality overhead. While $H_t$ provides statistically orthogonal information, it is a slow-moving, structural metric computed over a long formation window. Consequently, it lacks the high-frequency predictive variance required for micro-execution timing compared to highly dynamic Z-Score. Given the finite representational capacity of the LSTM network, forcing the agent to process this non-actionable feature dilutes the signal-to-noise ratio (SNR). The network inefficiently allocates its gradient updates to filter out this independent statistical noise, which ultimately hinders its ability to isolate critical, short-term execution triggers, which is consistent with the better performance of the parsimonious Autonomous space.

\subsubsection{Evaluation of Seed Variance}
\label{subsubsec:seed_variance}

To ensure that the results obtained for Agent 2 are not a product of coincidental stochastic alignment, a robustness check was performed across five independent random seeds ($s \in \{0, 1, 42, 100, 999\}$). This approach addresses the common criticism in Deep Reinforcement Learning literature regarding sensitivity to initial weight states and sampling order \cite{henderson2018}.

\begin{figure}[H]
    \caption{Training Diagnostic: Seed Variance Analysis of Mean Episode Reward.}
    \label{fig:wandb_seeds_var}
    \centering
    \includegraphics[width=\linewidth]{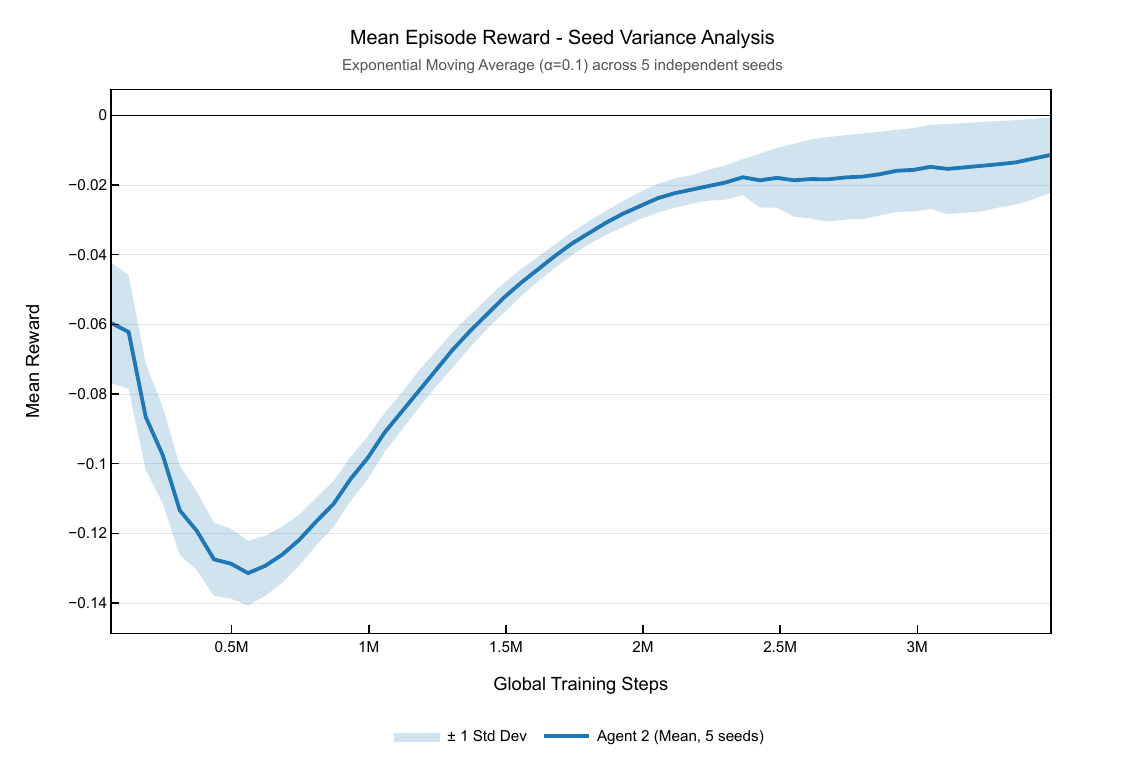}
    \justifying\noindent\scriptsize Note: The plot illustrates the training convergence of Agent 2 across five independent random seeds ($s \in \{0, 1, 42, 100, 999\}$). The solid line represents the mean of episode rewards smoothed with an Exponential Moving Average ($\alpha=0.1$), while the shaded area denotes the $\pm 1$ standard deviation interval.
\end{figure}

The training diagnostics presented in Figure \ref{fig:wandb_seeds_var} demonstrate that despite different random initializations of the neural networks, all agents followed a structurally similar and convergent learning trajectory. The initial high variance, characteristic of the exploration phase, gradually diminished as the agents learned the underlying mean-reverting dynamics of the cryptocurrency pairs. A notable observation is the stabilization of the standard deviation band until approximately 2.5M training steps, followed by a marginal widening in the final stage of the process. This specific dynamic is highly characteristic of deep reinforcement learning applied to noisy financial time series. During the initial training phases, all independently seeded agents successfully converge on the fundamental macro-level mean-reverting dynamics, resulting in tightly clustered performance. However, as training progresses into the final stages, the agents attempt to extract marginal gains by fine-tuning their policies to the high-frequency market noise \cite{lopez2018advances}. Driven by their distinct random initializations, individual agents begin to over-optimize for different micro-structural anomalies within the In-Sample dataset. This divergent fine-tuning naturally explains the slight widening of the performance variance at the end of the training cycle, a phenomenon consistent with the documented issues of policy instability and reproducibility in deep reinforcement learning \cite{henderson2018, engstrom2020}.

The translation of training stability into market performance is presented in Figure \ref{fig:seeds_var_oos} in \ref{app:sensitivity}. While the magnitude of cumulative returns exhibits noticeable dispersion, with Seed 42 acting as an upper outlier and Seed 100 underperforming, the overall directionality remains positive. The observed variance is a documented phenomenon in DRL applied to finance, where minor differences in trade timing under 10x leverage can lead to significantly different terminal equity levels. Detailed performance metrics for each run are summarized in Table \ref{tab:seed-variance-oos}.

\begin{table}[H]
\centering\footnotesize\renewcommand{\arraystretch}{1.2}
    \caption{Assumptions Verification: Out-Of-Sample Performance Stability Across Random Seeds (Agent 2, 2025).}
    \label{tab:seed-variance-oos}
    \begin{tabularx}{\linewidth}{l*{7}{>{\centering\arraybackslash}X}}
    \toprule Metric & Seed 0 & Seed 1 & Seed 42 & Seed 100 & Seed 999 & Mean & Std Dev \\ \midrule
        CAGR & 33.46\% & 48.69\% & 199.45\% & -3.00\% & 43.20\% & 64.36\% & 69.91\% \\
        Annual Volatility & 53.56\% & 59.23\% & 70.68\% & 40.23\% & 66.50\% & 58.04\% & 10.68\% \\
        Max Drawdown & 42.50\% & 47.95\% & 35.27\% & 42.48\% & 59.11\% & 45.46\% & 7.92\% \\[4pt]
        Win Rate & 61.21\% & 64.70\% & 73.18\% & 57.24\% & 67.90\% & 64.85\% & 5.47\% \\[4pt]
        Avg Trade Duration & 50.81 & 47.55 & 33.07 & 3.86 & 43.97 & 35.85 & 17.07 \\[4pt]
        Sharpe Ratio (Ann.) & 0.6247 & 0.8220 & 2.8220 & -0.0745 & 0.6497 & 0.9688 & 0.9762 \\
        Sortino Ratio (Ann.) & 0.6593 & 0.8906 & 3.2494 & -0.0532 & 0.7331 & 1.0958 & 1.1245 \\
        Calmar Ratio & 0.7872 & 1.0155 & 5.6557 & -0.0706 & 0.7308 & 1.6237 & 2.0491 \\
    \bottomrule
    \end{tabularx}
\justifying\noindent\scriptsize Note:  Out-of-Sample (2025) performance metrics for Agent 2, evaluated across 5 independent random initializations to assess policy robustness. Alongside individual seed performance, the aggregated mean and standard deviation (Std Dev) are provided to quantify outcome variance. All calculations assume 0.05\% trading fees and a 10x leverage multiplier.
\end{table}

The analysis of Out-Of-Sample performance across multiple seeds reveals significant dispersion in terminal returns, underscored by a CAGR standard deviation of 69.91\%. This variance highlights the inherent sensitivity of Deep RL agents to their initial weight states and sampling trajectories when navigating highly stochastic financial environments. While the originally selected Seed 42 (CAGR 199.45\%) represents an optimistic upper bound of the policy's potential, the underperformance of Seed 100 (CAGR -3.00\%) serves as a critical demonstration that even a structurally sound reward architecture can occasionally converge to a suboptimal local minimum during the final policy fine-tuning phase.

However, the evaluation should not rely on isolated extremes. The mean CAGR of 64.36\% and a positive mean Sortino Ratio of 1.0958 across all independent runs are considerably lower than the Seed 42 results, yet still roughly double those of the baseline (30.40\% and 0.5360, respectively). The fact that four out of five independent runs generated positive returns in the unseen 2025 period suggests that the agent captured part of the underlying mean-reverting signal rather than an isolated stochastic anomaly.

\subsubsection{Statistical Significance of Out-of-Sample Performance}

To address the risk of Out-of-Sample randomness, a stationary circular block bootstrap procedure \cite{Politis1994} was conducted (10,000 iterations, 168-hour block size to account for weekly seasonality and autocorrelation in hourly cryptocurrency returns). To ensure exact reproducibility of the statistical tests, the pseudo-random number generator was initialized with a fixed seed (42).

\begin{table}[H]
\centering\footnotesize\renewcommand{\arraystretch}{1.2}
    \caption{Out-Of-Sample Statistical Significance Testing of Performance Differences.}
    \label{tab:stat-sig-tests}
    \begin{tabular*}{\linewidth}{@{\extracolsep{\fill}} l c c c c c @{}}    
    \toprule
    Metric & Baseline & Agent 2 & Difference ($\Delta$) & 95\% CI of Diff. & p-value \\
    \midrule
    Sharpe Ratio (Ann.) & 0.4921 & 2.8220 & 2.3299 & [-0.4265, 11.3742] & 0.0603 \\
    Sortino Ratio (Ann.) & 0.5360 & 3.2494 & 2.7134 & [-0.5014, 13.9533] & 0.0593 \\
    \bottomrule
    \end{tabular*}
    \justifying \noindent \scriptsize Note: Baseline: 0.05\% fees, 10x leverage; Agent 2 (StepPnLReward, Autonomous, $\lambda=1.2$): 0.05\% fees, 10x leverage. Statistical significance computed via stationary circular block bootstrap methodology to account for autocorrelation and non-normality in hourly cryptocurrency returns (Block Bootstrap, 10 000 iterations, block = 168h, seed = 42). CI denotes Confidence Interval. The null hypothesis states that Agent 2 does not outperform the Baseline ($H_0: \Delta \le 0$).
\end{table}

As shown in Table \ref{tab:stat-sig-tests}, the differences in Sharpe and Sortino ratios between Agent 2 and the Baseline have p-values of 0.0603 and 0.0593, respectively, and both 95\% confidence intervals include zero, so the outperformance falls just short of the conventional 5\% significance threshold. These tests refer to the single reported Agent 2 run and were not corrected for multiple testing across the evaluated configurations and post-hoc analyses, and should therefore be read as exploratory. At the same time, the near-threshold p-values were obtained over a single year of hourly data, in which overcoming the inherent noise of cryptocurrency returns requires a substantial nominal outperformance; a longer Out-Of-Sample horizon would be needed to determine whether the effect reaches conventional significance.

\subsubsection{Sub-Period Performance Stability}

To address the risk of temporal overfitting, where aggregate annual outperformance might be disproportionately driven by a single, highly favorable market regime, a granular sub-period analysis of the Out-Of-Sample (2025) dataset was conducted. Cryptocurrency markets frequently exhibit rapid structural shifts in volatility and liquidity throughout a single calendar year. Consequently, evaluating the strategy in distinct quarterly segments (Q1–Q4) allows an assessment of temporal stability. This approach examines whether the outperformance of the Deep Reinforcement Learning execution overlay (Agent 2) over the heuristic baseline persists across varying market conditions, rather than being driven by an isolated anomaly.

\begin{table}[H]
    \centering
    \footnotesize
    \renewcommand{\arraystretch}{1.2}
    \setlength{\tabcolsep}{2pt}
    \caption{Sub-Period Out-Of-Sample Performance of the Baseline and Agent 2 (2025).}
    \label{tab:oos-sub-period}
    \vspace{6pt}
    \begin{tabularx}{\linewidth}{l*{8}{>{\centering\arraybackslash}X}}
    \toprule
        Metric & \multicolumn{2}{c}{Q1 2025} & \multicolumn{2}{c}{Q2 2025} & \multicolumn{2}{c}{Q3 2025} & \multicolumn{2}{c}{Q4 2025} \\ 
         & Baseline & Agent 2 & Baseline & Agent 2 & Baseline & Agent 2 & Baseline & Agent 2 \\ 
    \midrule
        CAGR & 90.77\% & 237.43\% & 10.13\% & 164.93\% & 116.47\% & 49.31\% & -33.29\% & 525.29\% \\
        Annual Volatility & 58.71\% & 64.70\% & 67.54\% & 78.39\% & 48.91\% & 53.09\% & 69.86\% & 82.74\% \\
        Max Drawdown & 25.85\% & 16.62\% & 33.27\% & 23.07\% & 23.78\% & 30.81\% & 25.44\% & 29.88\% \\
        \addlinespace[6pt]
        Sharpe Ratio (Ann.) & 1.5461 & 3.6699 & 0.1499 & 2.1040 & 2.3811 & 0.9287 & -0.4764 & 6.3485 \\
        Sortino Ratio (Ann.) & 1.9499 & 4.6306 & 0.1730 & 2.5028 & 2.9129 & 1.0773 & -0.4515 & 6.8420 \\
        Calmar Ratio & 3.5114 & 14.2839 & 0.3043 & 7.1485 & 4.8977 & 1.6004 & -1.3086 & 17.5815 \\
    \bottomrule
    \end{tabularx}
    \justifying \noindent \scriptsize \textit{Note: Performance metrics isolated for each calendar quarter of 2025. All metrics are calculated by rebasing the initial capital at the start of each period to ensure independent evaluation. Baseline and Agent 2 utilize 0.05\% fees and 10x leverage.}
\end{table}

The sub-period evaluation suggests that the outperformance of the RL execution overlay is not an artifact of a single favorable market phase. As illustrated in Table \ref{tab:oos-sub-period}, Agent 2 generated superior risk-adjusted returns (Sharpe, Sortino, and Calmar Ratios) in three out of the four analyzed quarters compared to the baseline. Most notably, during the adverse market conditions of Q4 2025, the baseline strategy suffered a severe degradation in performance, yielding negative returns (CAGR of -33.29\%) and failing to generate positive expectancy. In contrast, Agent 2 remained profitable in this quarter, achieving an annualized CAGR of 525.29\% and a Calmar Ratio of 17.58, although annualized figures based on a single quarter should be interpreted with caution. While the baseline temporarily outperformed in Q3, the reported Agent 2 run remained profitable over the evaluated OOS period.

\section{Conclusions and Further Research}
\label{sec:conclusion}
The primary contribution of this study to the quantitative finance literature lies in empirically showing that, in the evaluated setting, the persistent divergence risks inherent in highly volatile cryptocurrency markets \cite{Makarov2020, Fischer2019} can be effectively mitigated through a decoupled hybrid architecture. While contemporary research increasingly favors end-to-end Deep Reinforcement Learning (DRL) for financial portfolio optimization \cite{jiang2017cryptocurrency, guijarro2021}, the findings of this study validate the concerns regarding neural policy instability and out-of-sample brittleness \cite{henderson2018, bailey2014}. By empirically demonstrating that DRL can yield superior results when constrained as a specialized execution overlay, this research successfully bridges classical statistical arbitrage with modern expert systems. Rather than relying on the rigid applications of foundational frameworks, such as the static, distance-based method \citet{Gatev2006}, this research demonstrates that navigating non-stationary digital asset regimes requires a fundamentally dynamic approach. By introducing the "Filter-then-Rank" methodology and the "Fixed Risk, Adaptive Mean" model, this study proposes a highly adaptive alternative that effectively overcomes the structural limitations of static mean-reverting models highlighted by \citet{Do2010}. Ultimately, the application of this architecture provides a practical realization of "safe reinforcement learning via shielding" \cite{garcia2015comprehensive, alshiekh2018safe}, suggesting that neural execution timing should remain anchored within statistical boundaries.

The empirical findings of this study provide a comprehensive validation of the proposed methodology through the lens of two primary hypotheses. The first hypothesis (H1) posited that \textit{a dynamic statistical baseline, driven by a hierarchical pair selection framework and a custom execution concept, outperforms traditional benchmarks in the Out-Of-Sample period}. Addressing H1 and the first research question (RQ1), the results indicate that the baseline strategy, even without neural optimization, captured an edge in the evaluated period by employing cointegration and structural filters to mitigate the risk of divergence. This suggests that the "Filter-then-Rank" selection pipeline identifies pairs that exhibited mean-reverting properties in the evaluated period. The precise efficacy of the pair-selection mechanism was demonstrated during the sensitivity analysis of the neural execution model, which revealed a clear alpha-concentration versus risk-diversification trade-off: restricting the basket to fewer top-scoring pairs maximized profitability, while expanding cardinality systematically reduced volatility. Separately, the second hypothesis (H2) stated that \textit{deploying Deep Reinforcement Learning as an execution overlay enhances the profitability and risk-adjusted performance of a heuristic statistical arbitrage baseline}. To address H2 and the second research question (RQ2), the optimized PPO-LSTM agent demonstrated superior timing precision, with a Sortino Ratio of 3.2494 compared to 0.5360 for the baseline. Viewed holistically, the results for H1 and H2 suggest that the two components are complementary: while the statistical baseline provides the pair selection and risk boundaries, the DRL overlay acts as a non-linear optimizer that internalizes the natural half-life of spreads. However, a stationary circular block bootstrap procedure yields p-values of approximately 0.06, which fall just short of the conventional 5\% significance threshold, and the magnitude of the improvement varied substantially across random seeds, although four of the five seeds outperformed the baseline.

Addressing the third research question (RQ3), which asked how different reinforcement learning configurations and risk overlays impact Out-Of-Sample robustness, the extensive post-hoc ablation studies conducted at the conclusion of this research offer critical insights into the internal mechanics of the model. A statistical audit of the state-space dimensionality, including a Variance Inflation Factor (VIF) analysis, revealed that feeding slow-moving structural metrics into the agent’s observation space created informational friction rather than beneficial multicollinearity; therefore, representational autonomy yielded the highest signal-to-noise ratio. Furthermore, the ablation of the beta hedge constraint showed no statistically significant BTC exposure of the unhedged agent, although this does not establish market neutrality or explain the performance difference. Most importantly, the ablation of the Risk Management Overlay indicated that the deterministic shielding layer plays a key role in limiting losses; without it, the agent’s policy became unstable during out-of-distribution shocks, leading to catastrophic drawdowns. Additionally, the sub-period performance stability analysis revealed that while the heuristic baseline struggled during the adverse structural shifts in the final quarter (Q4) of the testing period, the reported Agent 2 run remained profitable in that quarter.

These insights suggest that an RL-driven execution overlay can enhance capital efficiency and adaptability in high-noise environments. However, these practical implications must be contextualized within explicit limitations. The Out-Of-Sample validation was deliberately constrained to a single calendar year (2025) to maintain high-resolution data integrity; thus, the model's resilience across multi-year macroeconomic cycles remains an open question. Model selection relied exclusively on In-Sample data; the Out-Of-Sample period was subsequently used in the post-hoc analyses, which serve a diagnostic purpose rather than as independent validation. Moreover, the ablation study of seed variance highlighted significant dispersion in terminal returns, emphasizing the inherent fragility of neural network convergence and suggesting the necessity of ensemble methods or "seed-averaging" for live deployment. Finally, while the simulation included elevated transaction fees as a simplified proxy for execution costs, the absence of a direct limit order book slippage simulation dictates caution regarding market impact in thin-margin scenarios, demanding strict capacity constraints for real-world capital allocation.

Building upon this hybrid foundation, several concrete next steps emerge for future research. Transitioning from static lookback windows to dynamic state-space models, such as the Kalman Filter, could provide a more responsive estimation of the spread's hidden mean and variance, further reducing the lag in Z-score derivation. Additionally, portfolio construction could be improved by introducing score-weighted capital allocation proportional to the Final Composite Score, or a secondary meta-RL agent tasked with dynamically adjusting pair cardinality based on global market risk indicators. Ultimately, given the agent’s documented sensitivity to informational friction, a comprehensive exploration of alternative state representations and advanced reward architectures will further enhance the efficiency of the proposed execution overlay.

\clearpage
\appendix

\section{Supplementary Figures}
\label{app:sensitivity}

This appendix consolidates the supplementary figures supporting the main analysis. It contains the auxiliary Critic-side training diagnostics for the RL agents (Section \ref{sec:rl_eval}), and the equity-curve figures for the sensitivity analysis and assumptions verification of both the baseline strategy (Section \ref{sec:baseline_eval}) and the RL execution overlay (Section \ref{sec:rl_eval}). The accompanying summary tables remain in the main text.

\subsection*{RL Training Diagnostics (Critic-Side)}

\begin{figure}[H]
    \caption{Training Diagnostic: Value Loss.}
    \label{fig:wandb_value_loss}
    \centering
    \includegraphics[width=\linewidth]{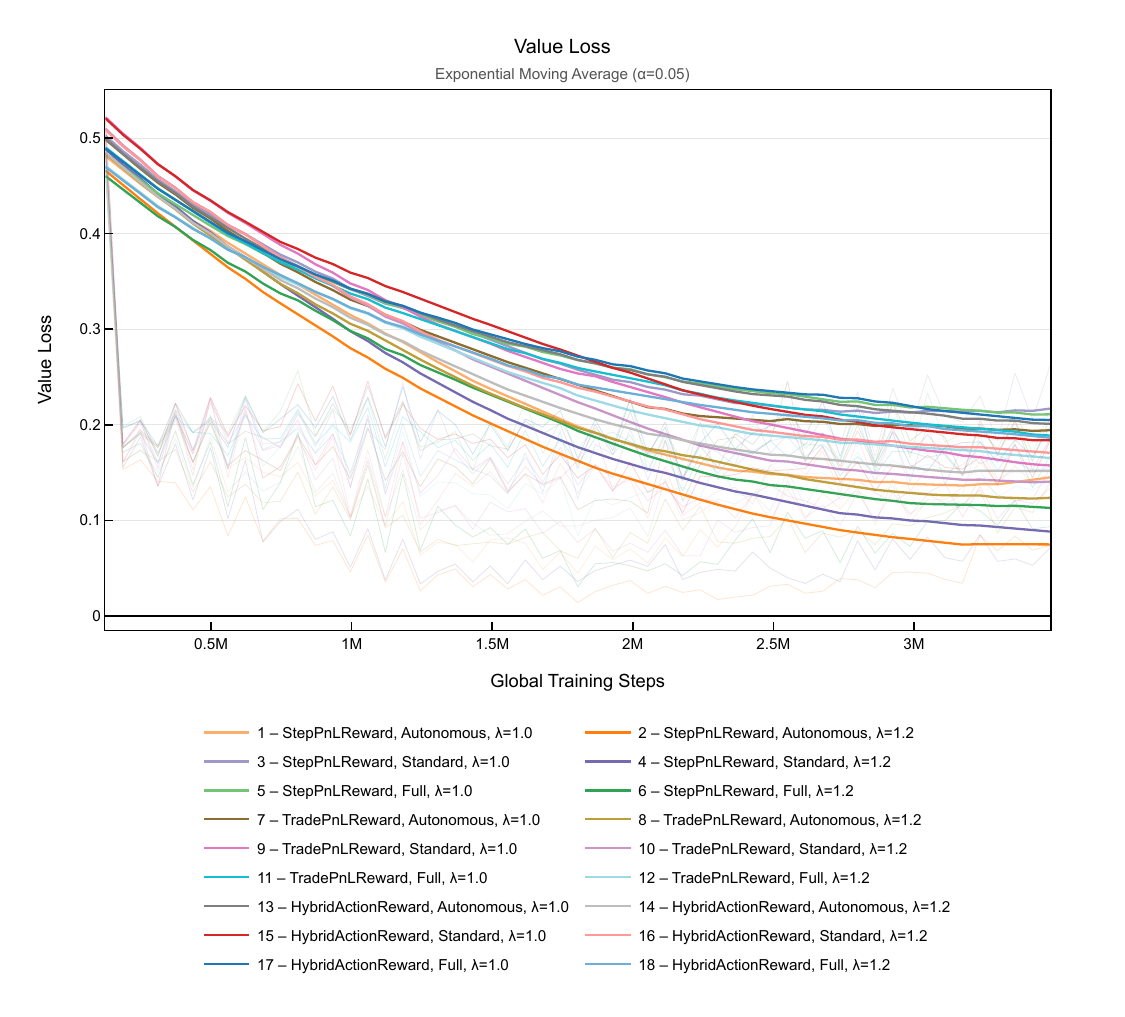}
    \justifying
    \noindent \justifying \noindent \scriptsize Note: 
    The figure illustrates prediction error of the critic network obtained by various RL agent configurations during single 1-month training episodes. Raw current values are displayed in the background, with an Exponential Moving Average ($\alpha=0.05$) overlay highlighting long-term trends. A decreasing and stabilizing loss indicates that the critic is becoming increasingly accurate at estimating the state-value function.
\end{figure}

\begin{figure}[H]
    \caption{Training Diagnostic: Explained Variance.}
    \label{fig:wandb_expl_var}
    \centering
    \includegraphics[width=\linewidth]{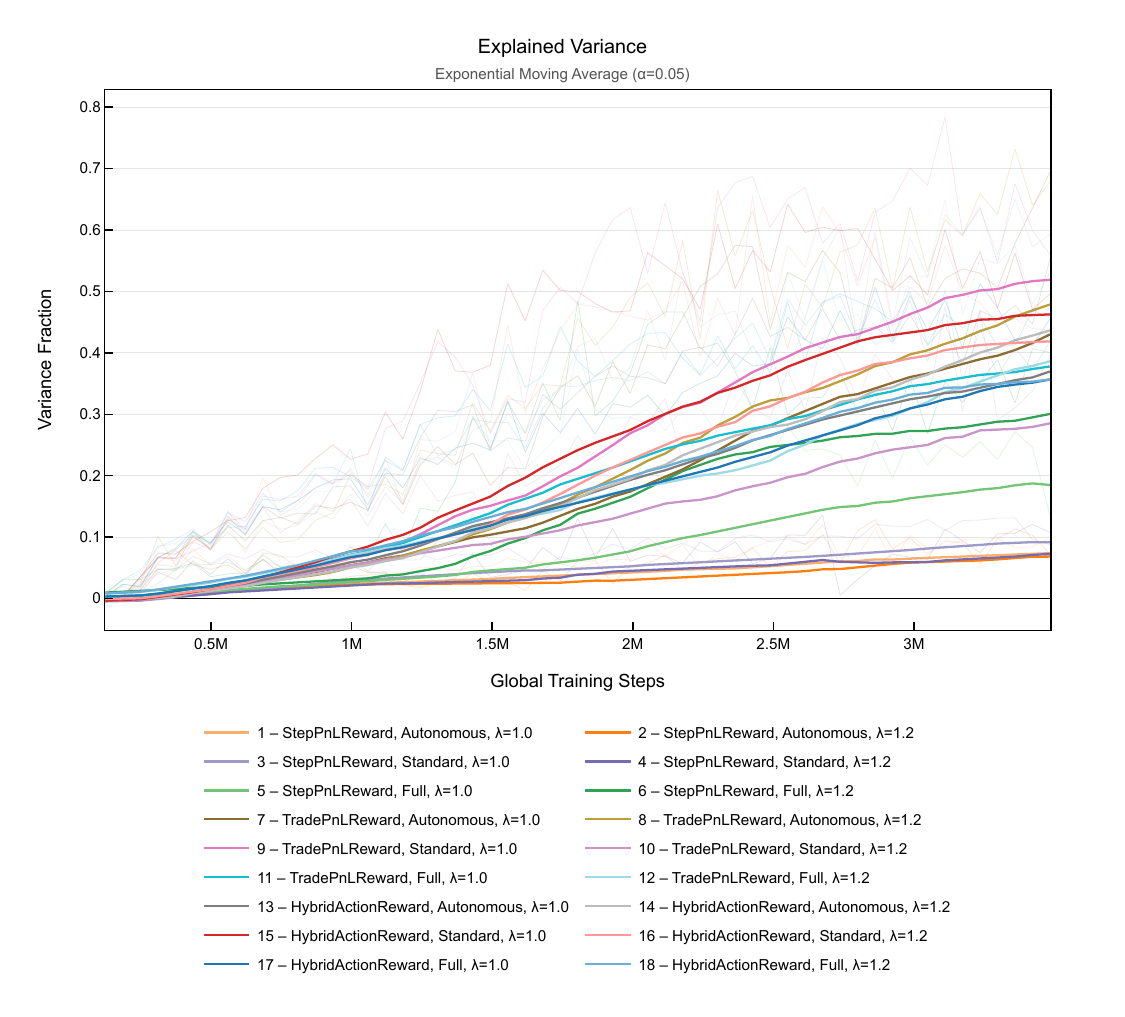}
    \justifying
    \noindent \justifying \noindent \scriptsize Note: 
    The figure illustrates proportion of the variance in rewards correctly predicted by the critic network, obtained by various RL agent configurations during single 1-month training episodes. Raw current values are displayed in the background, with an Exponential Moving Average ($\alpha=0.05$) overlay highlighting long-term trends. A high value indicates that the critic has developed a robust internal representation.
\end{figure}

\subsection*{Baseline Strategy: Sensitivity Analysis}

\begin{figure}[H]
    \caption{Sensitivity Analysis: Entry Threshold Variations.}
    \label{fig:sens_entry}
    \centering
    \includegraphics[width=\linewidth]{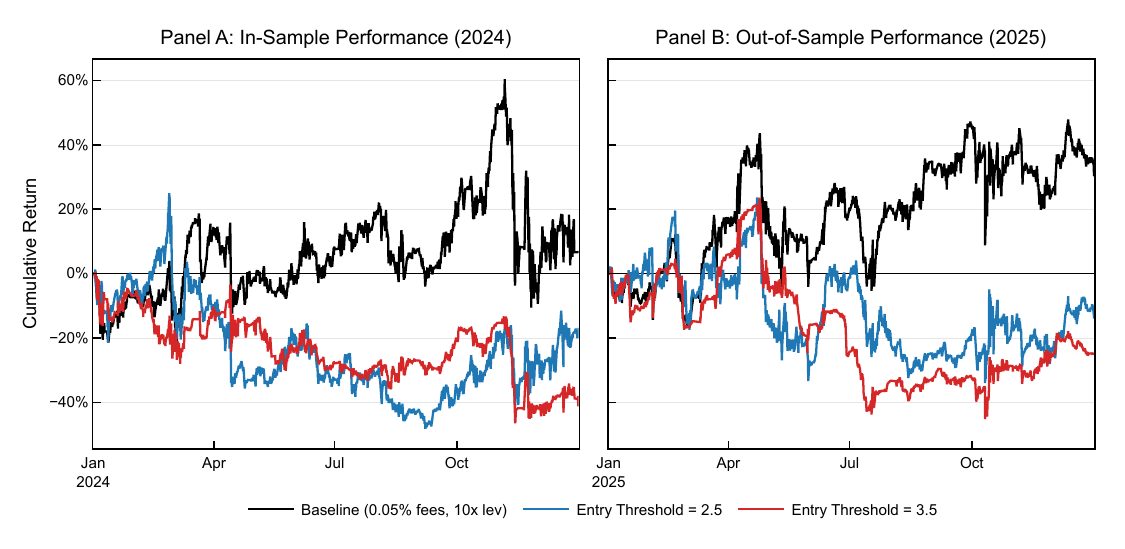}
    \justifying
    \noindent \justifying \noindent \scriptsize Note: 
    The figure illustrates the performance impact of varying the Entry Threshold (2.5 and 3.5) compared to the optimized baseline parameter (3.0). Panel A displays In-Sample performance (2024), and Panel B displays Out-Of-Sample performance (2025). Other parameters remain fixed at baseline defaults. All curves account for 0.05\% transaction fees and 10x leverage.
\end{figure}

\begin{figure}[H]
    \caption{Sensitivity Analysis: Exit Threshold Variations.}
    \label{fig:sens_exit}
    \centering
    \includegraphics[width=\linewidth]{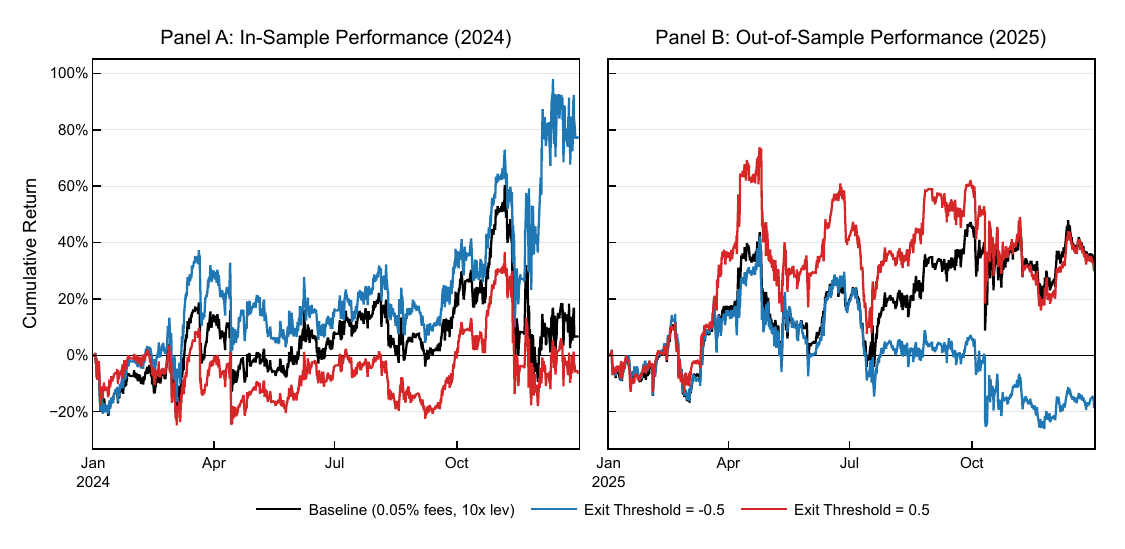}
    \justifying
    \noindent \justifying \noindent \scriptsize Note: 
    The figure illustrates the performance impact of varying the Exit Threshold (-0.5 and 0.5) compared to the default baseline parameter (0.0). Panel A displays In-Sample performance (2024), and Panel B displays Out-Of-Sample performance (2025). Other parameters remain fixed at baseline defaults. All curves account for 0.05\% transaction fees and 10x leverage.
\end{figure}

\begin{figure}[H]
    \caption{Sensitivity Analysis: Stop Loss Variations.}
    \label{fig:sens_sl}
    \centering
    \includegraphics[width=\linewidth]{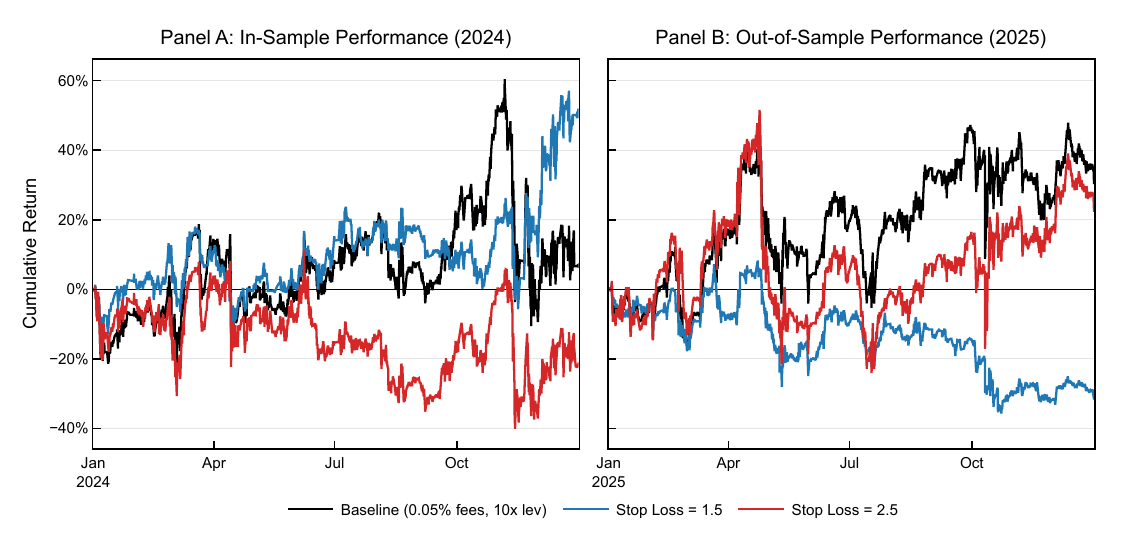}
    \justifying
    \noindent \justifying \noindent \scriptsize Note: 
    The figure illustrates the performance impact of varying the Stop Loss multiplier (1.5 and 2.5) compared to the optimized baseline parameter (2.0). Panel A displays In-Sample performance (2024), and Panel B displays Out-Of-Sample performance (2025). Other parameters remain fixed at baseline defaults. All curves account for 0.05\% transaction fees and 10x leverage.
\end{figure}

\begin{figure}[H]
    \caption{Sensitivity Analysis: Pairs Variations.}
    \label{fig:sens_top_n}
    \centering
    \includegraphics[width=\linewidth]{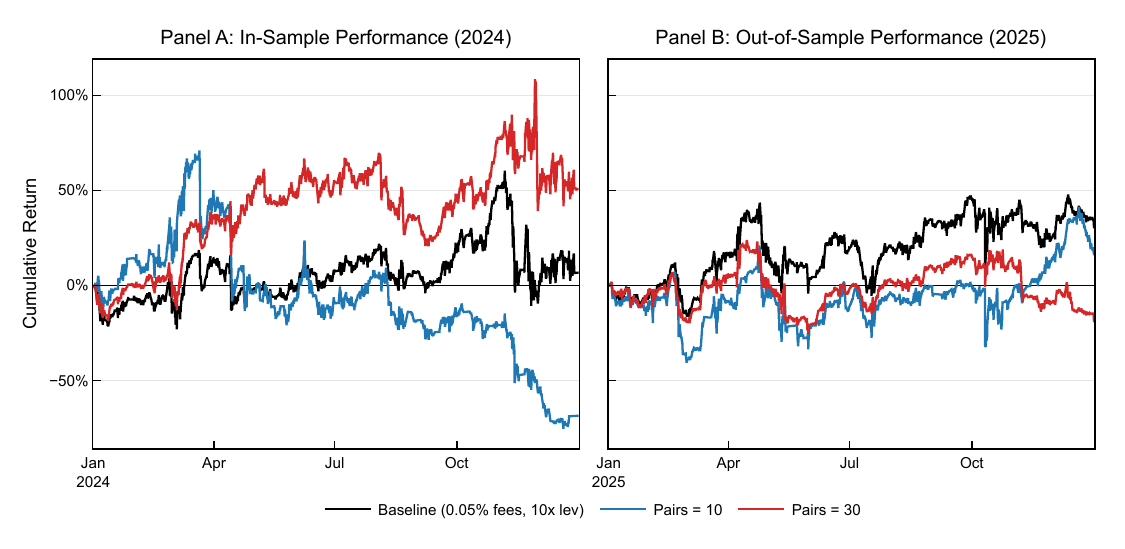}
    \justifying
    \noindent \justifying \noindent \scriptsize Note: 
    The figure illustrates the performance impact of varying the number of pairs (30 and 10) compared to the default baseline parameter (20). Panel A displays In-Sample performance (2024), and Panel B displays Out-Of-Sample performance (2025). Other parameters remain fixed at baseline defaults. All curves account for 0.05\% transaction fees and 10x leverage.
\end{figure}

\begin{figure}[H]
    \caption{Sensitivity Analysis: Z-Score Window Variations.}
    \label{fig:sens_z_score_win}
    \centering
    \includegraphics[width=\linewidth]{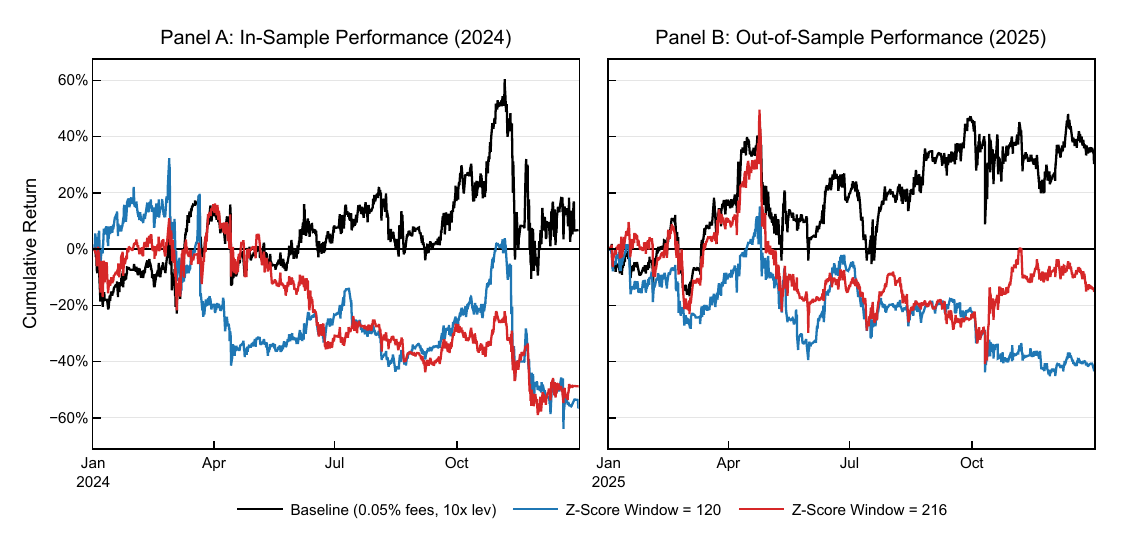}
    \justifying
    \noindent \justifying \noindent \scriptsize Note: 
    The figure illustrates the performance impact of varying the Z-Score Window length (216h and 120h) compared to the default baseline parameter (168h). Panel A displays In-Sample performance (2024), and Panel B displays Out-Of-Sample performance (2025). Other parameters remain fixed at baseline defaults. All curves account for 0.05\% transaction fees and 10x leverage.
\end{figure}

\subsection*{Baseline Strategy: Assumptions Verification}

\begin{figure}[H]
    \caption{Assumptions Verification: Fee Rate Variations.}
    \label{fig:mechanism_fee}
    \centering
    \includegraphics[width=\linewidth]{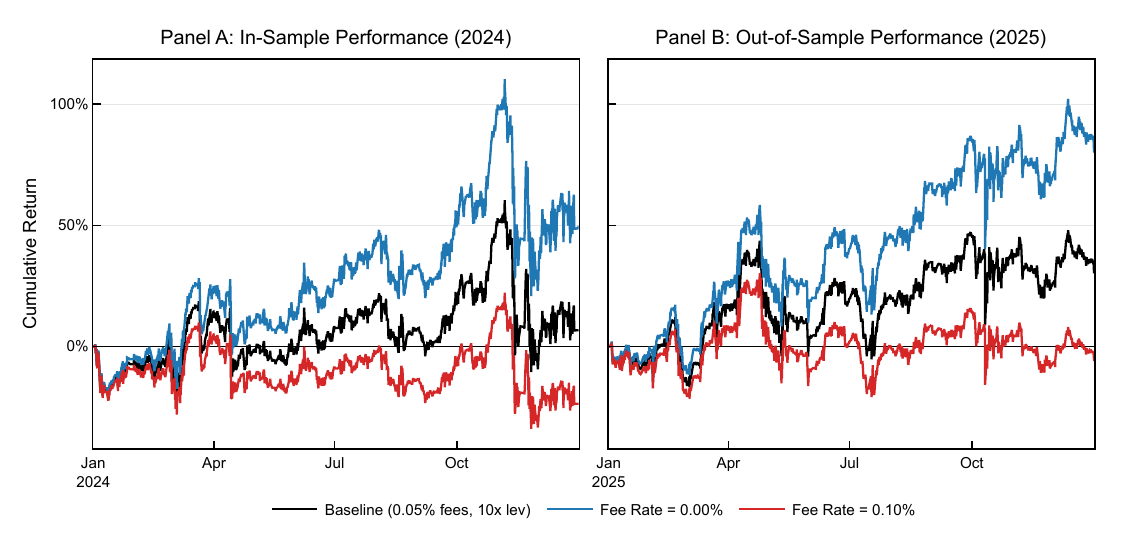}
    \justifying
    \noindent \justifying \noindent \scriptsize Note: 
    The figure illustrates the performance impact of varying the Fee Rate (0.00\% and 0.10\%) compared to the default baseline parameter (0.05\%). Panel A displays In-Sample performance (2024), and Panel B displays Out-Of-Sample performance (2025). Other parameters remain fixed at baseline defaults. All curves account for 0.05\% transaction fees and 10x leverage.
\end{figure}

\begin{figure}[H]
    \caption{Assumptions Verification: Beta Hedge disabled.}
    \label{fig:mechanism_beta_hedge}
    \centering
    \includegraphics[width=\linewidth]{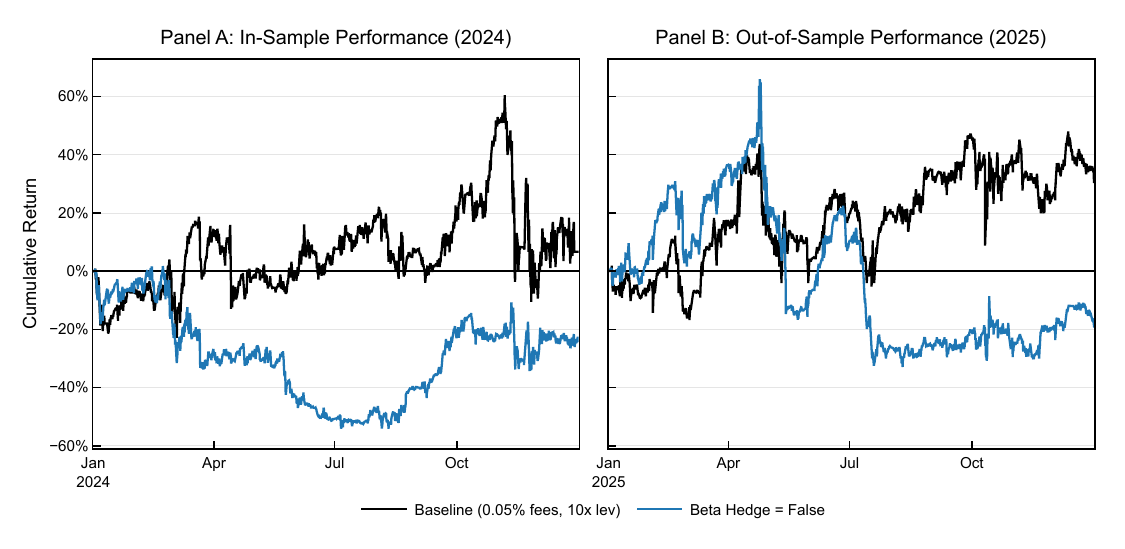}
    \justifying
    \noindent \justifying \noindent \scriptsize Note: 
    The figure illustrates the performance impact of disabling the beta hedge (fixed $\beta=1.0$) compared to the baseline. Panel A displays In-Sample performance (2024), and Panel B displays Out-Of-Sample performance (2025). Other parameters remain fixed at baseline defaults. All curves account for 0.05\% transaction fees and 10x leverage.
\end{figure}

\begin{figure}[H]
    \caption{Assumptions Verification: SL Lock disabled.}
    \label{fig:mechanism_sl_lock}
    \centering
    \includegraphics[width=\linewidth]{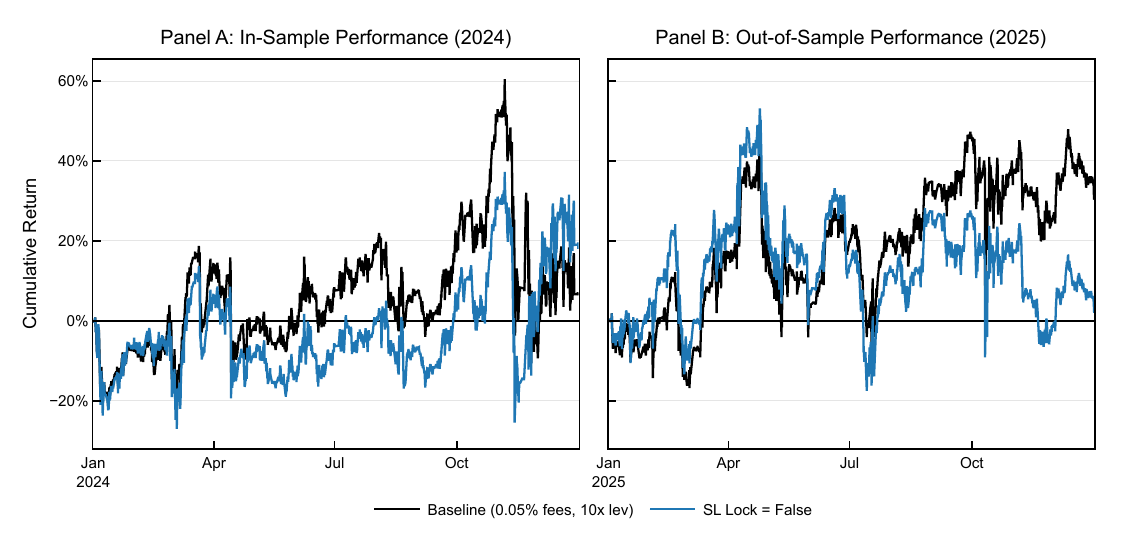}
    \justifying
    \noindent \justifying \noindent \scriptsize Note: 
    The figure illustrates the performance impact of disabling the SL Lock mechanism compared to the baseline. Panel A displays In-Sample performance (2024), and Panel B displays Out-Of-Sample performance (2025). Other parameters remain fixed at baseline defaults. All curves account for 0.05\% transaction fees and 10x leverage.
\end{figure}

\begin{figure}[H]
    \caption{Assumptions Verification: Time Decay SL disabled.}
    \label{fig:mechanism_time_decay_sl}
    \centering
    \includegraphics[width=\linewidth]{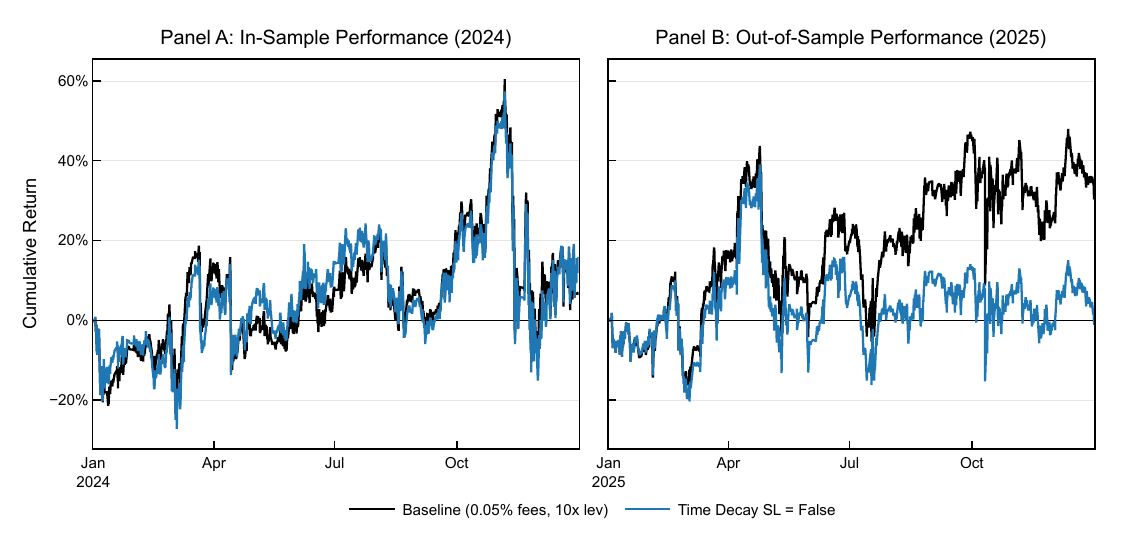}
    \justifying
    \noindent \justifying \noindent \scriptsize Note: 
    The figure illustrates the performance impact of disabling the Time Decay SL mechanism compared to the baseline. Panel A displays In-Sample performance (2024), and Panel B displays Out-Of-Sample performance (2025). Other parameters remain fixed at baseline defaults. All curves account for 0.05\% transaction fees and 10x leverage.
\end{figure}

\subsection*{RL Strategy: Sensitivity Analysis}

\begin{figure}[H]
    \caption{Sensitivity Analysis: Entry Threshold Variations (Agent 2).}
    \label{fig:rl_sens_entry}
    \centering
    \includegraphics[width=\linewidth]{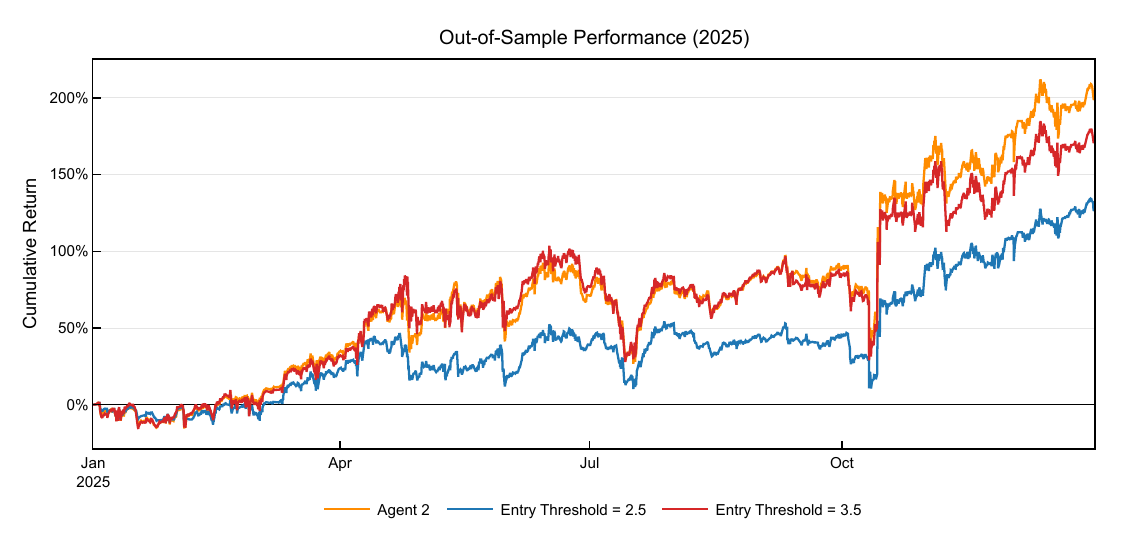}
    \justifying
    \noindent \justifying \noindent \scriptsize Note: The figure illustrates the Out-Of-Sample performance impact of varying the Entry Threshold (2.5 and 3.5) compared to the optimized baseline parameter (3.0). Agent 2 configuration: StepPnLReward, Autonomous Space, $\lambda=1.2$. Other parameters remain fixed at baseline defaults. All curves account for 0.05\% transaction fees and 10x leverage.
\end{figure}

\begin{figure}[H]
    \caption{Sensitivity Analysis: Exit Threshold Variations (Agent 2).}
    \label{fig:rl_sens_exit}
    \centering
    \includegraphics[width=\linewidth]{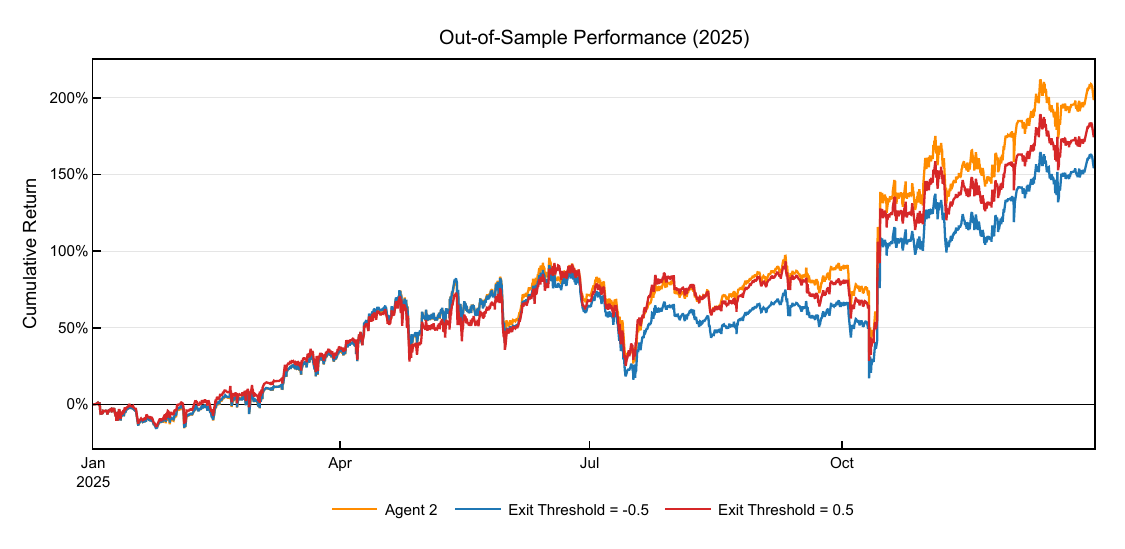}
    \justifying
    \noindent \justifying \noindent \scriptsize Note: The figure illustrates the Out-Of-Sample performance impact of varying the Exit Threshold (-0.5 and 0.5) compared to the default baseline parameter (0.0). Agent 2 configuration: StepPnLReward, Autonomous Space, $\lambda=1.2$. Other parameters remain fixed at baseline defaults. All curves account for 0.05\% transaction fees and 10x leverage.
\end{figure}

\begin{figure}[H]
    \caption{Sensitivity Analysis: Stop Loss Variations (Agent 2).}
    \label{fig:rl_sens_sl}
    \centering
    \includegraphics[width=\linewidth]{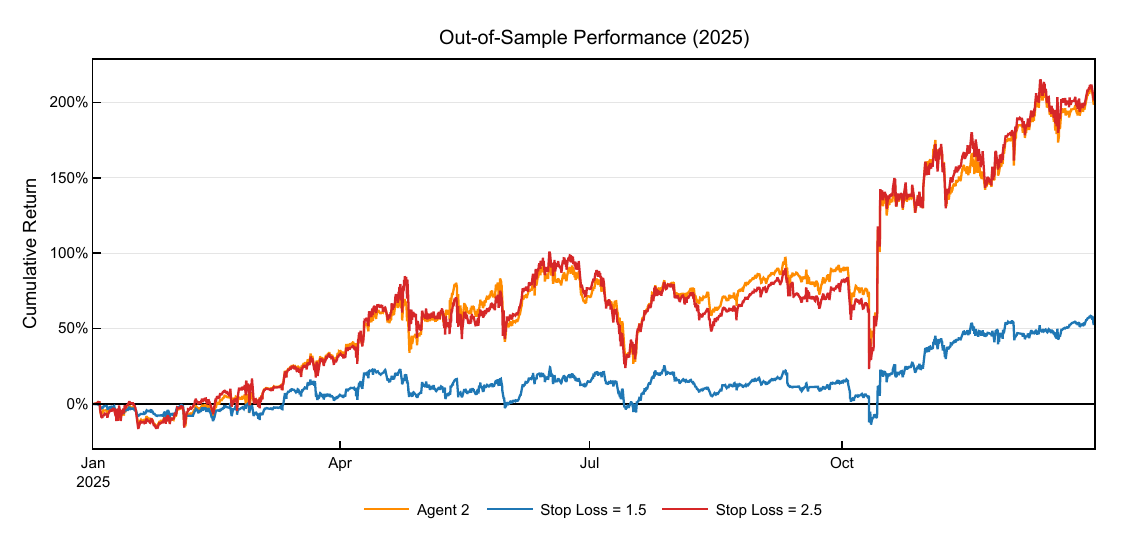}
    \justifying
    \noindent \justifying \noindent \scriptsize Note: The figure illustrates the Out-Of-Sample performance impact of varying the Stop Loss multiplier (2.5 and 1.5) compared to the optimized baseline parameter (2.0). Agent 2 configuration: StepPnLReward, Autonomous Space, $\lambda=1.2$. Other parameters remain fixed at baseline defaults. All curves account for 0.05\% transaction fees and 10x leverage.
\end{figure}

\begin{figure}[H]
    \caption{Sensitivity Analysis: Z-Score Window Variations (Agent 2).}
    \label{fig:rl_sens_z_score}
    \centering
    \includegraphics[width=\linewidth]{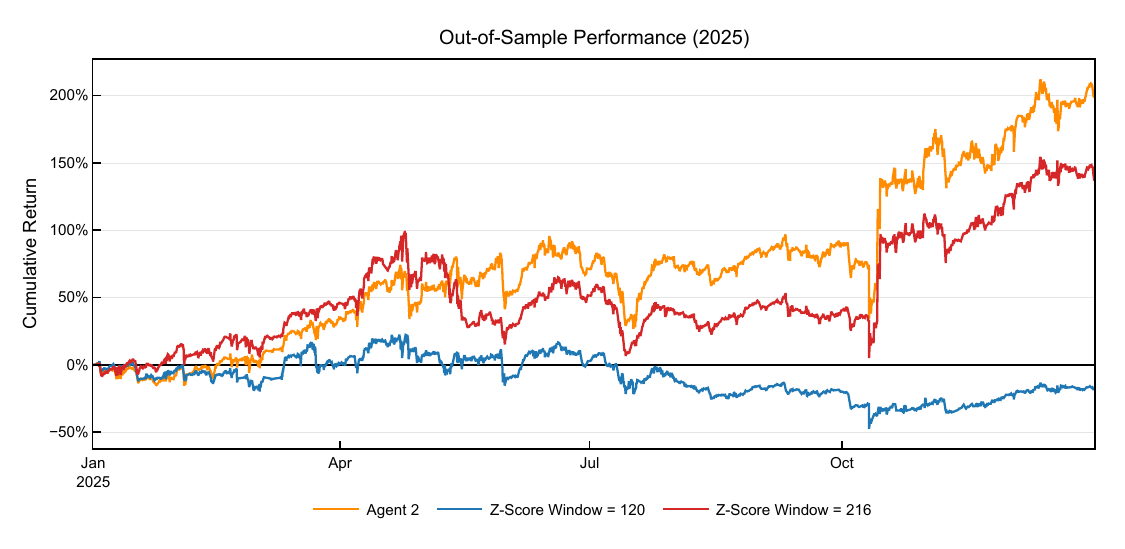}
    \justifying
    \noindent \justifying \noindent \scriptsize Note: The figure illustrates the Out-Of-Sample performance impact of varying the Z-Score Window length (216h and 120h) compared to the default baseline parameter (168h). Agent 2 configuration: StepPnLReward, Autonomous Space, $\lambda=1.2$. Other parameters remain fixed at baseline defaults. All curves account for 0.05\% transaction fees and 10x leverage.
\end{figure}

\begin{figure}[H]
    \caption{Sensitivity Analysis: Pairs Variations (Agent 2).}
    \label{fig:rl_sens_top_n}
    \centering
    \includegraphics[width=\linewidth]{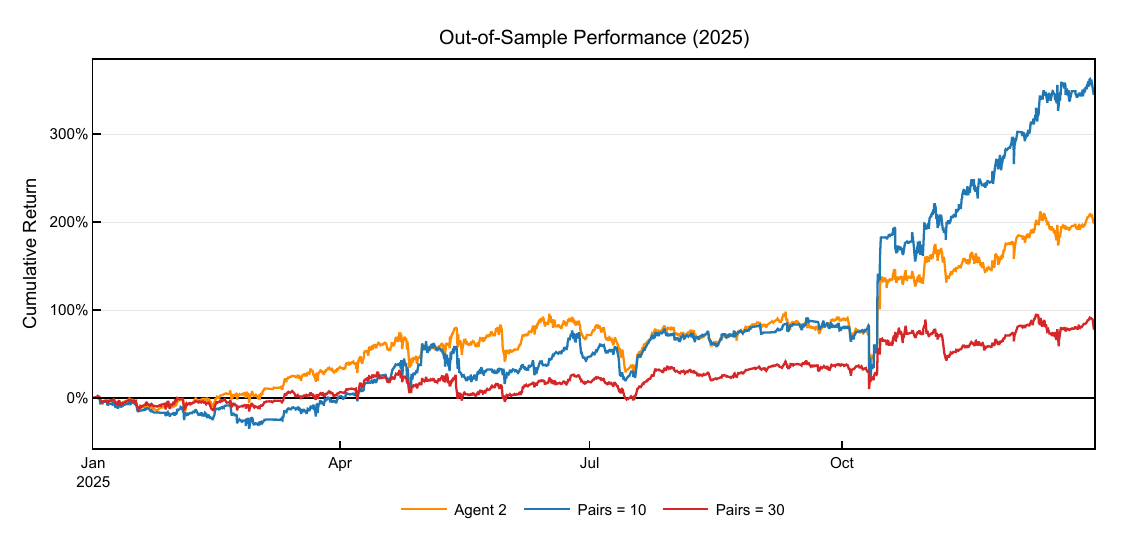}
    \justifying
    \noindent \justifying \noindent \scriptsize Note: The figure illustrates the Out-Of-Sample performance impact of varying the number of active pairs (30 and 10) compared to the default baseline parameter (20). Agent 2 configuration: StepPnLReward, Autonomous Space, $\lambda=1.2$. Other parameters remain fixed at baseline defaults. All curves account for 0.05\% transaction fees and 10x leverage.
\end{figure}

\subsection*{RL Strategy: Assumptions Verification}

\begin{figure}[H]
    \caption{Assumptions Verification: Fee Rate Variations (Agent 2).}
    \label{fig:rl_mech_fee}
    \centering
    \includegraphics[width=\linewidth]{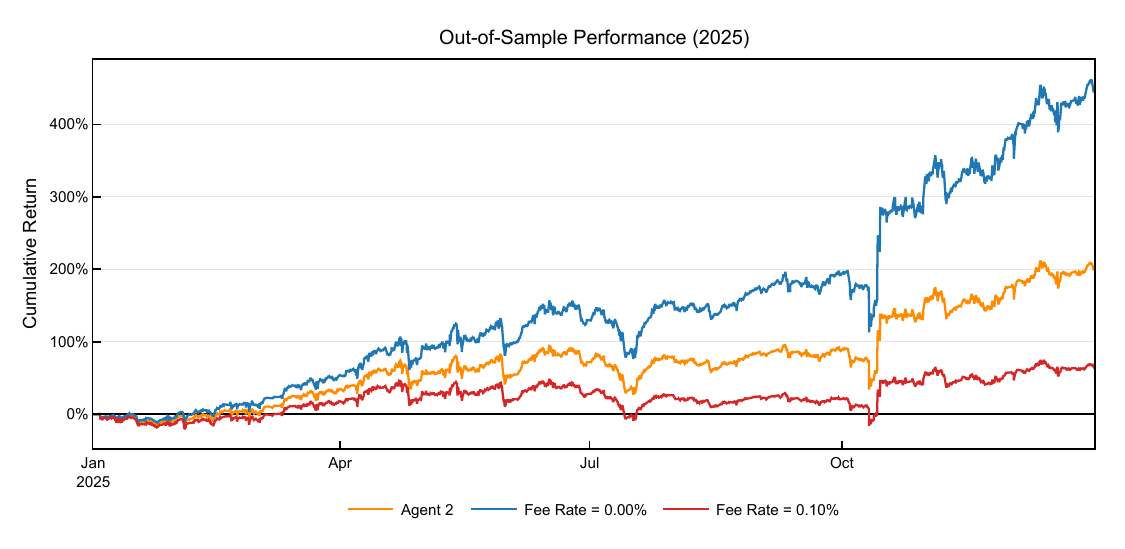}
    \justifying
    \noindent \justifying \noindent \scriptsize Note: The figure illustrates the Out-Of-Sample performance impact of varying the transaction Fee Rate (0.00\% and 0.10\%) compared to the default parameter (0.05\%). Agent 2 configuration: StepPnLReward, Autonomous Space, $\lambda=1.2$. Other parameters remain fixed at baseline defaults. All curves account for 0.05\% transaction fees and 10x leverage.
\end{figure}

\begin{figure}[H]
    \caption{Assumptions Verification: Beta Hedge Disabled (Agent 2).}
    \label{fig:rl_mech_beta}
    \centering
    \includegraphics[width=\linewidth]{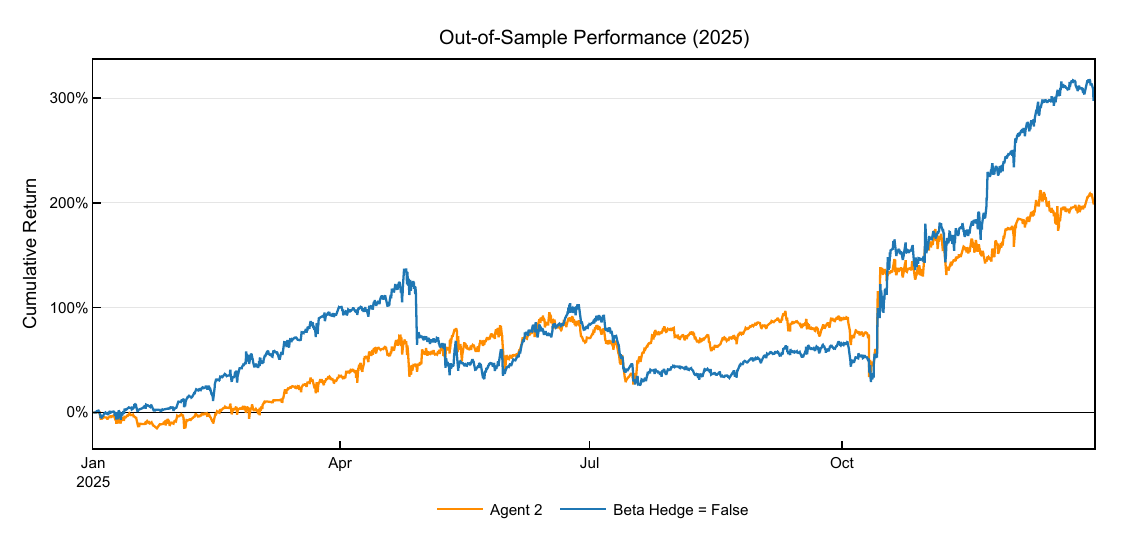}
    \justifying
    \noindent \justifying \noindent \scriptsize Note: The figure illustrates the Out-Of-Sample performance impact of disabling the beta hedge (reverting to a fixed $\beta=1.0$ allocation) compared to the default dynamic empirical hedge. Agent 2 configuration: StepPnLReward, Autonomous Space, $\lambda=1.2$. Other parameters remain fixed at baseline defaults. All curves account for 0.05\% transaction fees and 10x leverage.
\end{figure}

\begin{figure}[H]
    \caption{Assumptions Verification: SL Lock Disabled (Agent 2).}
    \label{fig:rl_mech_sl_lock}
    \centering
    \includegraphics[width=\linewidth]{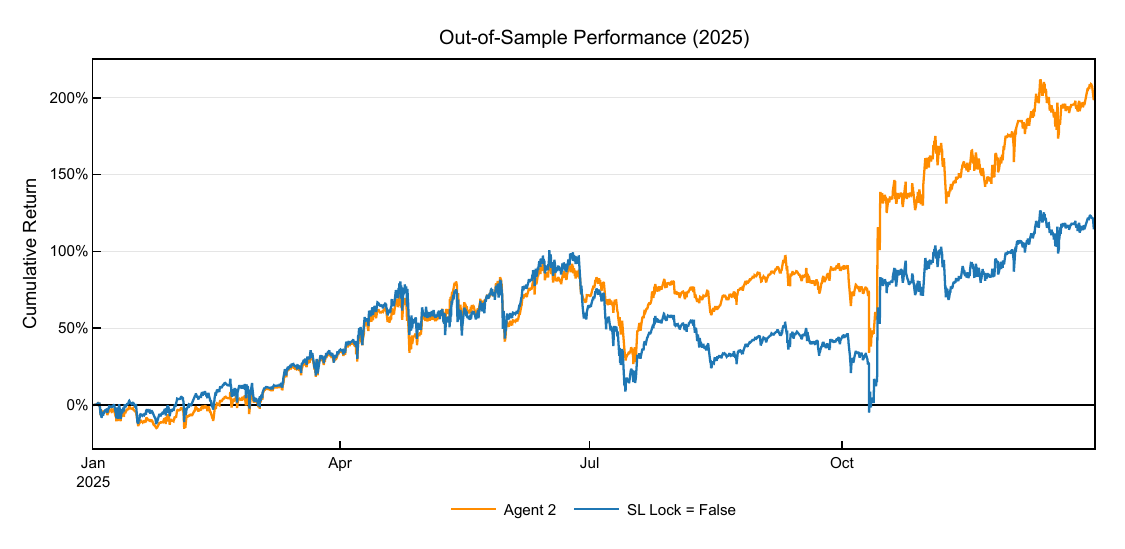}
    \justifying
    \noindent \justifying \noindent \scriptsize Note: The figure illustrates the Out-Of-Sample performance impact of disabling the Stop Loss Lock regime filter compared to the fully constrained strategy. Agent 2 configuration: StepPnLReward, Autonomous Space, $\lambda=1.2$. Other parameters remain fixed at baseline defaults. All curves account for 0.05\% transaction fees and 10x leverage.
\end{figure}

\begin{figure}[H]
    \caption{Assumptions Verification: Time Decay SL Disabled (Agent 2).}
    \label{fig:rl_mech_time_decay}
    \centering
    \includegraphics[width=\linewidth]{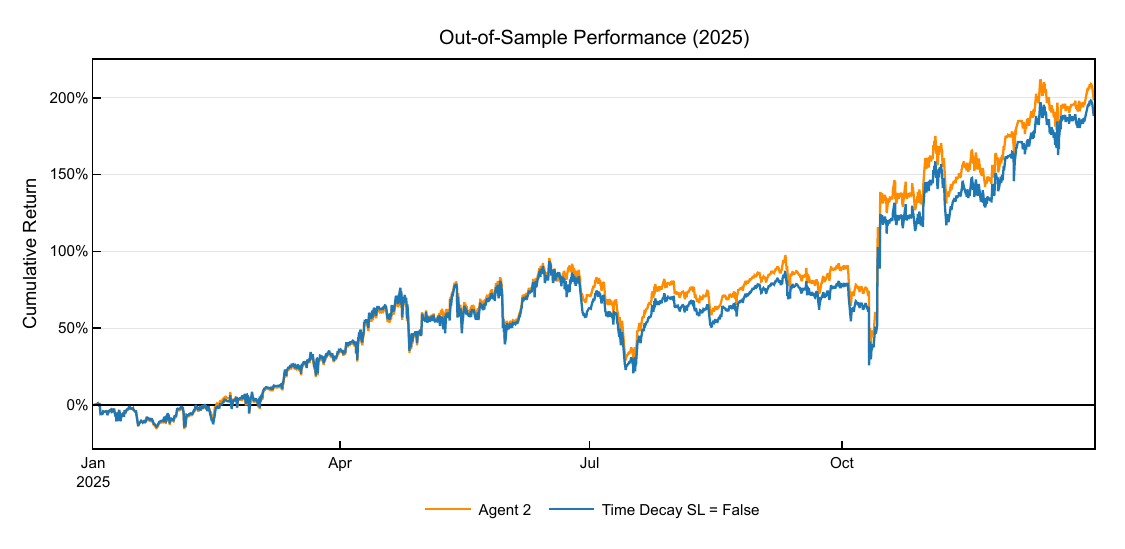}
    \justifying
    \noindent \justifying \noindent \scriptsize Note: The figure illustrates the Out-Of-Sample performance impact of disabling the temporal tightening of the Stop Loss limit compared to the fully constrained strategy. Agent 2 configuration: StepPnLReward, Autonomous Space, $\lambda=1.2$. Other parameters remain fixed at baseline defaults. All curves account for 0.05\% transaction fees and 10x leverage.
\end{figure}

\begin{figure}[H]
    \caption{Assumptions Verification: Risk Management Overlay Disabled (Agent 2).}
    \label{fig:rl_mech_auto}
    \centering
    \includegraphics[width=\linewidth]{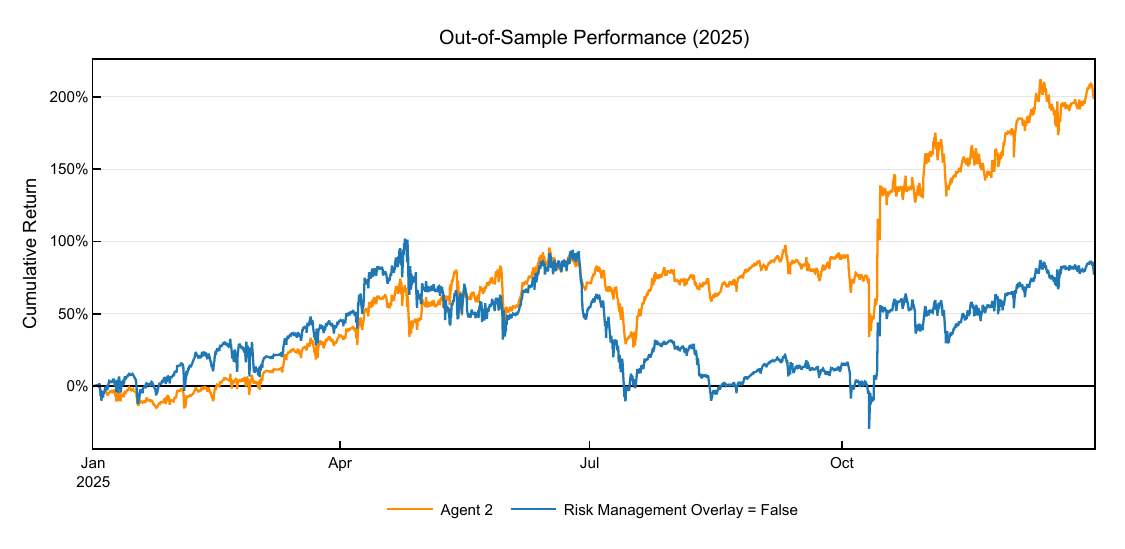}
    \justifying
    \noindent \justifying \noindent \scriptsize Note: The figure demonstrates the Out-Of-Sample performance degradation resulting from disabling the heuristic Take-Profit and Stop Loss enforcement and the associated SL Lock; the time-based close at the end of the Z-Score Window and the margin-exhaustion liquidation remain in place. Agent 2 configuration: StepPnLReward, Autonomous Space, $\lambda=1.2$. Other parameters remain fixed at baseline defaults. All curves account for 0.05\% transaction fees and 10x leverage.
\end{figure}

\begin{figure}[H]
    \caption{Assumptions Verification: Out-of-Sample Performance Stability across 5 Seeds (Agent 2).}
    \label{fig:seeds_var_oos}
    \centering
    \includegraphics[width=\linewidth]{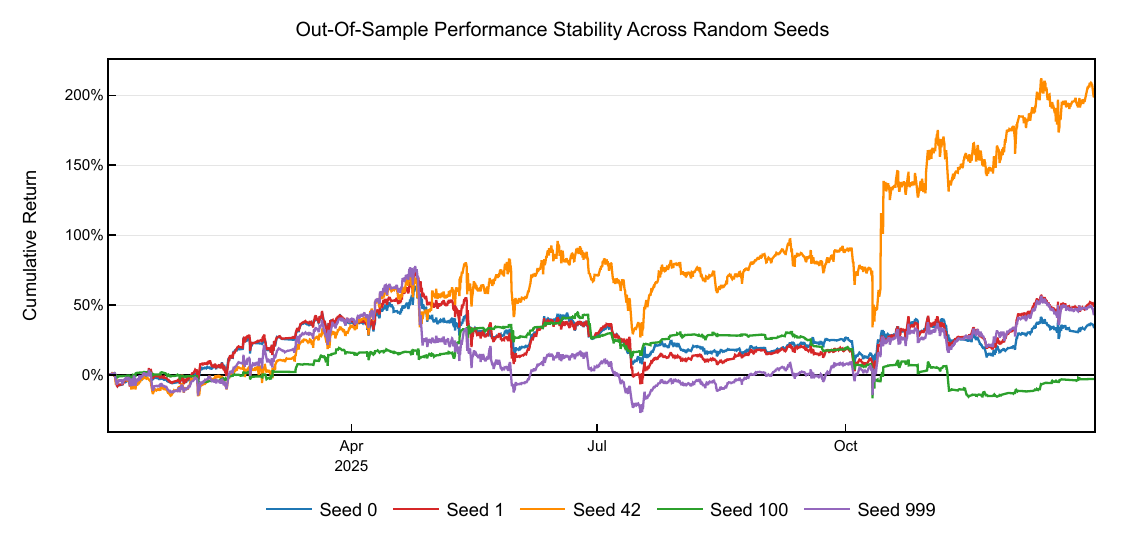}
    \justifying\noindent\scriptsize Note: The plot presents the cumulative returns for Agent 2 evaluated across five independent random seeds during the Out-of-Sample period (2025). All results account for 0.05\% fee rates and a leverage of 10x. Seed 42 represents the original baseline configuration.
\end{figure}

\clearpage
\bibliographystyle{elsarticle-harv}
\bibliography{references}

@article{Gatev2006,
    author = {Gatev, Evan and Goetzmann, William N. and Rouwenhorst, K. Geert},
    title = {Pairs Trading: Performance of a Relative-Value Arbitrage Rule},
    journal = {The Review of Financial Studies},
    volume = {19},
    number = {3},
    pages = {797-827},
    year = {2006},
    month = {10},
    url = {https://doi.org/10.1093/rfs/hhj020},
}

@article{Do2010,
    author = {Binh Do and Robert Faff},
    title = {Does Simple Pairs Trading Still Work?},
    journal = {Financial Analysts Journal},
    volume = {66},
    number = {4},
    pages = {83--95},
    year = {2010},
    publisher = {Routledge},
    url = {https://doi.org/10.2469/faj.v66.n4.1},
}

@article{Engle1987,
  author = {Robert F. Engle and C. W. J. Granger},
  journal = {Econometrica},
  number = {2},
  pages = {251--276},
  publisher = {[Wiley, Econometric Society]},
  title = {Co-Integration and Error Correction: Representation, Estimation, and Testing},
  volume = {55},
  year = {1987},
  url = {https://doi.org/10.2307/1913236},
}

@article{Hurst1951,
  title={Long-term storage capacity of reservoirs},
  author={Hurst, Harold E},
  journal={Transactions of the American Society of Civil Engineers},
  volume={116},
  number={1},
  pages={770--799},
  year={1951},
  publisher={American Society of Civil Engineers},
  url = {https://doi.org/10.1061/TACEAT.0006518}
}

@article{Qian2007,
  author={Qian, Bo and Rasheed, Khaled},
  title={Stock market prediction with multiple classifiers},
  journal={Applied Intelligence},
  year={2007},
  date={2007-02-01},
  volume={26},
  number={1},
  pages={25--33},
  url={https://doi.org/10.1007/s10489-006-0001-7},
}

@article{Krauss2017,
  title={Statistical arbitrage pairs trading strategies: Review and outlook},
  author={Krauss, Christopher},
  journal={Journal of Economic Surveys},
  volume={31},
  number={2},
  pages={513--545},
  year={2017},
  publisher={Wiley Online Library},
  url={https://doi.org/10.1111/joes.12153}
}

@Article{lin2024optimal,
AUTHOR = {Lin, Hsio-Yi and ChiangLin, Chieh-Yow and Tseng, Hsuan-Wei},
TITLE = {Optimal Parameter Selection and Indicator Design for Technical Analysis Strategies by Computer Software: An Empirical Analysis of the Taiwan Futures Market},
JOURNAL = {Engineering Proceedings},
VOLUME = {74},
YEAR = {2024},
NUMBER = {1},
ARTICLE-NUMBER = {56},
URL = {https://doi.org/10.3390/engproc2024074056},
}

@misc{zeng2025regimefolio,
      title={RegimeFolio: A Regime Aware ML System for Sectoral Portfolio Optimization in Dynamic Markets}, 
      author={Yiyao Zhang and Diksha Goel and Hussain Ahmad and Claudia Szabo},
      year={2025},
      eprint={2510.14986},
      archivePrefix={arXiv},
      primaryClass={q-fin.PM},
      url={https://doi.org/10.48550/arXiv.2510.14986},
}

@article{kahneman1979,
  URL = {http://doi.org/10.2307/1914185},
  author = {Daniel Kahneman and Amos Tversky},
  journal = {Econometrica},
  number = {2},
  pages = {263--291},
  publisher = {[Wiley, Econometric Society]},
  title = {Prospect Theory: An Analysis of Decision under Risk},
  volume = {47},
  year = {1979}
}

@misc{schulman2017ppo,
  title={Proximal Policy Optimization Algorithms},
  author={Schulman, John and Wolski, Filip and Dhariwal, Prafulla and Radford, Alec and Klimov, Oleg},
  year={2017},
  eprint={1707.06347},
  archivePrefix={arXiv},
  primaryClass={cs.LG},
  url={https://doi.org/10.48550/arXiv.1707.06347},
}

@article{kim2019,
  title={Optimizing the Pairs-Trading Strategy Using Deep Reinforcement Learning with Trading and Stop-Loss Boundaries},
  author={Kim, Taewook and Kim, Ha Young},
  journal={Complexity},
  volume={2019},
  year={2019},
  publisher={Hindawi},
  url={https://doi.org/10.1155/2019/3582516}
}

@article{yang2024,
  title={Reinforcement Learning Pair Trading: A Dynamic Scaling Approach},
  volume={17},
  url={http://doi.org/10.3390/jrfm17120555},
  number={12},
  journal={Journal of Risk and Financial Management},
  publisher={MDPI AG},
  author={Yang, Hongshen and Malik, Avinash},
  year={2024},
  pages={555}
}

@misc{jiang2017cryptocurrency,
      title={A Deep Reinforcement Learning Framework for the Financial Portfolio Management Problem}, 
      author={Zhengyao Jiang and Dixing Xu and Jinjun Liang},
      year={2017},
      eprint={1706.10059},
      archivePrefix={arXiv},
      primaryClass={q-fin.CP},
      url={https://arxiv.org/abs/1706.10059}, 
}

@article{garcia2015comprehensive,
  author  = {Javier Garc{{\'i}}a and Fern and o Fern{{\'a}}ndez},
  title   = {A Comprehensive Survey on Safe Reinforcement Learning},
  journal = {Journal of Machine Learning Research},
  year    = {2015},
  volume  = {16},
  number  = {42},
  pages   = {1437-1480},
  url     = {http://jmlr.org/papers/v16/garcia15a.html}
}

@article{alshiekh2018safe, 
    title={Safe Reinforcement Learning via Shielding}, 
    volume={32}, 
    url={https://doi.org/10.1609/aaai.v32i1.11797}, 
    number={1}, 
    journal={Proceedings of the AAAI Conference on Artificial Intelligence}, 
    author={Alshiekh, Mohammed and Bloem, Roderick and Ehlers, Rüdiger and Könighofer, Bettina and Niekum, Scott and Topcu, Ufuk}, 
    year={2018}, 
}

@book{vidyamurthy2004pairs,
  title={Pairs Trading: Quantitative Methods and Analysis},
  author={Vidyamurthy, Ganapathy},
  year={2004},
  publisher={John Wiley \& Sons},
}

@misc{pardo2018time,
    title={Time Limits in Reinforcement Learning}, 
    author={Fabio Pardo and Arash Tavakoli and Vitaly Levdik and Petar Kormushev},
    year={2022},
    eprint={1712.00378},
    archivePrefix={arXiv},
    primaryClass={cs.LG},
    url={10.48550/arXiv.1712.00378}
}

@article{dulac2019challenges,
  author       = {Dulac-Arnold, Gabriel and Levine, Nir and Mankowitz, Daniel J. and Li, Jerry and Paduraru, Cosmin and Gowal, Sven and Hester, Todd},
  title        = {Challenges of real-world reinforcement learning: definitions, benchmarks and analysis},
  journal = {Machine Learning},
  year         = {2021},
  volume       = {110},
  number       = {9},
  pages        = {2419--2468},
  url          = {https://doi.org/10.1007/s10994-021-05961-4},
}

@book{lopez2018advances,
  author    = {L{\'o}pez de Prado, Marcos},
  title     = {Advances in Financial Machine Learning},
  publisher = {John Wiley \& Sons},
  year      = {2018},
  url={https://books.google.pl/books?id=v0RKDwAAQBAJ},
}

@article{ramos2017hurst,
title = {Introducing Hurst exponent in pair trading},
journal = {Physica A: Statistical Mechanics and its Applications},
volume = {488},
pages = {39-45},
year = {2017},
issn = {0378-4371},
url = {https://doi.org/10.1016/j.physa.2017.06.032},
author = {J.P. Ramos-Requena and J.E. Trinidad-Segovia and M.A. Sánchez-Granero},
}

@Article{fischer2019,
  author={Fischer, Thomas Günter and Krauss, Christopher and Deinert, Alexander},
  title ={Statistical Arbitrage in Cryptocurrency Markets},
  journal={Journal of Risk and Financial Management},
  volume={12},
  year={2019},
  number={1},
  article-number={31},
  url={https:/doi.org/10.3390/jrfm12010031},
}

@article{Ardia2019,
title = {Regime changes in Bitcoin GARCH volatility dynamics},
journal = {Finance Research Letters},
volume = {29},
pages = {266-271},
year = {2019},
url = {https://doi.org/10.1016/j.frl.2018.08.009},
author = {David Ardia and Keven Bluteau and Maxime Rüede},
}

@article{hochreiter1997,
  author = {Hochreiter, Sepp and Schmidhuber, J\"{u}rgen},
  title = {Long Short-Term Memory},
  year = {1997},
  publisher = {MIT Press},
  volume = {9},
  number = {8},
  url = {https://doi.org/10.1162/neco.1997.9.8.1735},
  journal = {Neural Computation},
  pages = {1735--1780}
}

@book{sutton2018,
  title={Reinforcement learning: An introduction},
  author={Sutton, Richard S and Barto, Andrew G},
  edition={Second},
  year={2018},
  publisher={The MIT Press},
  address={Cambridge, Massachusetts},
  url={http://incompleteideas.net/book/the-book-2nd.html}
}

@article{bailey2014,
  author = {Bailey, David and Borwein, Jonathan (Jon) and Lopez de Prado, Marcos and Zhu, Qiji},
  year = {2014},
  pages = {458},
  title = {Pseudo-Mathematics and Financial Charlatanism: The Effects of Backtest Overfitting on Out-of-Sample Performance},
  volume = {61},
  journal = {Notices of the American Mathematical Society},
  url = {https://doi.org/10.1090/noti1105}
}

@article{avellaneda2010,
  author={Marco Avellaneda and Jeong-Hyun Lee},
  title={Statistical arbitrage in the US equities market},
  journal={Quantitative Finance},
  volume={10},
  number={7},
  pages={761--782},
  year={2010},
  publisher={Routledge},
  URL={https://doi.org/10.1080/14697680903124632}
}

@article{mihatsch2002,
  author = {Mihatsch, Oliver and Neuneier, Ralph},
  year = {2002},
  pages = {267-290},
  title = {Risk-Sensitive Reinforcement Learning},
  volume = {49},
  journal = {Machine Learning},
  url = {https://doi.org/10.1023/A:1017940631555}
}

@article{makarov2020,
  title = {Trading and arbitrage in cryptocurrency markets},
  journal = {Journal of Financial Economics},
  volume = {135},
  number = {2},
  pages = {293-319},
  year = {2020},
  url = {https://doi.org/10.1016/j.jfineco.2019.07.001},
  author = {Igor Makarov and Antoinette Schoar}
}

@misc{liu2020,
      title={FinRL: A Deep Reinforcement Learning Library for Automated Stock Trading in Quantitative Finance}, 
      author={Xiao-Yang Liu and Hongyang Yang and Qian Chen and Runjia Zhang and Liuqing Yang and Bowen Xiao and Christina Dan Wang},
      year={2022},
      eprint={2011.09607},
      archivePrefix={arXiv},
      primaryClass={q-fin.TR},
      url={https://arxiv.org/abs/2011.09607}, 
}

@misc{guijarro2021,
      title={Deep Learning Statistical Arbitrage}, 
      author={Jorge Guijarro-Ordonez and Markus Pelger and Greg Zanotti},
      year={2022},
      eprint={2106.04028},
      archivePrefix={arXiv},
      primaryClass={cs.LG},
      url={https://arxiv.org/abs/2106.04028}, 
}

@article{sharpe1966,
 URL = {http://doi.org/10.1086/294846},
 author = {William F. Sharpe},
 journal = {The Journal of Business},
 number = {1},
 pages = {119--138},
 publisher = {University of Chicago Press},
 title = {Mutual Fund Performance},
 volume = {39},
 year = {1966}
}

@article{sortino1994,
  title={Performance Measurement in a Downside Risk Framework},
  author={Sortino, Frank A and Price, Lee N},
  journal={The Journal of Investing},
  volume={3},
  number={3},
  pages={59--64},
  year={1994},
  url={https://doi.org/10.3905/joi.3.3.59}
}

@article{magdon2004,
 URL = {http://doi.org/10.1239/jap/1077134674},
 author = {Malik Magdon-Ismail and Amir F. Atiya and Amrit Pratap and Yaser S. Abu-Mostafa},
 journal = {Journal of Applied Probability},
 number = {1},
 pages = {147--161},
 title = {On the Maximum Drawdown of a Brownian Motion},
 volume = {41},
 year = {2004}
}

@misc{engstrom2020,
      title={Implementation Matters in Deep Policy Gradients: A Case Study on PPO and TRPO}, 
      author={Logan Engstrom and Andrew Ilyas and Shibani Santurkar and Dimitris Tsipras and Firdaus Janoos and Larry Rudolph and Aleksander Madry},
      year={2020},
      eprint={2005.12729},
      archivePrefix={arXiv},
      primaryClass={cs.LG},
      url={https://arxiv.org/abs/2005.12729}, 
}

@misc{korniejczuk2024,
      title={Statistical arbitrage in multi-pair trading strategy based on graph clustering algorithms in US equities market}, 
      author={Adam Korniejczuk and Robert Ślepaczuk},
      year={2024},
      eprint={2406.10695},
      archivePrefix={arXiv},
      primaryClass={q-fin.PM},
      url={https://arxiv.org/abs/2406.10695}, 
}

@Article{johansen1988,
journal={Journal of Economic Dynamics and Control},
author={Johansen, Soren},
title={Statistical analysis of cointegration vectors},
year={1988},
pages={231-254},
volume={12},
number={2},
url={https://doi.org/10.1016/0165-1889(88)90041-3},
}

@article{asness2012,
author = {Asness, Clifford and Frazzini, Andrea and Pedersen, Lasse},
year = {2012},
pages = {47--59},
title = {Leverage Aversion and Risk Parity},
volume = {68},
number = {1},
journal = {Financial Analysts Journal},
url = {https://doi.org/10.2469/faj.v68.n1.1}
}

@article{henderson2018, 
title={Deep Reinforcement Learning That Matters}, 
volume={32}, 
url={https://doi.org/10.1609/aaai.v32i1.11694},
number={1}, 
journal={Proceedings of the AAAI Conference on Artificial Intelligence}, 
author={Henderson, Peter and Islam, Riashat and Bachman, Philip and Pineau, Joelle and Precup, Doina and Meger, David}, 
year={2018}, 
}

@article{thorp2011,
title = {Chaper 9 - The kelly criterion in blackjack sports betting, and the stock market*},
editor = {S.A. Zenios and W.T. Ziemba},
journal = {Handbook of Asset and Liability Management},
pages = {385-428},
year = {2008},
volume = {1},
url = {https://doi.org/10.1016/B978-044453248-0.50015-0},
author = {Edward O. Thorp},
}

@article{Elliott2005,
author = {Robert J. Elliott and John Van Der Hoek * and William P. Malcolm},
title = {Pairs trading},
journal = {Quantitative Finance},
volume = {5},
number = {3},
pages = {271--276},
year = {2005},
publisher = {Routledge},
URL = { https://doi.org/10.1080/14697680500149370
},
}

@Article{Moura2013,
journal={Brazilian Review of Finance},
author={João Frois Caldeira and Gulherme Valle Moura},
title={Selection of a Portfolio of Pairs Based on Cointegration: A Statistical Arbitrage Strategy},
year={2013},
pages={49-80},
volume={11},
number={1},
url={https://doi.org/10.12660/rbfin.v11n1.2013.4785},
}

@article{Vergara2024,
title = {Deep reinforcement learning applied to statistical arbitrage investment strategy on cryptomarket},
journal = {Applied Soft Computing},
volume = {153},
pages = {111255},
year = {2024},
url = {https://doi.org/10.1016/j.asoc.2024.111255},
author = {Gabriel Vergara and Werner Kristjanpoller},
}

@article{Politis1994,
author = {Dimitris N. Politis and Joseph P. Romano},
title = {The Stationary Bootstrap},
journal = {Journal of the American Statistical Association},
volume = {89},
number = {428},
pages = {1303--1313},
year = {1994},
publisher = {Taylor \& Francis},
URL = { https://doi.org/10.1080/01621459.1994.10476870
},
}

\end{document}